\documentclass[lettersize,journal]{IEEEtran}
\usepackage{amsmath,amsfonts}
\usepackage[noend]{algorithmic}
\usepackage{algorithm}
\usepackage{array}
\usepackage[caption=false,font=normalsize,labelfont=sf,textfont=sf]{subfig}
\usepackage{textcomp}
\usepackage{siunitx}
\usepackage{stfloats}
\usepackage{url}
\usepackage{verbatim}
\usepackage{graphicx}
\usepackage{cite}
\usepackage{orcidlink}
\usepackage{booktabs}
\usepackage{multirow}
\begin{document}

\title{FugSeg: Fast Uncertainty-aware Ground Segmentation for 3D Point Cloud}

\author{Yu Li\orcidlink{0000-0002-4652-2037}, Volker Schwieger\orcidlink{0000-0001-9055-9809}
\thanks{Yu Li is with the Institute of Engineering Geodesy, University of Stuttgart, Stuttgart, Germany, and also with Daimler Truck AG, Leinfelden-Echterdingen, Germany (e-mail: yu.y.li@daimlertruck.com)}
\thanks{Prof. Dr.-Ing. habil. Volker Schwieger is with the Institute of Engineering Geodesy, University of Stuttgart, Stuttgart, Germany (e-mail: volker.schwieger@iigs.uni-stuttgart.de)}}

\markboth{IEEE TRANSACTIONS ON INTELLIGENT TRANSPORTATION SYSTEMS}%
{Li \MakeLowercase{\textit{et al.}}: FugSeg: Fast Uncertainty-aware Ground Segmentation for 3D Point Cloud}

\maketitle
\begingroup
\renewcommand{\thefootnote}{}
\footnotetext{This is the author accepted manuscript of an article published in \textit{IEEE Transactions on Intelligent Transportation Systems}. The final authenticated version is available at \url{https://doi.org/10.1109/TITS.2026.3682176}. \copyright~2026~IEEE. Personal use of this material is permitted. Permission from IEEE must be obtained for all other uses, in any current or future media, including reprinting/republishing this material for advertising or promotional purposes, creating new collective works, for resale or redistribution to servers or lists, or reuse of any copyrighted component of this work in other works.}
\addtocounter{footnote}{-1}
\endgroup
\begin{abstract}
In LiDAR-based environment perception systems, ground segmentation is a key preprocessing step supporting various applications such as mapping and navigation. Although extensively studied, problems such as reflection noise and isolated ground remain challenging. To address these issues, we propose FugSeg, a fast uncertainty-aware ground segmentation method. A polar grid map is adopted as the point cloud representation to ensure generalizability across LiDAR types. Building on that, we develop a within- and cross-segment ground labeling strategy that identifies not only directly visible ground cells but also those that are isolated or occluded. During this process, an adaptive slope is introduced, which incorporates measurement uncertainties to enhance its reliability under complex terrain. Finally, to achieve point-level ground segmentation, a fine-grained ground elevation estimation method is introduced. Throughout the complete workflow, reflection noise is explicitly handled via the proposed \textit{noisy ground} cells. We conduct comprehensive evaluations on four public datasets covering both structured and unstructured environments. Results show that FugSeg outperforms state-of-the-art non-learning methods, achieving the highest $\mathbf{F_1}$, accuracy, and mIoU across all datasets, while maintaining the fastest runtime (135 Hz and 487 Hz for 64- and 32-layer LiDARs) using a single CPU thread, making it suitable for resource-limited systems. The code will be available at \url{https://github.com/Leo-YuLi/FugSeg}.
\end{abstract}


\section{Introduction}
\begin{figure}[!t]
    \centering
    \includegraphics[width=3.5in]{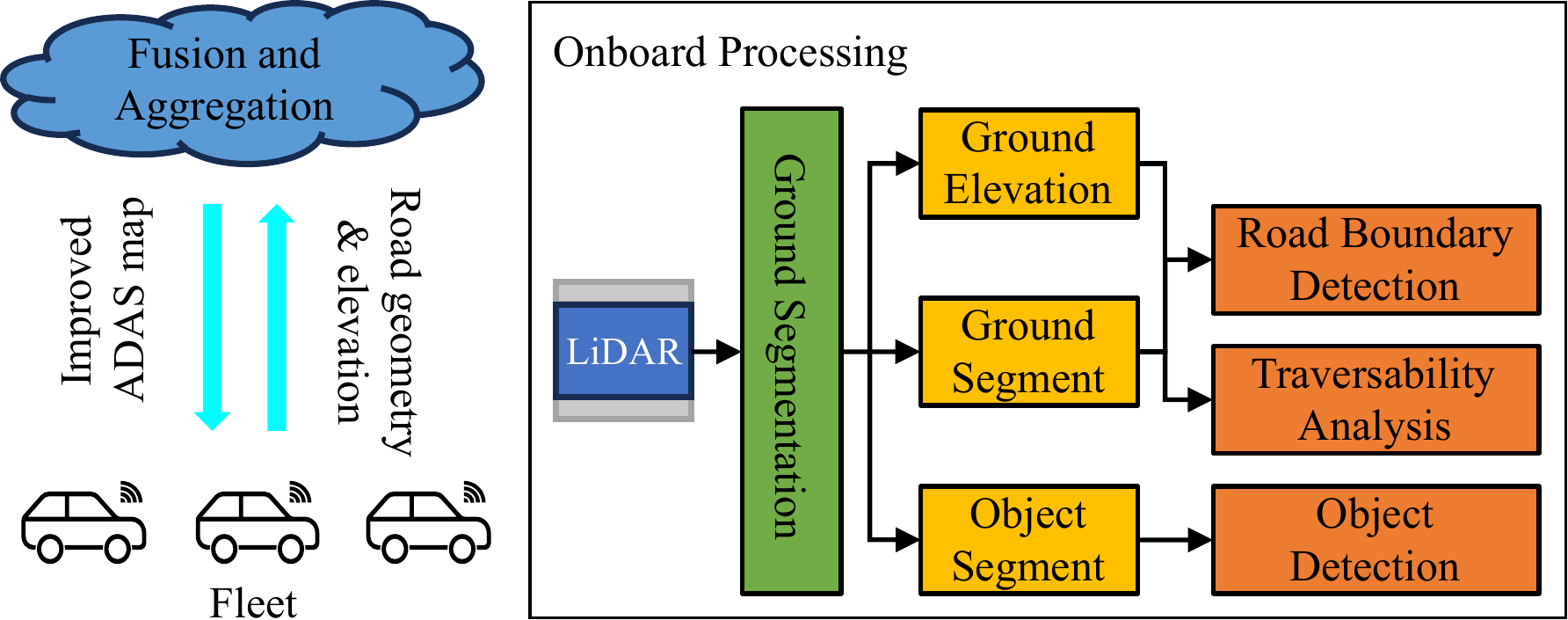}
    \caption{The role of ground segmentation in a LiDAR-centric intelligent transportation system. Onboard processing supports real-time environment perception for automated and connected vehicles, while offboard processing aggregates road geometric features measured by the fleet to enhance map-based ADAS functions and improve transportation and logistics efficiency.}
    \label{fig10}
\end{figure}

\IEEEPARstart{W}{ith} the rapid development of advanced driver assistance systems (ADAS), LiDAR has become a key enabler for numerous automated applications. 3D point clouds generated by LiDAR sensors provide rich geometric information about the surrounding environment, making them suitable for various perception tasks such as road boundary detection\cite{8291612}, traversability analysis\cite{Lim2024}, mapping\cite{ZhangRSS14} and object detection\cite{Petrovskaya2009}. Moreover, the onboard perception results can be further aggregated offboard to enhance map-based ADAS functions, thereby improving transportation and logistics efficiency\cite{8468109}. As illustrated in Figure~\ref{fig10}, as a fundamental requirement shared across these applications, ground segmentation not only reduces scene complexity but also improves the robustness and efficiency of subsequent perception modules.

For efficient ground segmentation, both fitting- and filtering-based solutions have been explored. Fitting-based methods typically estimate a 3D plane within a predefined region and classify ground points by evaluating point-to-plane fitness\cite{358692,7989591,8569534,9361109,9466396,9981561,9794594}, whereas filtering-based approaches identify ground points by thresholding the slope between adjacent laser channels\cite{Petrovskaya2009,Chu2017AFG,Chu2019,bonn2017}. Despite these advances, ground segmentation remains challenging in real-world deployments. First, noise caused by reflective materials often introduces spurious low-elevation points that are easily misclassified as ground. Second, isolated or occluded ground regions in cluttered scenes are difficult to detect using purely local geometric cues. Finally, complex or irregular terrain poses challenges for the traditional slope calculation method that is widely used in existing approaches.

A variety of methods have been proposed to address these challenges. Jiménez et al.\cite{9548034} labeled points below an estimated ground plane as noise; however, this method focuses only on regions near the ego-vehicle. The recent work in\cite{10319084} identified reflection noise by evaluating line-of-sight plausibility using a constantly tracked elevation map, but this method requires precise ego-motion estimation. To remove mirrored reflections, Yun et al.\cite{8792082} estimated glass planes to determine the optimal reflection trajectory for each point. Beyond these geometric criteria, intensity-based features have also been employed for noise filtering in various automotive and terrestrial applications\cite{rs13163058,9981561,FANG2025217}. Although effective in specific setups, these methods often require sensor-dependent fine-tuning, which limits their generalizability across different hardware platforms. To detect isolated ground regions, some works have incorporated horizontal processing\cite{Chu2019} or hybrid representations\cite{rs13163239,9548034} into the pipeline. However, these methods often introduce additional computational overhead and complexity. Regarding slope estimation, Wen et al.\cite{10359455} proposed a heuristic proxy-based method to mitigate short-baseline effects, but it relies on detailed intrinsic laser layout and is not directly applicable to heterogeneous LiDAR types.

To address these issues, FugSeg, a fast uncertainty-aware ground segmentation method, is proposed. Evaluated across eleven mechanical-spinning and solid-state LiDARs from five datasets, FugSeg demonstrates superior performance over other non-learning methods, while also achieving the fastest runtime under single-threaded execution. The key contributions of this work are summarized as follows:
\begin{itemize}
\item{An adaptive slope calculation method is proposed to improve the traditional slope calculation by incorporating measurement uncertainties.}
\item{A hybrid within- and cross-segment ground labeling strategy is developed to handle occlusions and isolated ground, thereby achieving a balance between precision and recall.}
\item{A fine-grained ground elevation estimation method is introduced for point-level segmentation, which accounts for the effects of non-flat terrain and reflection noise.}
\end{itemize}

\section{Related Work}
Recent studies have shown that using multiple point cloud representations can improve ground segmentation performance\cite{rs13163239,9548034}. Consequently, a purely representation-based taxonomy is no longer sufficient. Instead, we categorize existing approaches by their algorithmic characteristics, distinguishing between iterative optimization methods and deterministic single-pass methods. Machine learning based approaches typically involve an iterative training process followed by deterministic inference; therefore, they are discussed separately.

\subsection{Iterative Methods}
Methods that explicitly fit a ground model generally rely on iterative optimization. The most intuitive example is RANSAC\cite{358692}, which approximates the ground as a single 3D plane. To better accommodate non-flat terrain, a piecewise plane-fitting approach was proposed in\cite{7989591}, where the point cloud is divided into 3 sub-regions along the driving direction, and each region is independently fitted with a plane. In this framework, the RANSAC-based plane estimator is replaced by an iterative PCA-based procedure, significantly improving the efficiency.

Similarly, Narksri et al.\cite{8569534} projected the point cloud into a polar grid map and applied RANSAC-based plane fitting to individual cells, enhanced by a continuity constraint between cells sharing the same azimuth. This multi-region plane-fitting paradigm has since been further advanced. Patchwork\cite{9466396} introduced a concentric-zone polar grid representation to improve the efficiency of\cite{9361109}, and later Patchwork++\cite{9981561} incorporated adaptive parameterization to refine the initial ground estimates. Most recently, TRAVEL\cite{9794594} strengthened the enforcement of plane continuity between adjacent cells through local convexity-concavity analysis. Despite these developments, iterative methods inherently depend on initialization conditions and iteration counts, potentially leading to unstable performance under challenging scenarios.

\subsection{Deterministic Methods}
In contrast to iterative methods, typical deterministic, single-pass approaches offer predictable and bounded computational cost. For mechanical-spinning LiDARs, a full scan comprises a sequence of vertical laser slices acquired at consecutive time instances. By filtering slope changes along each slice from near to far, ground and non-ground points can be identified. This idea was successfully applied in the DARPA Urban Challenge 2007\cite{Petrovskaya2009}. Chu et al.\cite{Chu2017AFG} later introduced additional constraints to mitigate over-segmentation of distant points, followed by horizontal ground labeling to improve recall\cite{Chu2019}. However, these methods rely heavily on the LiDAR's intrinsic layout and often result in over-parameterized solutions.

A major advancement in laser-channel based methods is the introduction of the range-image representation\cite{5164280,bonn2017}, which enables 3D segmentation to be performed in a 2D space using standard image-processing techniques such as smoothing and dilation\cite{bonn2017,rs13163239,10359455}. While effective for suppressing isolated measurement noise, the 3D-to-2D projection can introduce geometric and topological distortions between adjacent pixels, making subsequent 2D processing unreliable\cite{9578697}.

Unlike sensor-dependent range images, grid maps offer a generic representation for 3D point clouds. A representative method is LineFit\cite{5548059}, which identifies ground points within each discretized segment by incrementally fitting 3D lines to consecutive cells under slope and $y$-intercept constraints. Beyond slope, additional statistical features such as Z-variance and point count were also employed for ground segmentation\cite{10319084}. Although effective in specific setups, these features are often sensor-dependent and require extensive fine-tuning. To leverage the strengths of different representations, Shen et al.\cite{rs13163239} used a polar grid map for coarse segmentation followed by a range-image based refinement. A similar strategy was also adopted in\cite{9548034}, where a Markov Random Field (MRF) is constructed to achieve point-level ground segmentation. However, besides the increased computational cost, these hybrid methods still inherit the limitations of their individual representations.

\subsection{Learning-based Methods}
\label{sec.2.c}
Instead of relying on hand-crafted features, learning-based methods enable models to implicitly learn discriminative representations from data. Velas et al.\cite{8374167} formulated ground segmentation as a semantic segmentation problem and applied a convolutional neural network (CNN) directly to the range image. An indirect strategy was proposed by Paigwar et al.\cite{9340979}, in which the point cloud is discretized to a $1m\times 1m$ grid map, and the ground elevation of each cell is regressed using a cascaded PointNet\cite{8099499}--CNN architecture. Point-level ground labels are then obtained by thresholding the distance between points and the estimated elevation surface. A similar approach is adopted in\cite{9691325}. However, direct semantic segmentation may yield suboptimal results when the image resolution does not preserve the geometric structure of the raw point cloud, and indirect methods cannot capture in-cell elevation variations on sloped surfaces. Moreover, the demand for substantial training data and high computational effort may limit their deployment in real-time or resource-constrained applications. 

\section{Methodology}
\begin{figure}[!t]
    \centering
    \includegraphics[width=3.5in]{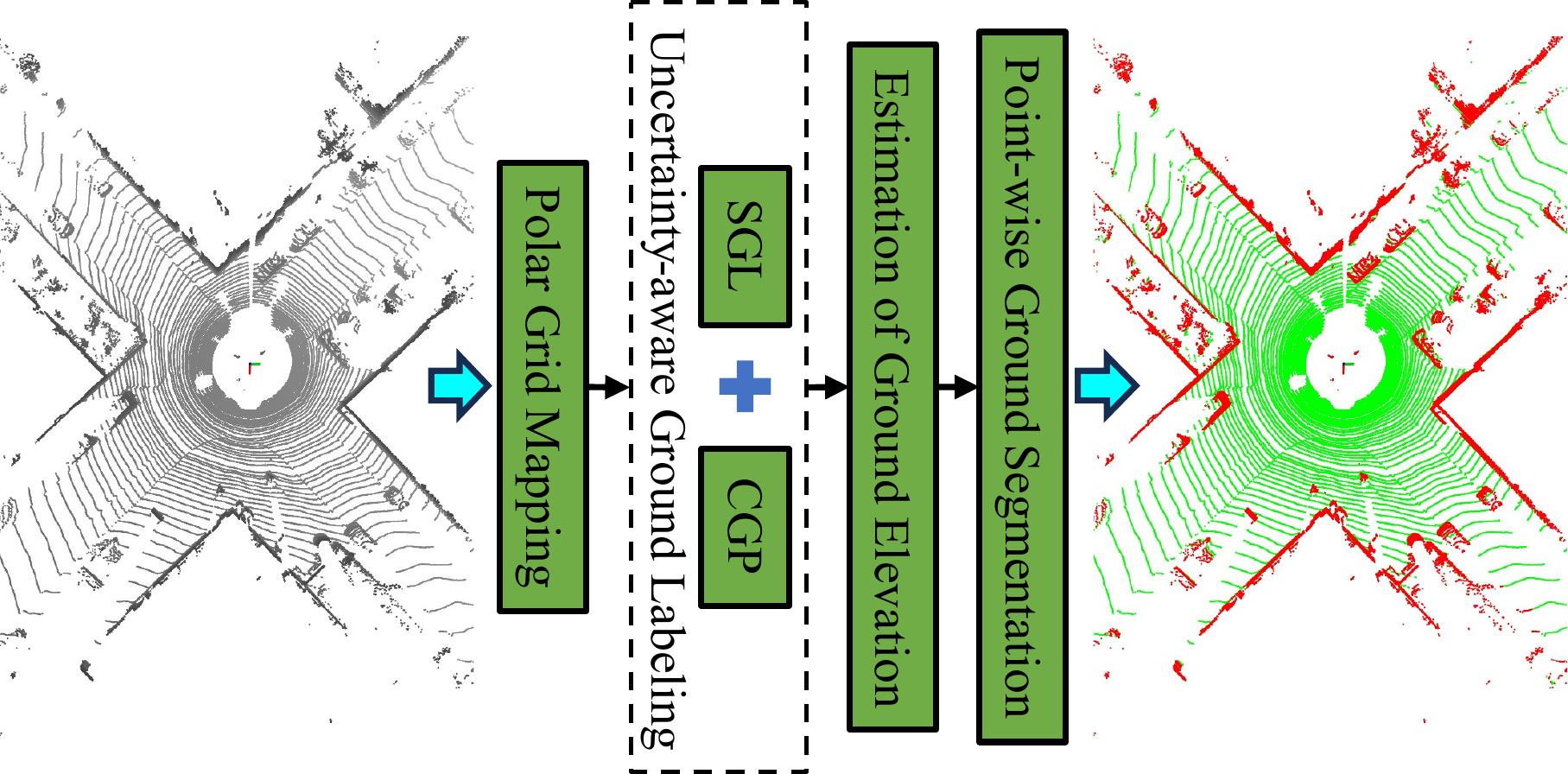}
    \caption{Overview of FugSeg. SGL: segment-wise ground labeling; CGP: cross-segment ground propagation.}
    \label{fig1}
\end{figure}

Figure~\ref{fig1} provides an overview of FugSeg. First, the input 3D point cloud is projected onto a polar grid map to enable efficient neighborhood retrieval. Next, ground cells are identified through a two-stage process: segment-wise ground labeling followed by cross-segment ground propagation. Then, ground elevation is estimated at the point level using all cells labeled as ground, which enables the fine-grained ground segmentation for individual points within each cell in the final step.

\subsection{Polar Grid Mapping (PGM)}
Typically, a 3D point cloud consists of a set of unordered points represented by their 3D Euclidean coordinates in the sensor coordinate system: $\mathcal{P}_t=\{p_1,\dots,p_K\}$, with $p_k=(x_k,y_k,z_k)$, and $K$ representing the total number of points. To ensure the generalizability of the resulting algorithm, we adopt a polar grid map as the point cloud representation. As illustrated in Figure~\ref{fig2}a, a polar grid map partitions the 3D space into a 2.5D structure on the XOY plane. Specifically, the XOY plane is divided into $L$ segments with an equal angular resolution $\Delta\alpha$, where $L=\frac{2\pi}{\Delta\alpha}$. Each segment is then further subdivided into $M$ cells along the radial direction from the origin. Consequently, for a given 3D point $p_k$, its corresponding segment and cell indices can be computed as:
\begin{align}
    \label{equ.3.a.1}
    \begin{cases}
        i = floor\big(\frac{\pi-\text{atan2}(y_k,x_k)}{\Delta\alpha}\big), &i\in[0,L) \\[1ex]
        j \iff r_j\leq \big(x_k^2+y_k^2\big)^\frac{1}{2} < r_{j+1}, &j\in[0,M)
    \end{cases}
\end{align}

After polar grid mapping, 3D points within the valid range $[r_0,r_M)$ are projected into their corresponding cells, while all points outside this range are directly classified as non-ground. For each non-empty cell, the point with the lowest $Z$-coordinate is selected as the representative point, as it is most likely to lie on the ground surface. For clarity in the subsequent discussion, we define a \textit{row} as the set of all cells equidistant from the origin, ordered by ascending azimuth angle.

\subsection{Uncertainty-aware Ground Labeling (UGL)}
\begin{figure}[!t]
\centering
\begin{tabular}{cc}
    \includegraphics[width=1.4in]{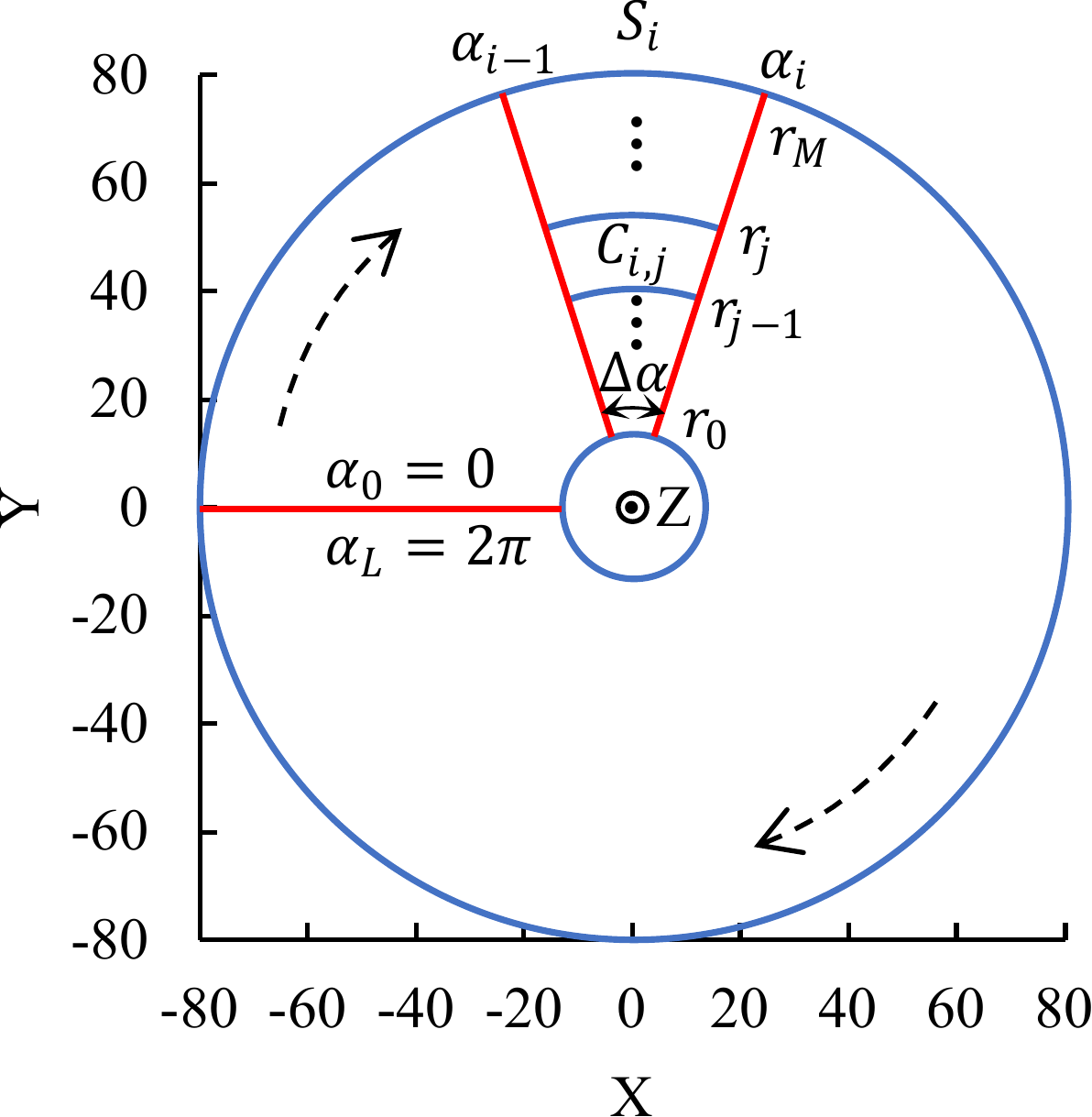} &
    \includegraphics[width=2.0in]{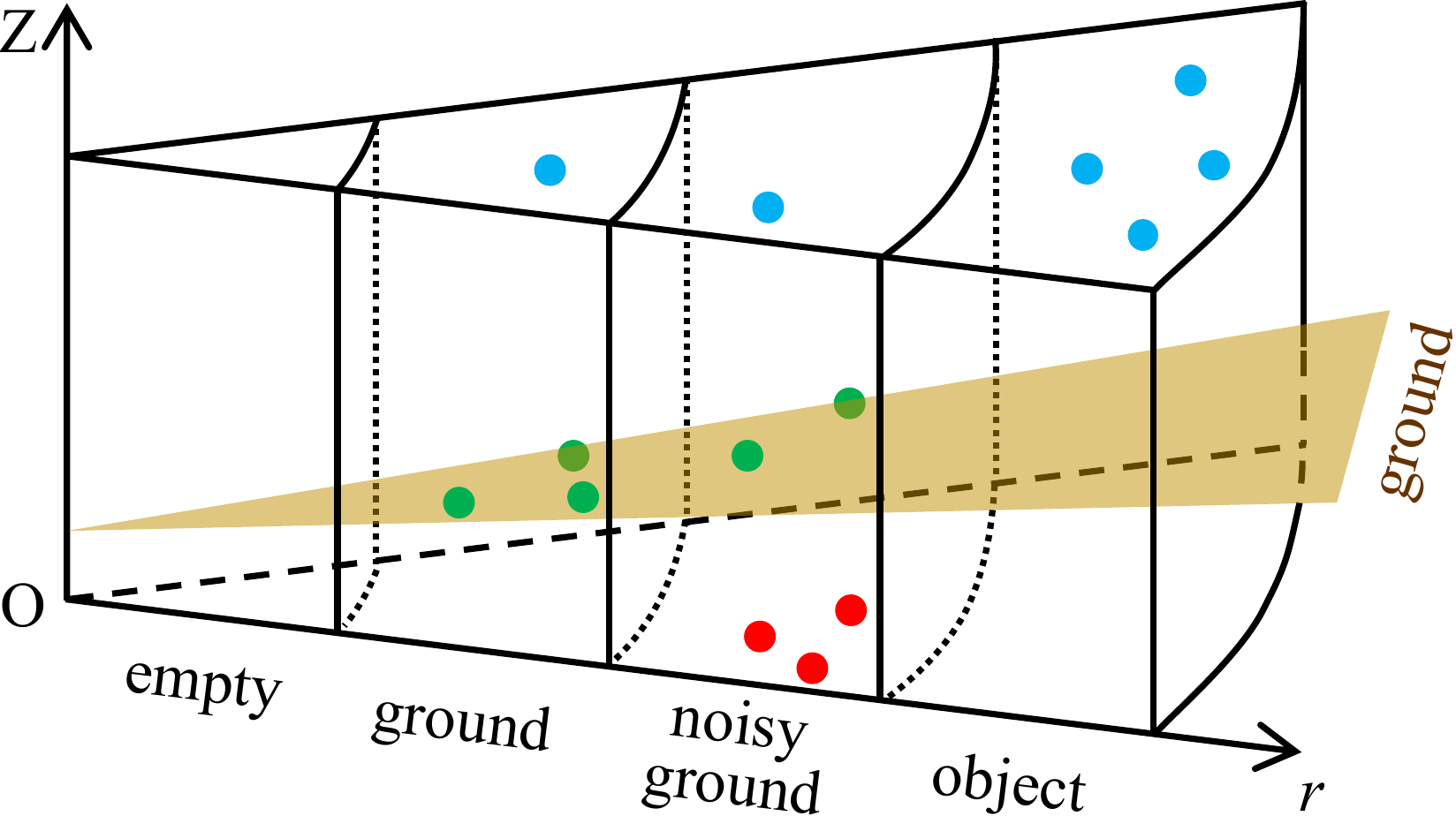} \\
    (a) & (b) \\
\end{tabular}
\caption{(a) Polar grid mapping, $C_{i,j}$ represents the $j^{th}$ cell in segment $S_i$. (b) Categorization of cells, Green: ground points; red: reflection noise; blue: above-ground objects.}
\label{fig2}
\end{figure}

As illustrated in Figure~\ref{fig2}b, the initially constructed polar grid map contains empty and non-empty cells, and non-empty cells can be further categorized into object and ground cells based on the presence of ground points. Note that some ground cells may contain reflection artifacts below the actual ground surface, which is caused by laser interference with reflective objects\cite{rs13163058,8792082,FANG2025217}. The objective of ground labeling is to identify all ground cells (including \textit{noisy ground} cells), which is achieved through the proposed slope criterion that is adaptively determined by incorporating measurement uncertainties.

\setlength{\tabcolsep}{1.0pt}
\setlength{\tabcolsep}{6pt}

\subsubsection{Uncertainty-aware Slope Calculation}
\begin{figure}[!t]
    \centering
    \includegraphics[width=2.8in]{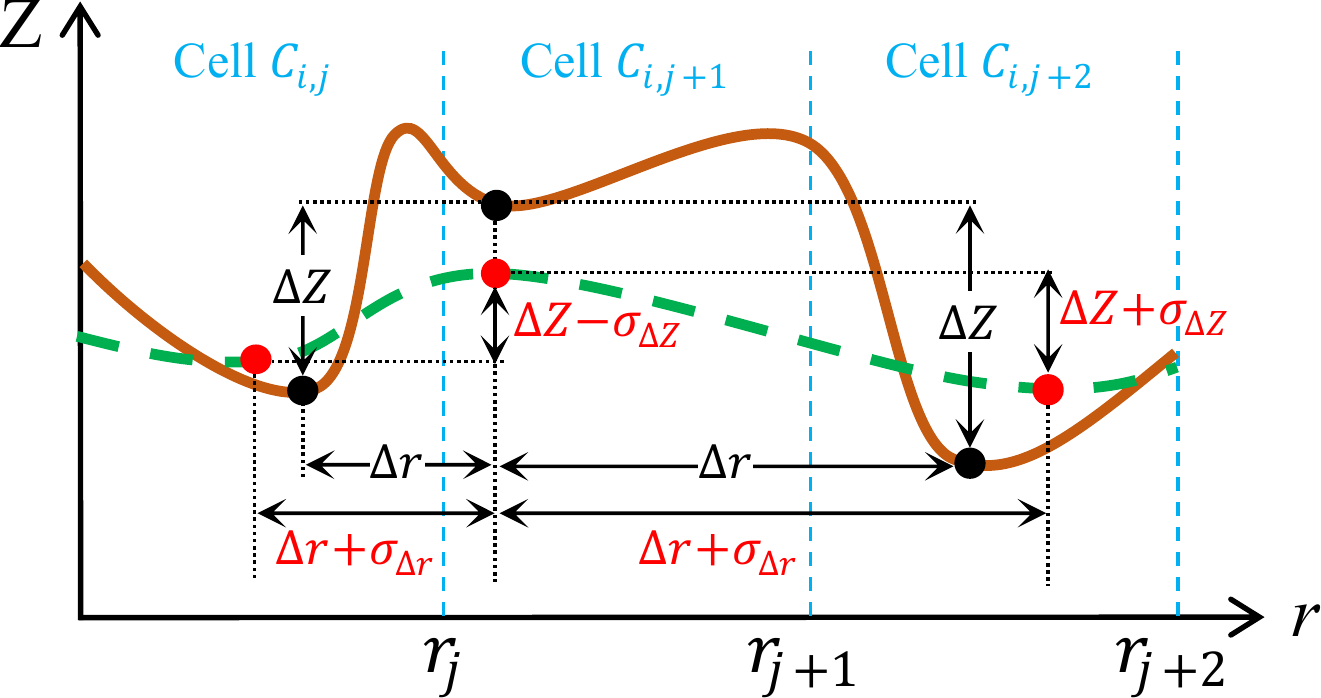}
    \caption{Traditional slope (black) versus the proposed adaptive slope (red) in complex terrain (denoted by brown curve). Green dashed curve: (virtually) smoothed terrain; black dot: representative point of a cell; red dots: compensated representative points. Note that $\Delta Z$ is a signed variable.}
    \label{fig9}
\end{figure}

The traditional slope (TS) formulation in~(\ref{equ.3.b.1}) generally works when vector $\overrightarrow{p_k p_l}$ is well formed. However, as illustrated in Figure~\ref{fig9}, due to the effects of irregular terrain undulations and measurement inaccuracies, Equation~(\ref{equ.3.b.1}) may fail to reflect the underlying terrain variation between neighboring cells, particularly in the case of short baselines (i.e., $\Delta r\to 0$) or abrupt height changes (i.e., $|\Delta Z|\gg |\Delta r|$). This may further result in misclassification of ground cells, thereby limiting its applicability across different terrain conditions. Wen et al.\cite{10359455} addressed this issue by extending the slope vector using a nearby proxy point that meets a minimum baseline requirement. However, this proxy selection is sensor-dependent and specific to range-image based representations. In this work, we introduce an adaptive slope calculation method.

\begin{equation}
    \label{equ.3.b.1}
    TS(p_k,p_l) = \frac{\Delta Z}{\Delta r} = \frac{z_l-z_k}{\big[(x_l-x_k)^2+(y_l-y_k)^2\big]^\frac{1}{2}}
\end{equation}

\begin{equation}
    \label{equ4}
    \begin{cases}
    x = R \cdot \cos\phi \cdot \sin\theta \\
    y = R \cdot \cos\phi \cdot \cos\theta \\
    z = R \cdot \sin\phi
    \end{cases}
\end{equation}

Taking a mechanical-spinning LiDAR as an example, Equation~(\ref{equ4}) defines the conversion from raw spherical coordinates $($radius $R,$ elevation angle $\phi,$ azimuth $\theta)$ to Cartesian coordinates $(x,y,z)$ for a single 3D point. As a measurement property, the raw spherical coordinates are associated with corresponding standard deviations $\sigma_R$, $\sigma_\phi$ and $\sigma_\theta$. Assuming that $R$, $\phi$ and $\theta$ are uncorrelated variables, the standard deviation of each Cartesian coordinate can be obtained following the law of variance-covariance propagation\cite{bevington2003data} as follows:
\begin{equation}
\begin{split}
    \label{equ5}
    \sigma_x = \big(\cos^2\phi \cdot \sin^2\theta \cdot \sigma^2_R + R^2 \cdot \sin^2\phi \cdot \sin^2\theta \cdot \sigma^2_\phi \\
                + R^2 \cdot \cos^2\phi \cdot \cos^2\theta \cdot \sigma^2_\theta\big)^\frac{1}{2}
\end{split}
\end{equation}
\begin{equation}
\begin{split}
    \label{equ6}
    \sigma_y = \big(\cos^2\phi \cdot \cos^2\theta \cdot \sigma^2_R + R^2 \cdot \sin^2\phi \cdot \cos^2\theta \cdot \sigma^2_\phi \\
                + R^2 \cdot \cos^2\phi \cdot \sin^2\theta \cdot \sigma^2_\theta\big)^\frac{1}{2}
\end{split}
\end{equation}
\begin{equation}
    \label{equ7}
    \sigma_z = \big(\sin^2\phi \cdot \sigma^2_R + R^2 \cdot \cos^2\phi \cdot \sigma^2_\phi\big)^\frac{1}{2}
\end{equation}
Likewise, standard deviations of the height difference $\Delta Z$ and the baseline length $\Delta r$ can then be determined as:
\begin{equation}
    \label{equ8}
    \sigma_{\Delta Z} = \big(\sigma^2_{z_l} + \sigma^2_{z_k}\big)^\frac{1}{2}
\end{equation}
\begin{equation}
\begin{split}
    \label{equ9}
        \sigma_{\Delta r} &= \Big[\big(\frac{\Delta X}{\Delta r}\big)^2\big(\sigma^2_{x_l}+\sigma^2_{x_k}\big) + \big(\frac{\Delta Y}{\Delta r}\big)^2\big(\sigma^2_{y_l}+\sigma^2_{y_k}\big)\Big]^\frac{1}{2}, \\
        &\qquad \text{with } \Delta X=x_l-x_k \text{ and } \Delta Y=y_l-y_k
\end{split}
\end{equation}
With these estimated uncertainties, the proposed adaptive slope (AS) is computed as follows:
\begin{align}
    \label{equ10}
    {AS(p_k,p_l)} = \begin{cases}
            0,&{\text{if }}\ |\Delta Z| \leq {\sigma_{\Delta Z}} \\[1ex]
            {\frac{\Delta Z-\sigma_{\Delta Z}}{\Delta r+\sigma_{\Delta r}},}&{\text{if }} \Delta Z>{\sigma_{\Delta Z}} \\[1ex]
            {\frac{\Delta Z+\sigma_{\Delta Z}}{\Delta r+\sigma_{\Delta r}},}&{\text{if }} \Delta Z<{-\sigma_{\Delta Z}}
    \end{cases}
\end{align}

As illustrated in Figure~\ref{fig9}, the proposed adaptive slope acts as a low-pass filter, which essentially smooths abrupt changes in both the horizontal and vertical directions. In Equation~(\ref{equ10}), the horizontal term $\Delta r+\sigma_{\Delta r}$ compensates for short baselines, whereas the vertical terms $\Delta Z-\sigma_{\Delta Z}$ and $\Delta Z+\sigma_{\Delta Z}$ account for height deviations caused by irregular terrain undulations. Note that a one-sigma uncertainty measure is used here; in practice, however, two or more sigmas may be selected depending on the measurement quality. For the ground segmentation task, the proposed adaptive slope demonstrates improved effectiveness over the traditional slope, as shown in Section~\ref{sec_ablation_uagl}.

\subsubsection{Segment-wise Ground Labeling (SGL)}
\begin{algorithm}[tb]
    \caption{Segment-wise Ground Labeling}
    \label{alg1}
    \begin{algorithmic}[1]
    \renewcommand{\algorithmicrequire}{\textbf{Input:}}
    \renewcommand{\algorithmicensure}{\textbf{Output:}}
    \REQUIRE $j_{seed}$ radial index of the ground seed in segment $S_i$
    \ENSURE  $label_{i,j}$ label of each cell in segment $S_i$
    \STATE $j_{last} \gets j_{seed}$, $s_{last} \gets slope(O,C_{i,j})$
    \\ \text{Forward Labeling:}
    \FOR {$j = j_{last}+1$ to $M-1$}
        \IF {$\Delta r({j_{last},j})<T_{\Delta r}$}
            \IF {$|s_{last}-slope(C_{i,j_{last}},C_{i,j})|<T_{\Delta slope}$}
                \STATE $j_{last} \gets j$
                \STATE $s_{last} \gets slope(C_{i,j_{last}},C_{i,j})$
                \STATE $label_{i,j} \gets ground$
            \ELSIF {$slope(C_{i,j_{last}},C_{i,j})<0$}
                \STATE $label_{i,j} \gets noisy\ ground$
            \ELSE
                \STATE $label_{i,j} \gets object$
            \ENDIF
        \ENDIF
    \ENDFOR
    \text{Backward Labeling:}
    \FOR {$j = j_{last}-2$ to $0$}
        \IF {$label_{i,j}\ne ground$ \textbf{and} $label_{i,j+1}=ground$ \\ \textbf{and} $label_{i,j+2}=ground$}
            \STATE $s_{last} \gets slope(C_{i,j+2},C_{i,j+1})$
            \IF {$|s_{last}-slope(C_{i,j+1},C_{i,j})|<T_{\Delta slope}$}
                \STATE $label_{i,j} \gets ground$
            \ELSIF {$slope(C_{i,j+1},C_{i,j})<0$}
                \STATE $label_{i,j} \gets noisy\ ground$
            \ELSE
                \STATE $label_{i,j} \gets object$
            \ENDIF
        \ENDIF
    \ENDFOR
    \end{algorithmic}
\end{algorithm}

To identify all ground cells, one common strategy is to carefully select a few cells as ground seeds and then progressively expand them toward more ground cells in a region-growing fashion. The proposed segment-wise ground labeling follows the same idea.

For horizontally leveled LiDAR sensors, two widely adopted assumptions are: (1) the ground slope does not change drastically between adjacent cells, and (2) the ground surface is more likely to be observed in the near field\cite{bonn2017,Chu2017AFG,9548034,rs13163239}. Motivated by these assumptions, for each segment $S_i$, we search along the radial direction from near to far and select the first cell $C_{i,j}$ that satisfies the following conditions as its ground seed:
\begin{itemize}
    \item{$Z(C_{i,j})<T_h$: the cell's $Z$-coordinate is below a threshold determined based on the sensor installation height $H_s$.}
    \item{$|slope(O,C_{i,j})|<T_{\Delta slope}$: the absolute slope with respect to the leveled ground defined by $O(0,0,-H_s)$ is below a given threshold.}
    \item{$|slope(O,C_{i,j})-slope(C_{i,j},C_{i,j+1})|<T_{\Delta slope}$: the slope change between this cell and its next neighbor is within the same threshold.}
\end{itemize}
Note that unlike most existing works, which filter directly on the slope itself, we instead constrain the slope change. This enhances the generalization ability of our approach in highly curved or inclined environments, where finding a proper absolute slope threshold is often challenging.

With the identified ground seed, additional ground cells within the same segment are progressively identified by incrementally evaluating the slope change between a confirmed ground cell and its unknown neighbors. Algorithm~\ref{alg1} presents our implementation. The segment-wise ground labeling is performed in two rounds. The forward round (line 2-11) expands from the initial ground seed toward the farthest cell, whereas the backward round (line 12-20) identifies additional ground cells between the last detected ground cell and the sensor origin. In the forward pass, a maximum baseline length threshold $T_{\Delta r}$ is introduced to reduce false positives in sparse or occluded regions (line 3). Likewise, a strict neighborhood constraint is applied in the backward pass to suppress false positives (line 13). Empty cells are automatically skipped in both rounds. After segment-wise ground labeling, four cell types are present in the resulting polar grid map: \textit{ground}, \textit{noisy ground}, \textit{object}, and \textit{empty}.

\subsubsection{Cross-segment Ground Propagation (CGP)}
\begin{figure}[!t]
    \centering
    \includegraphics[width=2.8in]{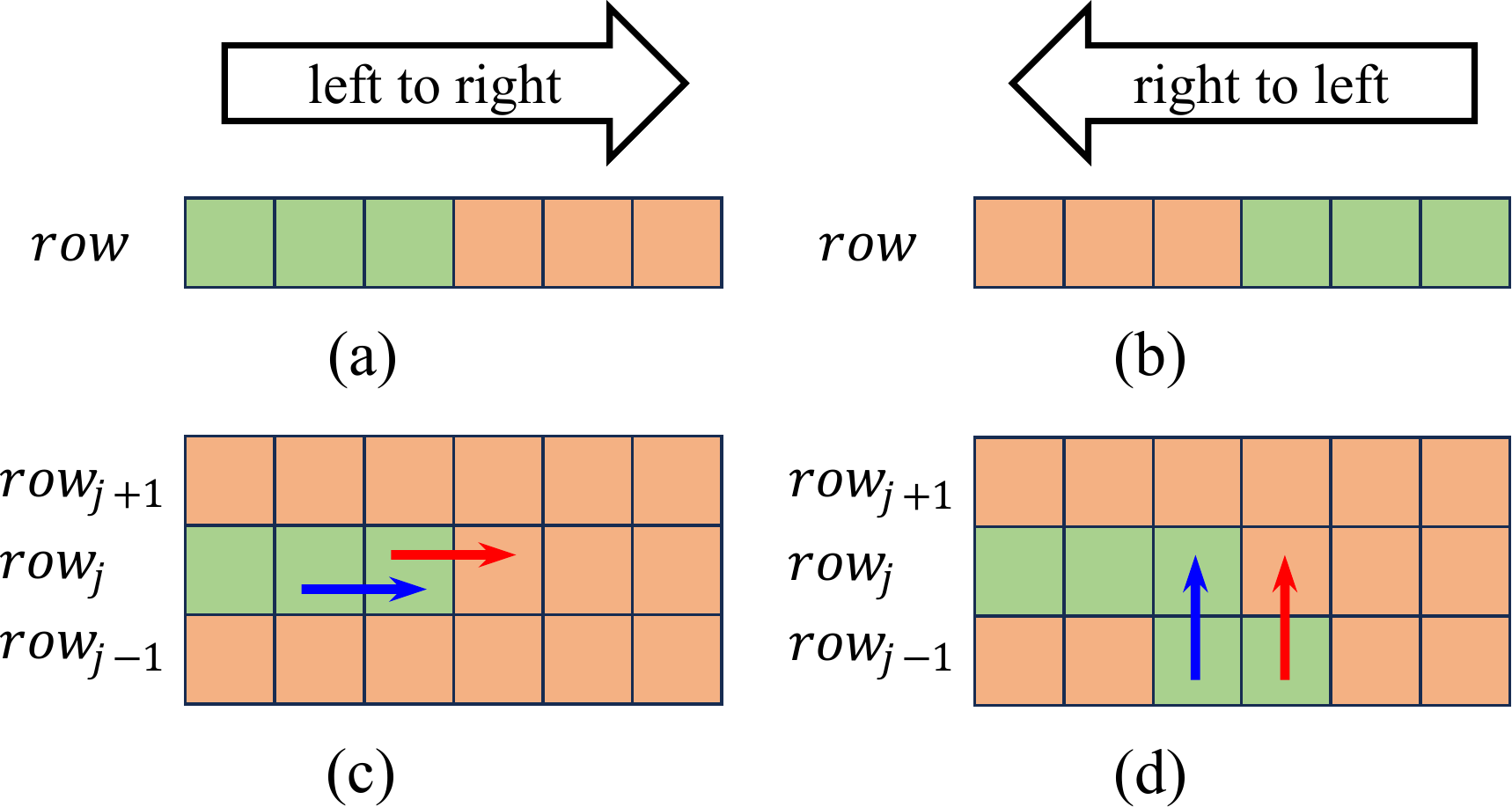}
    \caption{Cross-segment ground propagation. (a) Ground propagation from left to right. (b) Ground propagation from right to left. (c) Horizontal slope vectors constructed within the same row. (d) Vertical slope vectors constructed across rows. Green cell: ground; orange cell: non-ground; blue arrow: ground-slope baseline; red arrow: slope vector to be evaluated. Note that concentric sectors are depicted as rectangles for simplicity.}
    \label{fig4}
\end{figure}

\begin{algorithm}[tb]
    \caption{Cross-segment Ground Propagation (left to right)}
    \label{alg2}
    \begin{algorithmic}[1]
    \renewcommand{\algorithmicrequire}{\textbf{Input:}}
    \renewcommand{\algorithmicensure}{\textbf{Output:}}
    \REQUIRE current label of all cells in the polar grid map
    \ENSURE  $label_{i,j}$ updated label of non-ground cells
    \STATE $s_h \gets 0$, $s_v \gets 0$
    \FOR {$j = 0$ to $M-1$}
        \FOR {$i = 2$ to $L-1$}
            \IF {$label_{i-1,j}=ground$ \textbf{and} $label_{i,j}\ne ground$}
                \STATE $s_h \gets slope(C_{i-1,j},C_{i,j})$
                \STATE $s_v \gets \text{SlopeVertical}(i,j)$
                \IF {\big($label_{i-2,j}=ground$ \textbf{and} \\ $|slope(C_{i-2,j},C_{i-1,j})-s_h|<T_{\Delta slope}$\big) \\ \textbf{or} \big($|\text{SlopeVertical}(i-1,j)-s_v|<T_{\Delta slope}$\big)}
                    \STATE $label_{i,j} \gets ground$
                \ENDIF
            \ENDIF
        \ENDFOR
    \ENDFOR
    \STATE \textbf{function} \textsc{SlopeVertical}($i$, $j$)
    \begin{ALC@g}
        \IF{$label_{i,j-1}=ground$}
            \RETURN $slope(C_{i,j-1},C_{i,j})$
        \ELSIF{$label_{i,j+1}=ground$}
            \RETURN $slope(C_{i,j},C_{i,j+1})$
        \ELSE
            \RETURN $\inf$
        \ENDIF
    \end{ALC@g}
    \end{algorithmic}
\end{algorithm}

Segment-wise ground labeling primarily captures ground cells that are directly visible to the LiDAR sensor. However, cells occluded by other objects require an additional labeling mechanism. To address this, we develop a cross-segment ground propagation method, which aims to improve recall by extending ground labels from existing ground cells to their non-ground neighbors.

As illustrated in Figures~\ref{fig4}a and \ref{fig4}b, the proposed cross-segment ground propagation focuses on labeling along tangential directions. Depending on the availability of a second ground cell that is required to establish a ground-slope baseline, slope vectors can be constructed either horizontally or vertically, as shown in Figures~\ref{fig4}c and \ref{fig4}d, respectively. The evaluation of slope changes, however, must be performed between slope vectors of the same type.

Algorithm~\ref{alg2} illustrates the implementation for left-to-right propagation, from which the right-to-left propagation can be derived symmetrically. For a non-ground cell, propagation succeeds only if the slope change relative to its ground neighbor is below the threshold $T_{\Delta slope}$ (line 7). Building a horizontal ground-slope baseline requires at least two consecutive ground cells within the same row (line 4 and 7), whereas building a vertical ground-slope baseline requires at least one ground cell in adjacent rows (line 10 and 12). Because vertical slope vectors span multiple rows, propagation results between adjacent rows may be mutually influenced. To ensure completeness, Algorithm~\ref{alg2} can be executed twice in practice: once from near to far, and a second time from far to near.

\subsection{Estimation of Ground Elevation (EGE)}
\label{sec.3.c}

\begin{figure}[!t]
    \centering
    \includegraphics[width=2.7in]{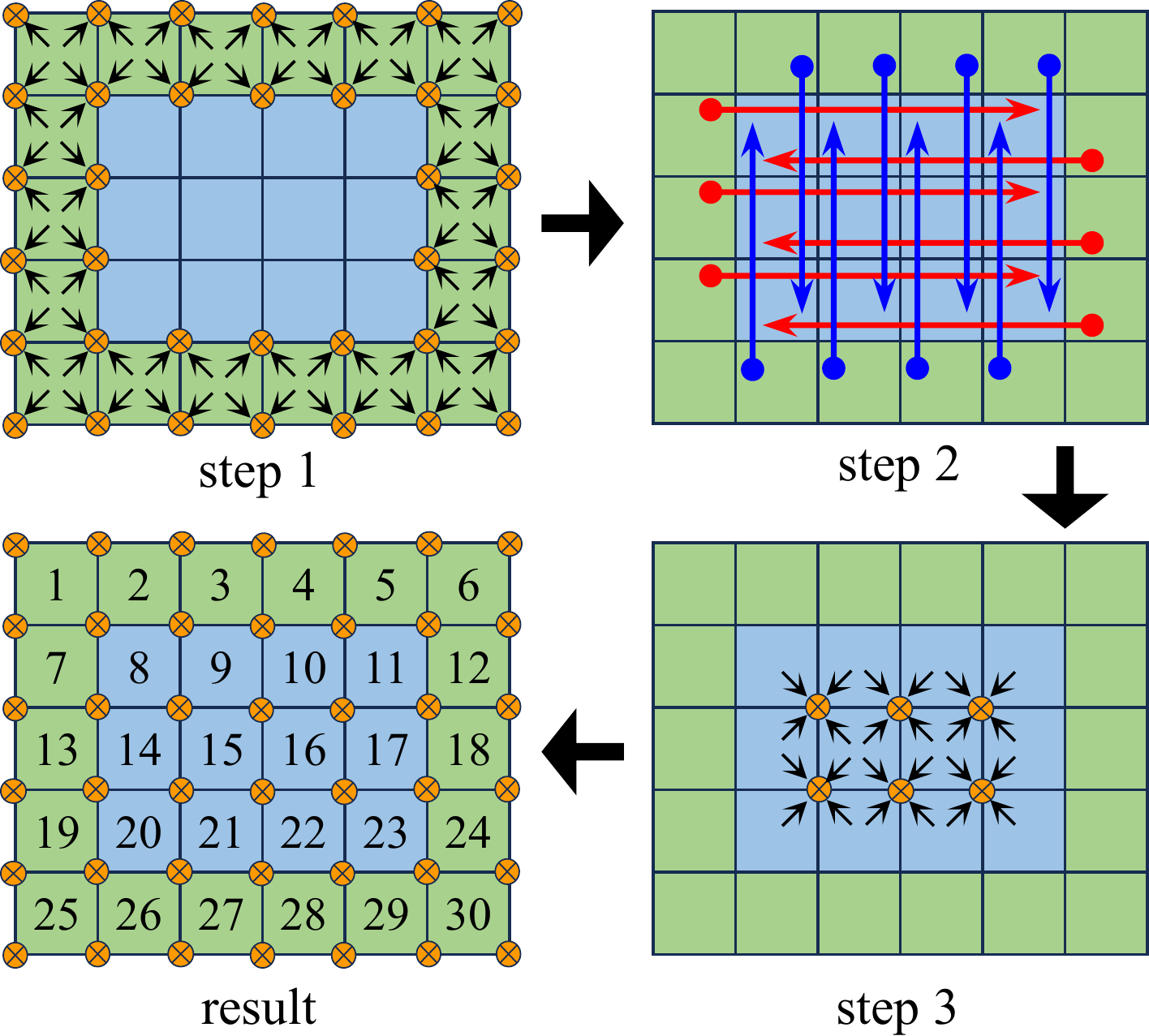}
    \caption{Three steps to determine the node height for \textit{ground} and \textit{noisy ground} cells. Step 1: height determination for \textit{ground} nodes. Step 2: four-path propagation of $Z$-coordinate from \textit{ground} cells to \textit{noisy ground} cells. Step 3: height determination for \textit{noisy ground} nodes.  Green cell: \textit{ground}; blue cell: \textit{noisy ground}; red and blue arrows: propagation paths of $Z$-coordinate, each path starts from a \textit{ground} cell and stops before reaching the next \textit{ground} cell; crossed circles in orange: nodes; black arrows (small): cells involved in the height calculation of a node.}
    \label{fig5}
\end{figure}

Ground labeling provides a coarse segmentation at the cell level. To achieve point-level ground segmentation, the next step is to estimate the ground elevation for all identified \textit{ground} and \textit{noisy ground} cells. As illustrated in Figure~\ref{fig6}, for a given cell, once the heights of its four nodes are known, the ground elevation at any position within this cell can be obtained via interpolation. Thus, the core of ground elevation estimation lies in determining the node height in the polar grid map.

As illustrated in Figure~\ref{fig5}, to determine the node height for \textit{ground} and \textit{noisy ground} cells, a three-step approach is proposed. The first step focuses on \textit{ground} nodes---nodes that are directly connected to \textit{ground} cells. For a \textit{ground} node $N_{i,j}$, $i\in[0,L)$ and $j\in[0,M]$, its height $H(N_{i,j})$ is computed as the weighted average $Z$-coordinate of its adjacent \textit{ground} cells:
\begin{equation}
\begin{split}
    \label{equ.3.c.1}
        H(N_{i,j}) = \frac{\sum (W\cdot Z)}{\sum W},
        \text{with } W =\exp(-\|\overrightarrow{N_{i,j} p}\|)
\end{split}
\end{equation}
where $W$ is a weighting factor determined based on the 2D distance $\|\overrightarrow{N_{i,j} p}\|$ between node $N_{i,j}$ and the representative point $p$ of a \textit{ground} cell, and $Z$ represents $p$'s $Z$-coordinate.

Unlike \textit{ground} cells, the $Z$-coordinates of \textit{noisy ground} cells represent the height of reflection noise rather than the actual ground. Therefore, before computing the node height for \textit{noisy ground} cells, their actual ground heights need to be determined first. For this purpose, a four-path $Z$-coordinate propagation method is proposed. As illustrated in step 2 of Figure~\ref{fig5}, the propagation is performed along four directions: row-wise from left to right and right to left (red arrows), and segment-wise from near to far and far to near (blue arrows). Each propagation starts from a \textit{ground} cell and stops before reaching the next \textit{ground} cell. During propagation, the $Z$-coordinate of the last visited \textit{ground} cell is assigned to the current \textit{noisy ground} cell.

After all four propagations, each \textit{noisy ground} cell may have up to four propagated $Z$-coordinates from different directions. The ground height of a \textit{noisy ground} cell is then estimated as the weighted average of these propagated $Z$-coordinates, with weighting factors determined by the distances to the corresponding source \textit{ground} cells. In the example of Figure~\ref{fig5}, \textit{noisy ground} cell 15 receives propagated $Z$-coordinates from all four directions (from \textit{ground} cells 13, 18, 3 and 27), and its ground height is consequently estimated as:
\begin{equation}
\begin{split}
    \label{equ.3.c.2}
        \hat{Z}_{15} &= \frac{w_{13}\cdot Z_{13} + w_{18}\cdot Z_{18} + w_3\cdot Z_3 + w_{27}\cdot Z_{27}}{w_{13} + w_{18} + w_3 + w_{27}}, \\
        &\quad \text{with } w_n =\exp(-d_{15,n}), n \in \{13, 18, 3, 27\}
\end{split}
\end{equation}
where $d_{15,n}$ denotes the distance between cells 15 and $n$.

Once the estimated ground heights of \textit{noisy ground} cells are available, the third step computes the node height for these cells using the same Equation~(\ref{equ.3.c.1}) applied to \textit{ground} cells, but with the estimated $Z$-coordinates from step 2. As a byproduct, the proposed elevation estimation produces a fine-grained ground elevation map that is free from reflection noise and more accurately represents the actual ground surface. This elevation map is particularly beneficial for the downstream tasks presented in Figure~\ref{fig10}, including traversability analysis and the fuel-efficiency oriented road elevation aggregation.

\subsection{Point-level Ground Segmentation (PGS)}

\begin{figure}[!t]
    \centering
    \includegraphics[width=1.8in]{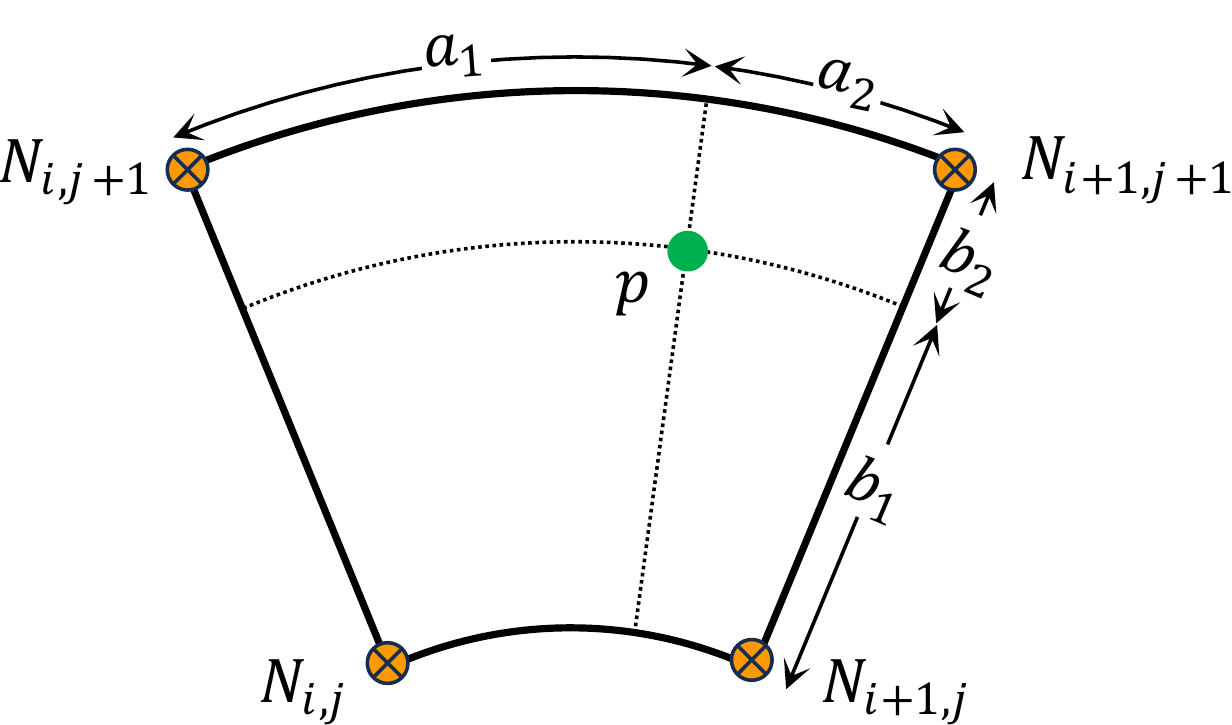}
    \caption{Elevation interpolation for arbitrary point $p$ within cell $C_{i,j}$.}
    \label{fig6}
\end{figure}

As shown in Figure~\ref{fig6}, for a point $p$ in \textit{ground} or \textit{noisy ground} cell $C_{i,j}$, its corresponding ground elevation $E(p)$ can be estimated through interpolation as follows:
\begin{equation}
    \label{equ.3.d.0}
    E(p) = \frac{
        \begin{aligned}
            &W_{bl} \cdot H(N_{i,j}) + W_{br} \cdot H(N_{i+1,j}) \\
            &\quad + W_{tl} \cdot H(N_{i,j+1}) + W_{tr} \cdot H(N_{i+1,j+1})
        \end{aligned}
    }{
        \sum W
    }
\end{equation}
where $i\in[0,L)$ and $j\in[0,M)$. $W_{bl}$, $W_{br}$, $W_{tl}$ and $W_{tr}$ are the weighting factors for the four nodes of cell $C_{i,j}$, which are determined based on their distances from $p$. The weighting Equation~(\ref{equ.3.c.1}) can be used to calculate these factors; however, it requires the time-consuming computation of an exponential function and a square root. In practice, a less-demanding alternative can be formulated by employing a linear weighting mechanism. As illustrated in Figure~\ref{fig6}, these four weighting factors can be calculated as follows:
\begin{equation}
    \begin{gathered}
    \label{equ.3.d.1}
        W_{bl} = a_2+b_2,\quad W_{br} = a_1+b_2, \\
        W_{tl} = a_2+b_1,\quad W_{tr} = a_1+b_1
    \end{gathered}
\end{equation}
where $a_1$ and $a_2$, and $b_1$ and $b_2$, represent the distance ratios along the tangential and radial directions, respectively.

Once the ground elevation is available, point $p$ can be segmented into ground or non-ground by comparing its $Z$-coordinate $Z(p)$ against the estimated $E(p)$. For points in \textit{ground} cells, the fine-grained segmentation is achieved via:
\begin{align}
    \label{equ.3.d.2}
    {label(p)} \gets \begin{cases}
            ground, &{\text{if}}\  Z(p)<E(p)+T_Z\\
            non\text{-}ground, &{\text{otherwise}}
    \end{cases}
\end{align}
For points in \textit{noisy ground} cells, another rule is applied to filter out the below-ground reflection noise:
\begin{align}
    \label{equ.3.d.3}
    {label(p)} \gets \begin{cases}
            ground, &{\text{if}}\  |Z(p)-E(p)|<T_Z\\
            non\text{-}ground, &{\text{otherwise}}
    \end{cases}
\end{align}
where $T_Z$ specifies the height threshold.

\section{Experimental Setup}
\begin{table}[!t]
\footnotesize
\setlength{\tabcolsep}{2.5pt}
\begin{center}
\caption{List of evaluated LiDARs. Unit: angle in [$^{\circ}$], distance in [$m$].}
\label{tab7}
        \begin{tabular}{*{9}c}
        \toprule
        \textbf{Dataset} & \textbf{Sensor} & \textbf{Type} & \textbf{\#scan} & {$\sigma_R$} & {$\sigma_\phi$} & {$\sigma_\theta$} & {$H_s$} & {$T_h$}  \\
        \midrule
        \scriptsize SemanticKITTI   & \scriptsize HDL64E & \scriptsize spinning & 23201 & 0.02 & 0.033 & 0.009 & 1.73 & -1.43  \\
        \hline
        \scriptsize nuScenes        & \scriptsize HDL32E & \scriptsize spinning & 34149 & 0.02 & 0.033 & 0.008 & 1.84 & -1.54  \\
        \hline
        \scriptsize KITTI-360       & \scriptsize HDL64E & \scriptsize spinning & 64640 & 0.02 & 0.033 & 0.009 & 1.73 & -1.43  \\
        \hline
        \multirow{6}{*}{\begin{tabular}[c]{@{}l@{}}{\scriptsize LiDARDustX}\end{tabular}}
                                    & \scriptsize LS128S2 & \scriptsize solid-state & 629 & 0.03 & 0.020 & 0.009 & 1.35 & -1.05  \\
                                    & \scriptsize CB64S1 & \scriptsize solid-state & 1116 & 0.03 & 0.063 & 0.012 & 1.4 & -1.1  \\
                                    & \scriptsize FalconK1 & \scriptsize solid-state & 410 & 0.02 & 0.010 & 0.010 & 2.5 & -2.2  \\
                                    & \scriptsize FalconK3 & \scriptsize solid-state & 671 & 0.02 & 0.010 & 0.007 & 2.6 & -2.3  \\
                                    & \scriptsize RS-M1 & \scriptsize solid-state & 3112 & 0.025 & 0.010 & 0.010 & 0.8 & -0.5  \\
                                    & \scriptsize Ouster2 & \scriptsize spinning & 1624 & 0.02 & 0.010 & 0.010 & 1.8 & -1.5  \\
        \hline
        \multirow{2}{*}{\begin{tabular}[c]{@{}l@{}}{\scriptsize self-collected}\end{tabular}}
                                    & \scriptsize VLP32C & \scriptsize spinning & -- & 0.03 & 0.033 & 0.010 & 0.7 & -0.4  \\
                                    & \scriptsize OS1-128 & \scriptsize spinning & -- & 0.03 & 0.010 & 0.010 & 0.5 & -0.2  \\
        \bottomrule
        \end{tabular}
\end{center}
\end{table}
\setlength{\tabcolsep}{6pt}

To ensure cross-platform software compatibility for both onboard and offboard deployments, we implement FugSeg in C++ and test it on an Ubuntu 18.04 desktop equipped with 32 GB of RAM, an Intel i7-9700K CPU and an RTX4000 GPU.
\subsection{Datasets}
For a comprehensive evaluation, we conduct experiments on four public datasets: SemanticKITTI\cite{geiger2012cvpr,behley2019iccv}, nuScenes\cite{nuscenes}, KITTI-360\cite{9786676} and LiDARDustX\cite{11127917}. The first three datasets cover a large variety of urban and highway driving scenarios with diverse lighting and weather conditions, whereas the LiDARDustX dataset focuses on unstructured environments such as mines and sand quarries. To derive ground labels for these datasets, their existing semantic annotations are utilized. For SemanticKITTI, we follow the same convention as\cite{9981561} and\cite{10319084}: $\{$\textit{road}, \textit{sidewalk}, \textit{other-ground}, \textit{terrain}, \textit{parking}, \textit{lane-marking}$\}$ compose the ground labels, $\{$\textit{unlabeled}, \textit{outlier}, \textit{vegetation}$\}$ are excluded from numerical evaluations, and all other labels are considered non-ground. Similarly, the ground and invalid sets are defined as $\{$\textit{drivable surface}, \textit{sidewalk}, \textit{terrain}, \textit{flat.other}$\}$ and $\{$\textit{noise}$\}$ for nuScenes, $\{$\textit{ground}, \textit{road}, \textit{sidewalk}, \textit{parking}, \textit{rail track}, \textit{terrain}$\}$ and $\{$\textit{unlabeled}, \textit{outlier}, \textit{ego vehicle}$\}$ for KITTI-360, and $\{$\textit{ground}, \textit{grass}$\}$ and $\{$\textit{ignore}, \textit{unknown}$\}$ for LiDARDustX. In addition to public datasets, we also validate FugSeg using self-collected LiDAR measurements. As summarized in Table~\ref{tab7}, 11 LiDAR sensors are evaluated in total, covering both spinning and solid-state units with varying resolutions and fields of view.

\subsection{Evaluation Metrics}
In this work, the following evaluation metrics are employed: precision, recall, $\mathrm{F_1}$, accuracy and mIoU (mean intersection over union). They are defined as follows:
\begin{equation}
    \begin{gathered}
    \label{equ.4.b.1}
        \text{precision} = \frac{TP}{TP+FP},\quad \text{recall} = \frac{TP}{TP+FN}, \\
        \mathrm{F_1} = \frac{2\cdot TP}{2\cdot TP+FP+FN}, \\
        \text{accuracy} = \frac{TP+TN}{TP+FP+TN+FN}, \\
        \text{mIoU} = \frac{1}{2}\cdot (\frac{TP}{TP+FP+FN}+\frac{TN}{TN+FP+FN})
    \end{gathered}
\end{equation}
where $TP$, $FP$, $TN$, and $FN$ represent the number of true positive, false positive, true negative, and false negative ground points, respectively. Additionally, the average runtime is reported to quantify the efficiency of a method, which is given in milliseconds per scan.

\subsection{Parameter Setup}
\begin{figure}[!t]
    \centering
    \includegraphics[width=3.5in]{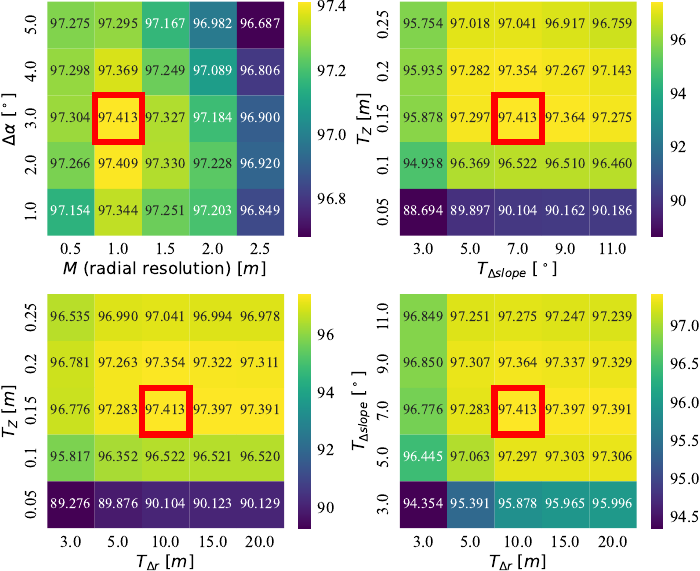}
    \caption{$\mathrm{F_1}$-based grid search for algorithmic parameters $\Delta\alpha$, $M$, $T_{\Delta r}$, $T_Z$ and $T_{\Delta slope}$ on sequence 08 of the SemanticKITTI dataset. The optimal configurations are highlighted in red rectangles.}
    \label{fig11}
\end{figure}
\begin{table}
\begin{center}
\caption{Algorithmic Parameter Setup.}
\label{tab1}
    \begin{tabular}{*{8}c}
    \toprule
    \textbf{Parameter} & $\Delta\alpha$ & $M$ & $r_0$ & $r_M$ & $T_{\Delta slope}$ & $T_{\Delta r}$ & $T_Z$  \\
    \midrule
    \textbf{Value} & $\ang{3}$ & $80$ & $0.5m$ & $80m$ & $\tan(\ang{7})$ & $10m$ & $0.15m$   \\
    \bottomrule
    \end{tabular}
\end{center}
\end{table}

To enhance the scalability and generalizability of FugSeg, a key design objective is minimizing the number of parameters. Consequently, FugSeg comprises twelve parameters in total, including five sensor-specific and seven algorithmic parameters. Sensor-specific parameters include the measurement accuracies $\sigma_R$, $\sigma_\phi$ and $\sigma_\theta$, and LiDAR's installation height $H_s$ (thus also $T_h$). The first three parameters are typically documented in the corresponding datasheet or can be empirically estimated through sensor calibration\cite{5651382,rs2061610,Levinson2014}. In this study, these three parameters are obtained from the corresponding sensor datasheets. Parameters $H_s$ and $T_h$ depend on the actual hardware setup. Table~\ref{tab7} lists the sensor-specific parameters of each evaluated LiDAR.

Algorithmic parameters consist of the range parameters $r_0$ and $r_M$, the polar grid's spatial resolutions $\Delta\alpha$ and $M$, and three thresholds $T_{\Delta slope}$, $T_{\Delta r}$ and $T_Z$. Parameters $r_0$ and $r_M$ indicate the valid measurement range of a LiDAR sensor, which can be statistically determined from the corresponding point clouds. For the remaining five algorithmic parameters, however, due to the diversity of input point clouds and the complex interactions among individual components, a fully theoretical derivation is non-trivial. Therefore, as presented in Figure~\ref{fig11}, to optimize $\Delta\alpha$, $M$, $T_{\Delta slope}$, $T_{\Delta r}$ and $T_Z$, we conduct a systematic grid search on sequence 08 (containing 4071 scans) of the SemanticKITTI dataset. During the grid search, the following parameter ranges are examined:
\begin{itemize}
    \item{$\Delta\alpha$: from $\ang{1}$ to $\ang{5}$ in $\ang{1}$ steps, balancing the azimuth resolution of the evaluated sensors and the level of detail in the resulting polar grid map\cite{5548059,rs13163239,9548034,9466396}.}
    \item{$M$: with the corresponding radial resolution ranging from $0.5m$ to $2.5m$ in $0.5m$ increments, covering the vertical resolution of both the sparse and high-definition LiDAR sensors\cite{5548059,rs13163239,9548034,9466396,geiger2012cvpr,nuscenes}.}
    \item{$T_{\Delta slope}$: $\{\tan(\ang{3}), \tan(\ang{5}), \tan(\ang{7}), \tan(\ang{9}), \tan(\ang{11})\}$, covering typical slope changes in both urban and non-urban environments\cite{rs13163239,9548034,10359455,bonn2017}.}
    \item{$T_{\Delta r}$: $\{3m, 5m, 10m, 15m, 20m\}$, covering typical measurement gaps in the radial direction, especially in sparse and occluded point clouds\cite{nuscenes,behley2019iccv}.}
    \item{$T_Z$: from $0.05m$ to $0.25m$ with a step size of $0.05m$, encompassing common elevation deviations of traversable ground\cite{rs13163239,9548034,9466396,9794594}.}
\end{itemize}
This results in a total of 3125 parameter combinations. The optimal configuration, as shown in Table~\ref{tab1}, is selected based on the highest average $\mathrm{F_1}$ score achieved across all 4071 scans.

\section{Results and Discussion}

\subsection{Comparison with State-of-the-Art Methods}
For a comprehensive evaluation, we compare FugSeg with other state-of-the-art approaches, including RANSAC\cite{358692}, GPF\cite{7989591}, Patchwork++\cite{9981561}, TRAVEL\cite{9794594}, LineFit\cite{5548059}, DepthClustering\cite{bonn2017}, JCP\cite{rs13163239}, GroundGrid\cite{10319084}, DipG-Seg\cite{10359455}, 2DPASS\cite{yan20222dpass}, SphereFormer\cite{10203552} and LSK3DNet\cite{10656196}. To ensure a fair comparison, the following measures were implemented: (1) the RANSAC method was implemented and optimized following the same strategy as FugSeg, and all other methods were evaluated based on the corresponding open-source implementations, with default parameter settings as documented in the respective publications; (2) learning-based methods were evaluated on the GPU, whereas all other methods were executed single-threaded on the CPU of the aforementioned hardware platform; (3) GroundGrid was adapted as a frame-wise segmentation method (by removing its odometry-based tracking); and (4) laser reflections from the ego-vehicle were directly labeled as non-ground using the bounding boxes defined in\cite{10359455}, as some of these methods require ground seeds in the immediate vicinity.

\subsubsection{Quantitative Comparison}
\begin{table*}
\footnotesize
\setlength{\tabcolsep}{2.5pt}
\begin{center}
\caption{Quantitative Comparison on SemanticKITTI, nuScenes and KITTI-360. Runtime in [$ms$], other metrics in [\%]. Type I: iterative, D: deterministic, L: learning based. $^*$ Performance evaluated on the corresponding validation set.}
\label{tab3}
        \begin{tabular}{*{7}c r *{5}c r *{5}c r}
        \toprule
        \multirow[b]{2}{*}{\textbf{Method}} & \multirow[b]{2}{*}{\textbf{Type}} & \multicolumn{6}{c}{{SemanticKITTI}} & \multicolumn{6}{c}{{nuScenes}} & \multicolumn{6}{c}{{KITTI-360}}  \\
        \cmidrule(lr){3-8}  \cmidrule(lr){9-14}  \cmidrule(lr){15-20}
        & & \textbf{P} & \textbf{R} & {$\mathbf{F_1}$} & \textbf{A} & \textbf{mIoU} & \multicolumn{1}{c}{\textbf{t}} & \textbf{P} & \textbf{R} & {$\mathbf{F_1}$} & \textbf{A} & \textbf{mIoU} & \multicolumn{1}{c}{\textbf{t}} & \textbf{P} & \textbf{R} & {$\mathbf{F_1}$} & \textbf{A} & \textbf{mIoU} & \multicolumn{1}{c}{\textbf{t}}  \\
        \midrule
        \scriptsize RANSAC-50\cite{358692}
                                       &  I  & 90.52 & 90.43 & 90.48 & 90.14 & 80.36 & 12.091
                                             & 92.66 & 94.21 & 93.43 & 94.97 & 90.12 & 2.156
                                             & 82.75 & 85.30 & 84.00 & 87.19 & 77.15 & 14.315  \\
        \scriptsize RANSAC-300\cite{358692}
                                       &  I  & 91.47 & 93.07 & 92.27 & 91.69 & 82.74 & 62.419
                                             & 92.68 & 94.84 & 93.74 & 95.16 & 90.49 & 11.102
                                             & 83.79 & 87.98 & 85.83 & 88.38 & 79.04 & 58.109  \\
        GPF-3\cite{7989591}            &  I  & 94.98 & 73.60 & 82.94 & 80.30 & 68.86 & 22.249
                                             & 94.09 & 74.70 & 83.28 & 88.74 & 78.67 & 3.821
                                             & 89.88 & 73.13 & 80.64 & 84.31 & 72.46 & 23.547  \\
        GPF-50\cite{7989591}           &  I  & 94.74 & 87.29 & 90.86 & 88.86 & 79.29 & 94.646
                                             & 94.72 & 85.71 & 89.99 & 92.76 & 85.94 & 17.698
                                             & 89.30 & 84.70 & 86.94 & 88.63 & 79.14 & 94.230  \\
        Patchwork++\cite{9981561}      &  I  & 93.16 & \textbf{98.32} & 95.67 & 95.02 & 88.36 & 27.244
                                             & 91.70 & 95.49 & 93.56 & 94.98 & 89.90 & 6.362
                                             & 84.69 & \textbf{94.45} & 89.30 & 90.06 & 81.20 & 27.846  \\
        TRAVEL\cite{9794594}           &  I  & 95.79 & 96.12 & 95.95 & 95.17 & 88.62 & 17.245
                                             & 94.94 & 93.60 & 94.26 & 95.51 & 91.02 & 3.736
                                             & 90.44 & 90.55 & 90.50 & 91.50 & 83.56 & 17.145  \\
        LineFit\cite{5548059}          &  D  & 94.01 & 96.68 & 95.33 & 94.43 & 87.01 & 31.912
                                             & 94.54 & 92.39 & 93.45 & 94.94 & 89.87 & 11.370
                                             & 86.92 & 93.16 & 89.93 & 90.74 & 82.31 & 30.346  \\
        \scriptsize DepthClustering\cite{bonn2017}
                                       &  D  & \textbf{99.34} & 77.39 & 87.00 & 85.66 & 73.14 & 11.679
                                             & \textbf{98.26} & 77.36 & 86.57 & 91.05 & 81.76 & 3.601
                                             & \textbf{97.66} & 69.48 & 81.19 & 85.10 & 72.86 & 13.903  \\
        JCP\cite{rs13163239}           &  D  & 93.56 & 96.47 & 94.99 & 94.15 & 86.73 & 22.218
                                             & 93.70 & 94.02 & 93.86 & 95.24 & 90.47 & 6.249
                                             & 89.55 & 91.12 & 90.32 & 91.41 & 83.38 & 23.394  \\
        GroundGrid\cite{10319084}      &  D  & 95.66 & 94.82 & 95.24 & 94.34 & 87.20 & 58.987
                                             & 87.37 & \textbf{96.60} & 91.75 & 93.28 & 86.91 & 50.407
                                             & 90.51 & 88.45 & 89.47 & 90.57 & 82.08 & 58.218  \\
        DipG-Seg\cite{10359455}        &  D  & 98.21 & 92.90 & 95.48 & 94.77 & 88.02 & 15.392
                                             & 96.62 & 93.59 & 95.08 & 96.38 & 92.48 & 4.650
                                             & 94.82 & 87.59 & 91.06 & 92.22 & 84.78 & 17.852  \\
        \hline
        \textbf{FugSeg (Ours)}         &  D  & 95.98 & 97.72 & \textbf{96.84} & \textbf{96.27} & \textbf{90.90} & \textbf{7.385}
                                             & 95.05 & 96.18 & \textbf{95.61} & \textbf{96.52} & \textbf{92.97}  & \textbf{2.053}
                                             & 90.85 & 92.33 & \textbf{91.58} & \textbf{92.41} & \textbf{85.17} & \textbf{7.372}  \\
        \hline
        \hline
        2DPASS\cite{yan20222dpass}$^*$ &  L  & 99.38 & 98.61 & 98.99 & 98.88 & 97.12 & 43.20
                                             & \textbf{98.89} & 97.04 & 97.92 & 97.78 & 95.40 & 41.34
                                             & \textbf{97.51} & 93.51 & 95.41 & 95.07 & 88.20 & 36.95  \\
        \scriptsize SphereFormer\cite{10203552}$^*$
                                       &  L  & \textbf{99.65} & \textbf{99.18} & \textbf{99.41} & \textbf{99.33} & \textbf{98.26} & 250.99
                                             & 98.69 & \textbf{98.44} & \textbf{98.55} & \textbf{98.44} & \textbf{96.77} & 67.50
                                             & 97.29 & \textbf{94.06} & \textbf{95.59} & \textbf{95.22} & \textbf{88.58} & 230.96  \\
        LSK3DNet\cite{10656196}$^*$    &  L  & 99.24 & 98.58 & 98.90 & 98.76 & 96.87 & 306.45
                                             & 98.73 & 97.37 & 98.01 & 97.89 & 95.62 & 322.93
                                             & 96.61 & 93.61 & 95.01 & 94.53 & 86.98 & 282.75  \\
        \textbf{FugSeg (Ours)}$^*$     &  D  & 97.17 & 97.66 & 97.41 & 97.00 & 92.78 & \textbf{7.60}
                                             & 95.44 & 96.02 & 95.73 & 96.54 & 93.07 & \textbf{2.14}
                                             & 90.04 & 91.72 & 90.87 & 91.98 & 84.38 & \textbf{7.62}  \\
        \bottomrule
        \end{tabular}
\end{center}
\end{table*}
\setlength{\tabcolsep}{6pt}

\begin{table}[!t]
\footnotesize
\setlength{\tabcolsep}{3.0pt}
\begin{center}
\caption{Quantitative Comparison on the LiDARDustX dataset.}
\label{tab6}
        \begin{tabular}{*{6}c r}
        \toprule
        \textbf{Method} & \textbf{P} & \textbf{R} & {$\mathbf{F_1}$} & \textbf{A} & \textbf{mIoU} & \multicolumn{1}{c}{\textbf{t}}  \\
        \midrule
        RANSAC-50\cite{358692}         & 80.18 & 82.10 & 81.13 & 81.72 & 67.94 & 3.787  \\
        RANSAC-300\cite{358692}        & 80.20 & 83.00 & 81.58 & 81.89 & 68.22 & 19.451  \\
        GPF-3\cite{7989591}            & 85.22 & 82.43 & 83.80 & 83.67 & 69.74 & 8.203  \\
        GPF-50\cite{7989591}           & 83.36 & 85.45 & 84.39 & 83.73 & 70.32 & 35.826  \\
        Patchwork++\cite{9981561}      & 76.44 & \textbf{93.73} & 84.21 & 82.25 & 64.78 & 10.407  \\
        TRAVEL\cite{9794594}           & \textbf{91.10} & 92.58 & 91.83 & 92.91 & 82.68 & 6.325  \\
        LineFit\cite{5548059}          & 80.23 & 73.95 & 76.96 & 79.78 & 64.05 & 13.370  \\
        GroundGrid\cite{10319084}      & 87.64 & 91.42 & 89.49 & 89.12 & 76.84 & 50.410  \\
        \hline
        \textbf{FugSeg (Ours)}         & 90.56 & 93.61 & \textbf{92.06} & \textbf{93.18} & \textbf{83.12} & \textbf{2.690}  \\
        \bottomrule
        \end{tabular}
\end{center}
\end{table}
\setlength{\tabcolsep}{6pt}

\begin{table}[!t]
\footnotesize
\setlength{\tabcolsep}{3.0pt}
\centering
\caption{Performance of FugSeg under various nuScenes Conditions.}
\label{tab5}
\begin{tabular}{llrcccccr}
\toprule
{\scriptsize \textbf{Location}}  & {\scriptsize \textbf{Condition}} & \textbf{\#scan} & \textbf{P} & \textbf{R} & $\mathbf{F_1}$ & \textbf{A} & \textbf{mIoU} & \multicolumn{1}{c}{\textbf{t}} \\
\midrule
\multirow{2}{*}{\begin{tabular}[c]{@{}l@{}}{\scriptsize Boston}\\[-0.2em]{\scriptsize Seaport}\end{tabular}}
                    &  rain    & 6028  & 95.43 & 98.12 & 96.76 & 97.90 & 95.30 & 1.815  \\
                    &  no rain & 12757 & 95.54 & 98.39 & 96.95 & 97.72 & 95.20 & 2.006  \\
\hline
\multirow{4}{*}{\begin{tabular}[c]{@{}l@{}}{\scriptsize Singapore}\\[-0.2em]{\scriptsize Holland}\\[-0.2em]{\scriptsize Village}\end{tabular}}
                    &  rain    & 642   & 95.15 & 95.71 & 95.43 & 96.42 & 92.73 & 2.198  \\
                    &  no rain & 2785  & 93.76 & 95.41 & 94.58 & 96.02 & 91.74 & 2.205  \\
                    \cline{2-9}
                    &  night   & 2657  & 94.28 & 95.43 & 94.85 & 96.20 & 92.12 & 2.190  \\
                    &  day     & 770   & 93.12 & 95.58 & 94.34 & 95.74 & 91.26 & 2.250  \\
\hline
{\scriptsize One-north}
                    &  all     & 7308  & 95.21 & 94.79 & 95.00 & 95.59 & 91.48 & 2.163  \\
\hline
\multirow{2}{*}{\begin{tabular}[c]{@{}l@{}}{\scriptsize Singapore}\\[-0.2em]{\scriptsize Queenstown}\end{tabular}}
                    &  night   & 1330  & 92.52 & 88.49 & 90.46 & 92.32 & 85.11 & 2.072  \\
                    &  day     & 3299  & 94.12 & 90.98 & 92.52 & 93.53 & 87.67 & 2.258  \\
\bottomrule
\end{tabular}
\end{table}
\setlength{\tabcolsep}{6pt}

As shown in Tables~\ref{tab3} and \ref{tab6}, among all non-learning methods, FugSeg achieves the best overall performance in terms of $\mathrm{F_1}$, accuracy and mIoU across all datasets, while also maintaining the fastest runtime. Except for sensor-specific parameters, all algorithmic parameters remain unchanged throughout all experiments, regardless of the evaluated dataset. This demonstrates FugSeg's compatibility across different sensors and environments (both structured and unstructured). Because the range-image representations used in JCP, DepthClustering and DipG-Seg are specific to spinning LiDARs, these methods are not evaluated on the LiDARDustX dataset. Compared with learning-based methods, FugSeg is markedly more efficient, though slightly inferior in overall performance. Nevertheless, even when accelerated by GPU hardware, SphereFormer and LSK3DNet still struggle to meet real-time requirements.

In general, methods such as DepthClustering, GPF and DipG-Seg achieve high precision but comparatively low recall, indicating potential under-segmentation. In contrast, Patchwork++ shows a tendency toward over-segmentation. GroundGrid and LineFit exhibit imbalanced performance across datasets, likely due to thresholds that are biased toward specific sensor models. Thanks to the proposed within- and cross-segment ground labeling, FugSeg achieves balanced precision and recall. For RANSAC and GPF, we additionally investigate the effect of the number of iterations on performance. As indicated by the suffixes, increasing the number of iterations generally improves performance, but at the cost of linearly increasing runtime, which can be challenging for onboard systems with limited computational resources.

To further validate FugSeg under varying weather and time conditions, a detailed analysis is conducted using the nuScenes dataset. As shown in Table~\ref{tab5}, FugSeg maintains consistent overall performance across diverse weather and time conditions at individual locations. However, the location-related performance variation suggests that localized parameter tuning may be needed in practice.

\subsubsection{Qualitative Comparison}
\setlength{\tabcolsep}{0.8pt}
\renewcommand{\arraystretch}{0.0}
\begin{figure*}[!t]
\begin{tabular}{cccc}
    &Scenario 1&Scenario 2&Scenario 3\\[.3em]
    \rotatebox{90}{\begin{minipage}{2.11cm}\centering{benchmark}\end{minipage}}&
    \includegraphics[width=.37\textwidth,height=2.11cm]{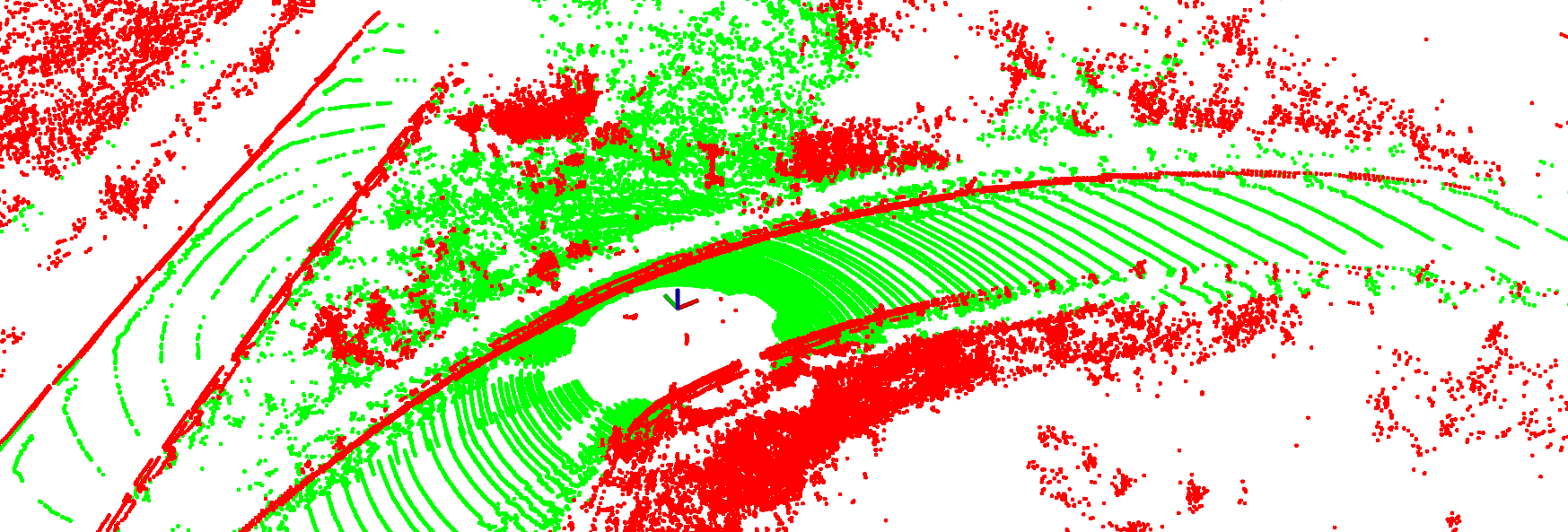}&\includegraphics[width=.27\textwidth,height=2.11cm]{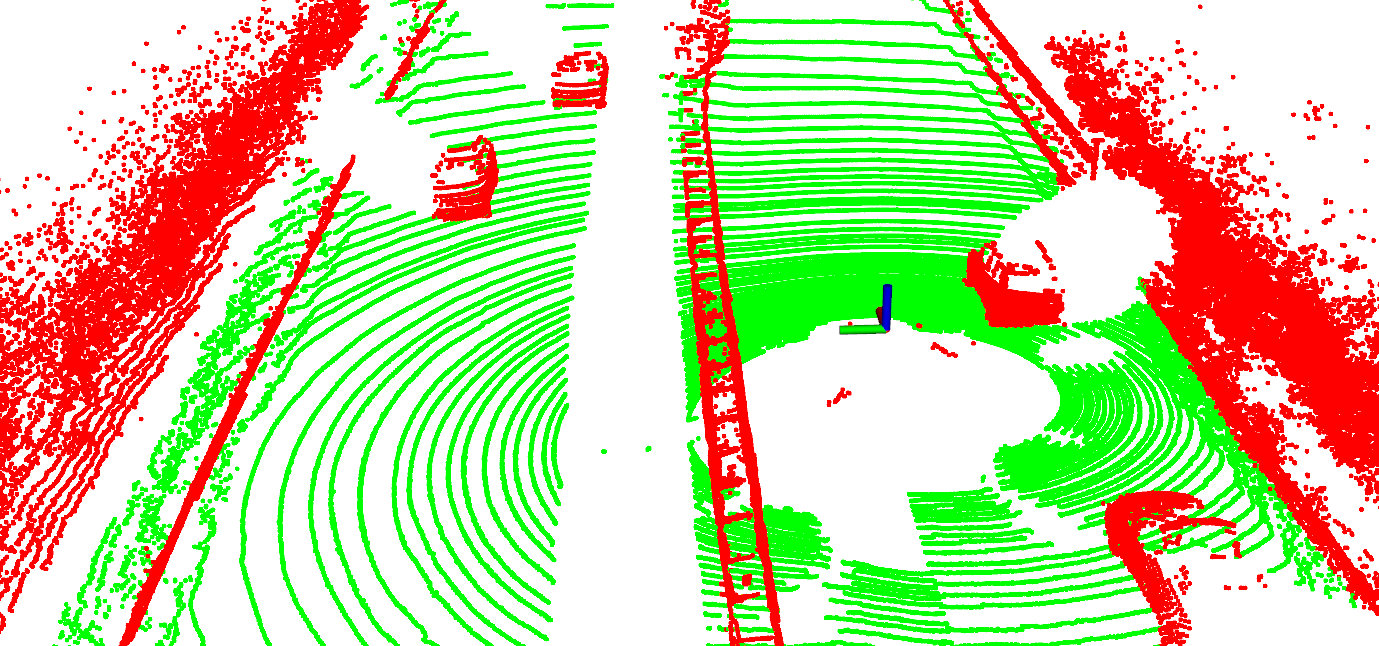}&\includegraphics[width=.33\textwidth,height=2.11cm]{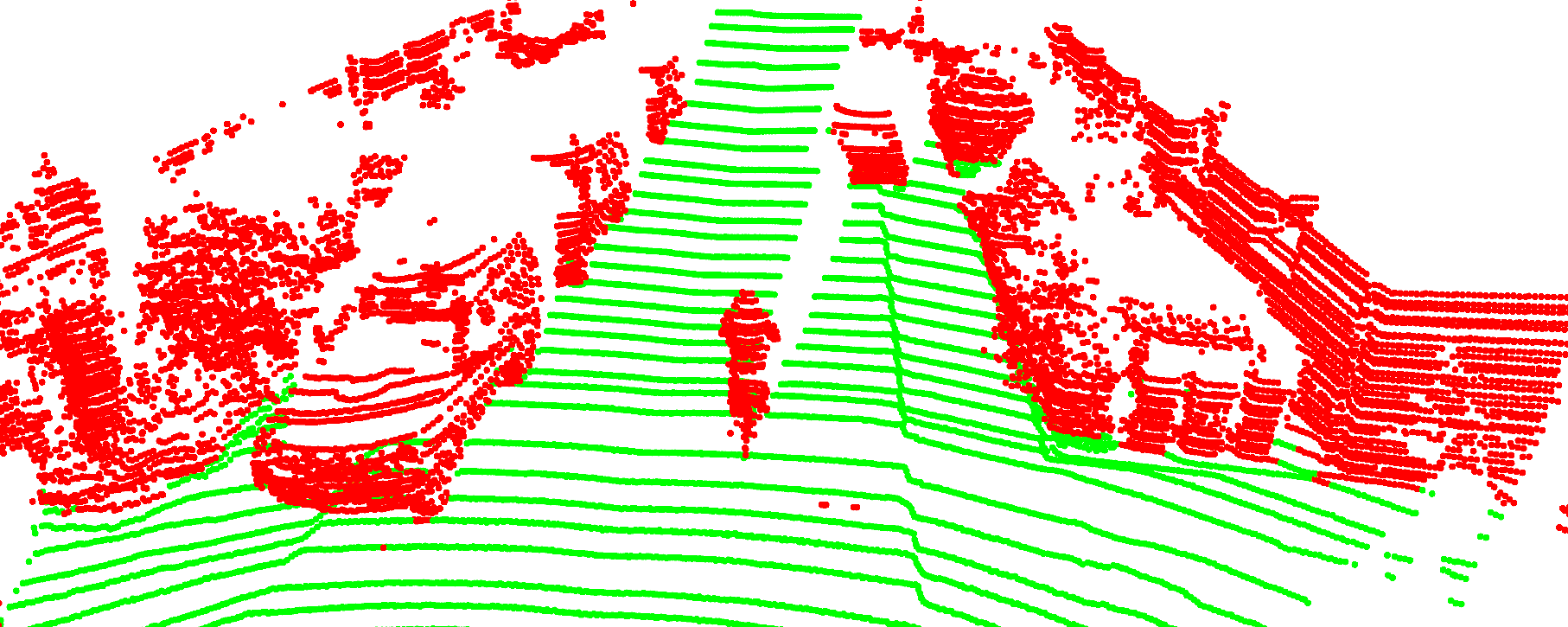}\\
    \rotatebox{90}{\begin{minipage}{2.11cm}\centering{RANSAC-300}\end{minipage}}&
    \includegraphics[width=.37\textwidth,height=2.11cm]{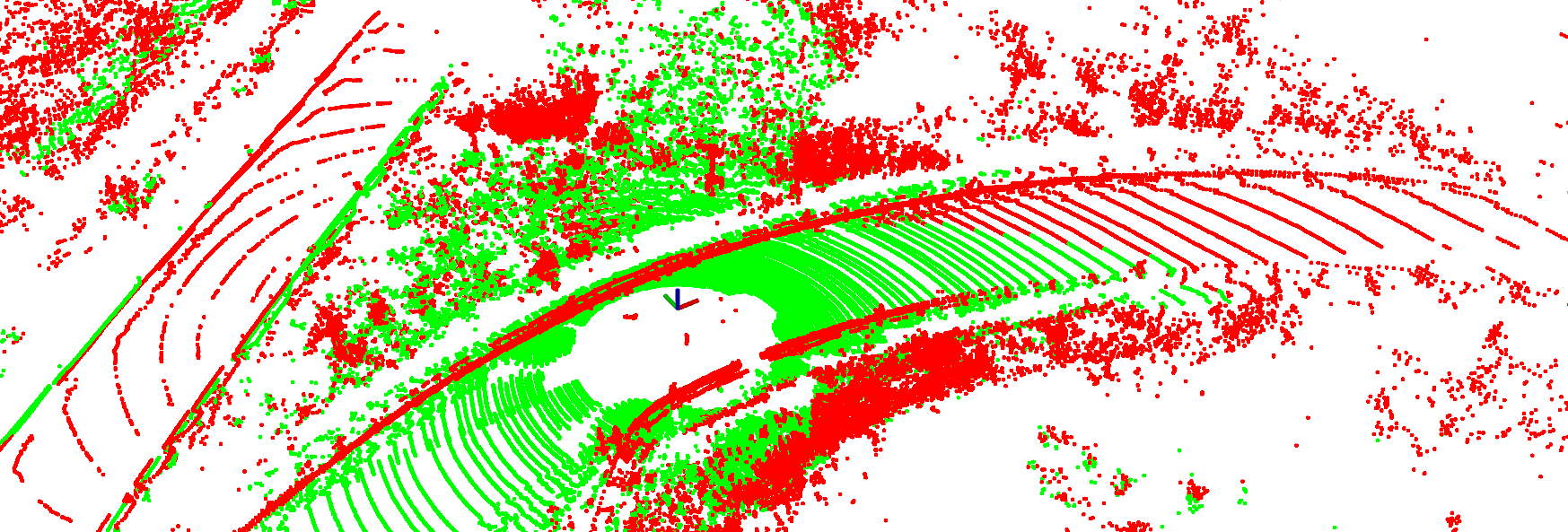}&\includegraphics[width=.27\textwidth,height=2.11cm]{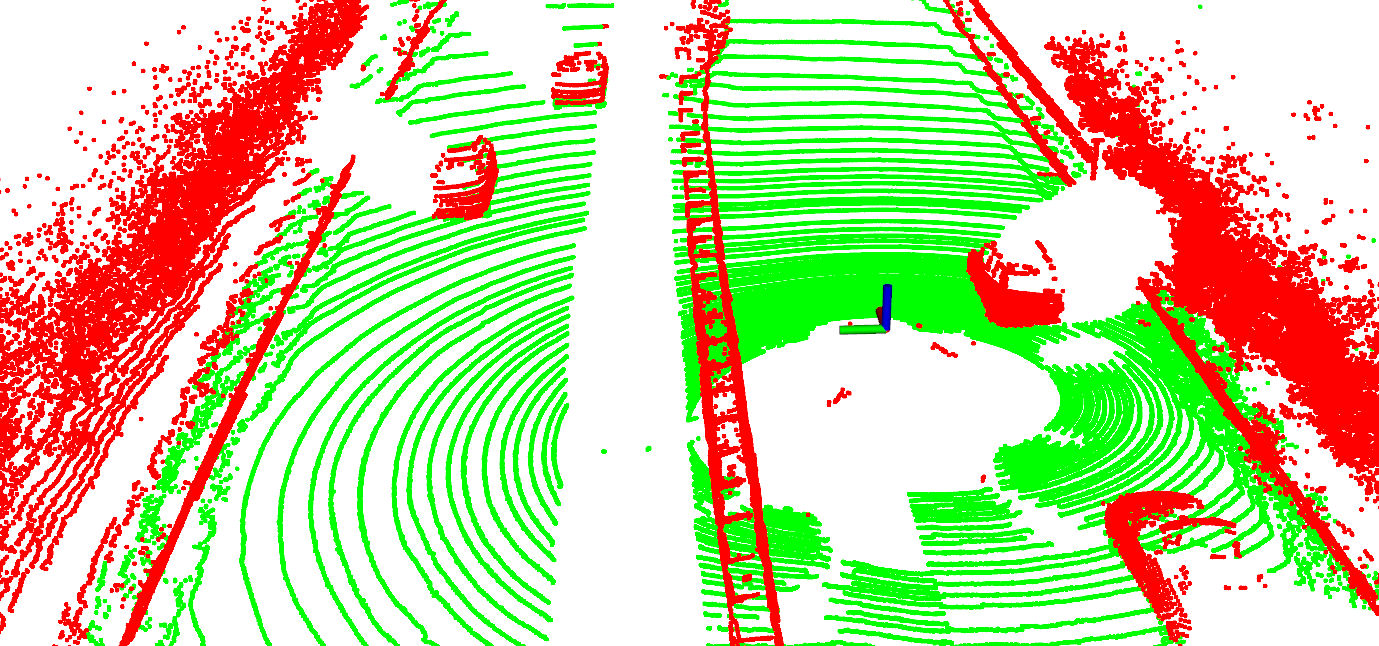}&\includegraphics[width=.33\textwidth,height=2.11cm]{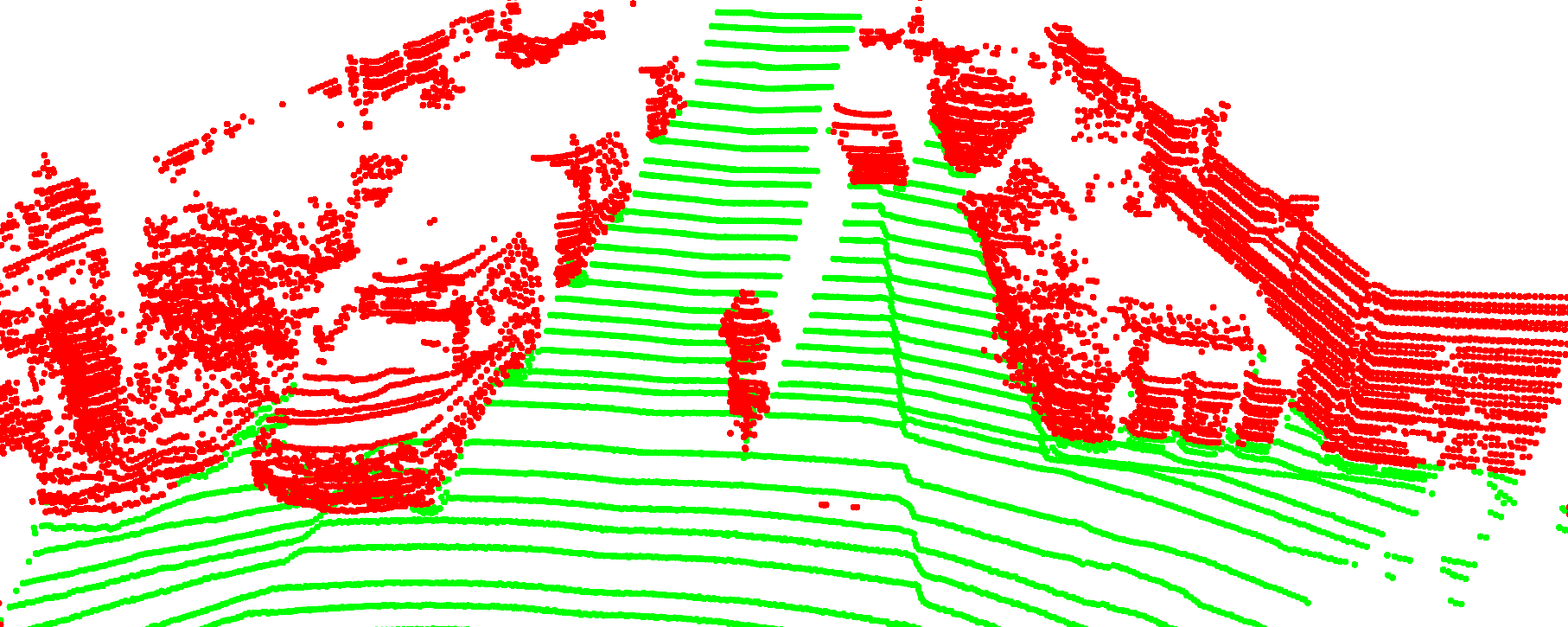}\\
    \rotatebox{90}{\begin{minipage}{2.11cm}\centering{GPF-50}\end{minipage}}&
    \includegraphics[width=.37\textwidth,height=2.11cm]{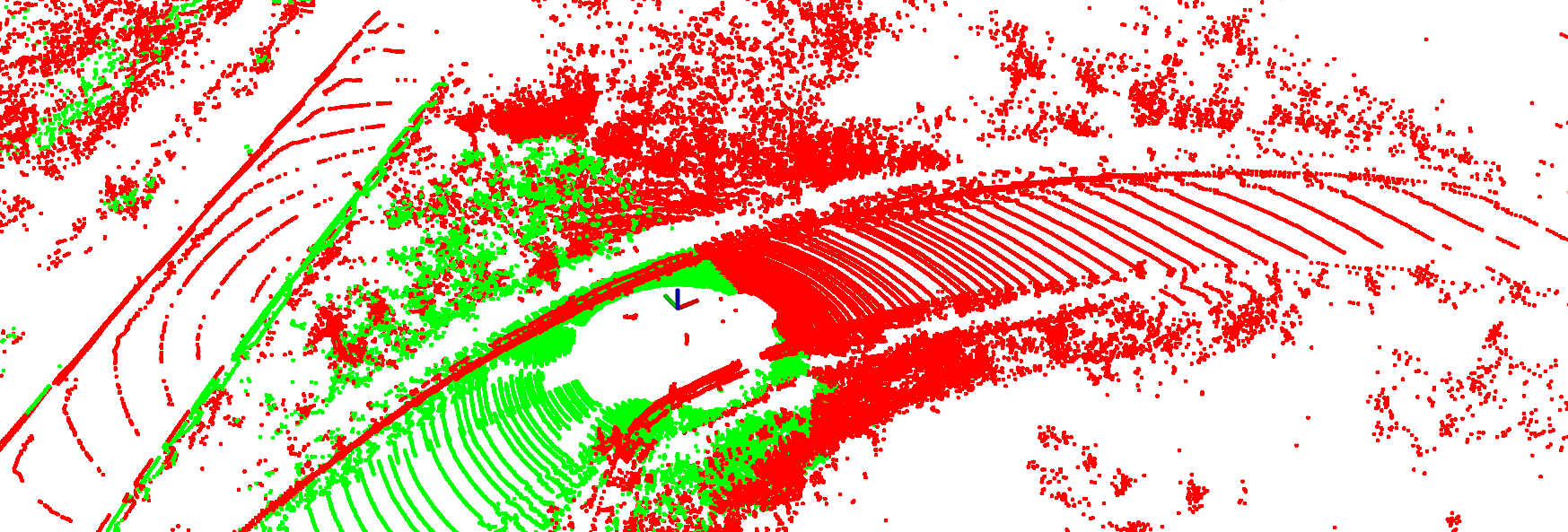}&\includegraphics[width=.27\textwidth,height=2.11cm]{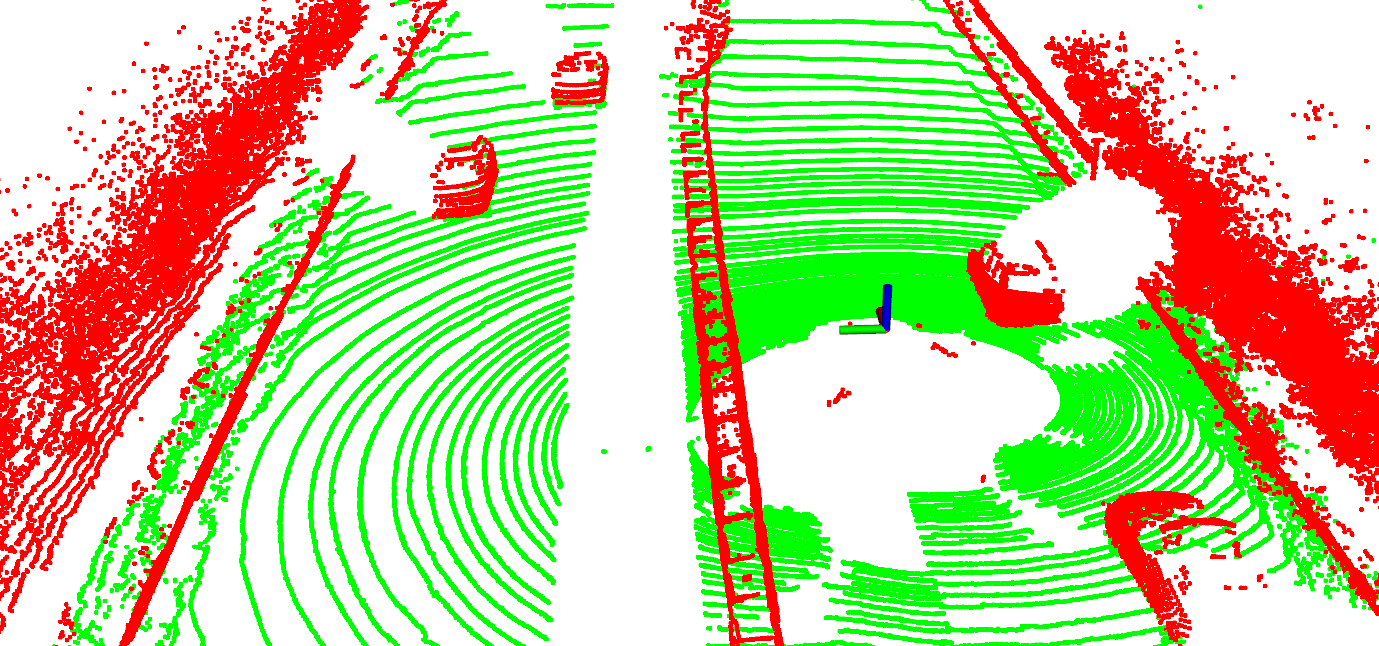}&\includegraphics[width=.33\textwidth,height=2.11cm]{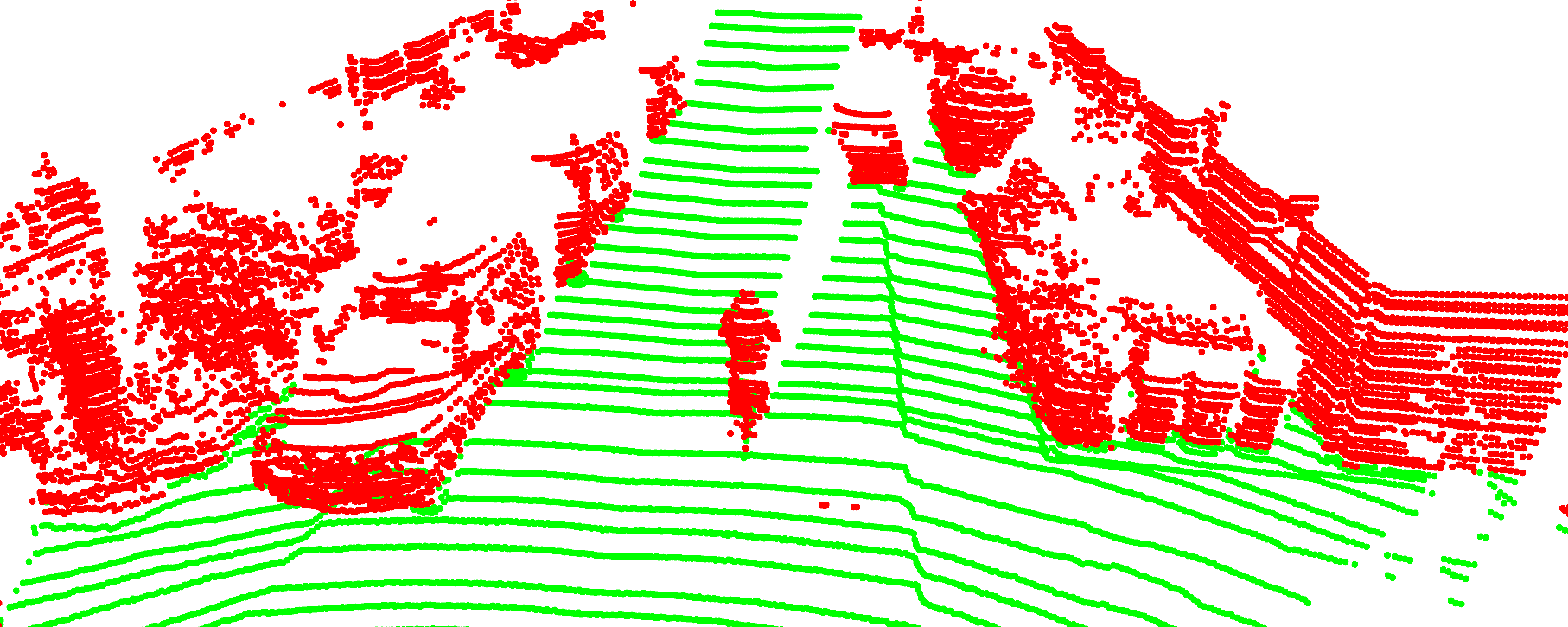}\\
    \rotatebox{90}{\begin{minipage}{2.11cm}\centering{Patchwork++}\end{minipage}}&
    \includegraphics[width=.37\textwidth,height=2.11cm]{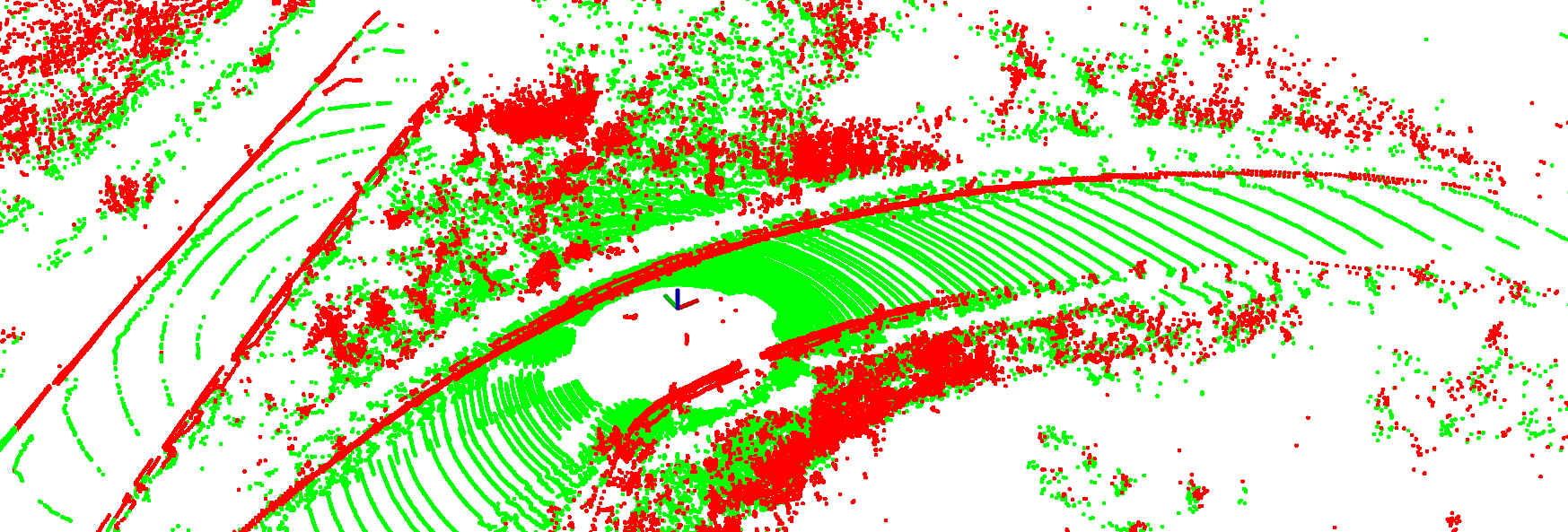}&\includegraphics[width=.27\textwidth,height=2.11cm]{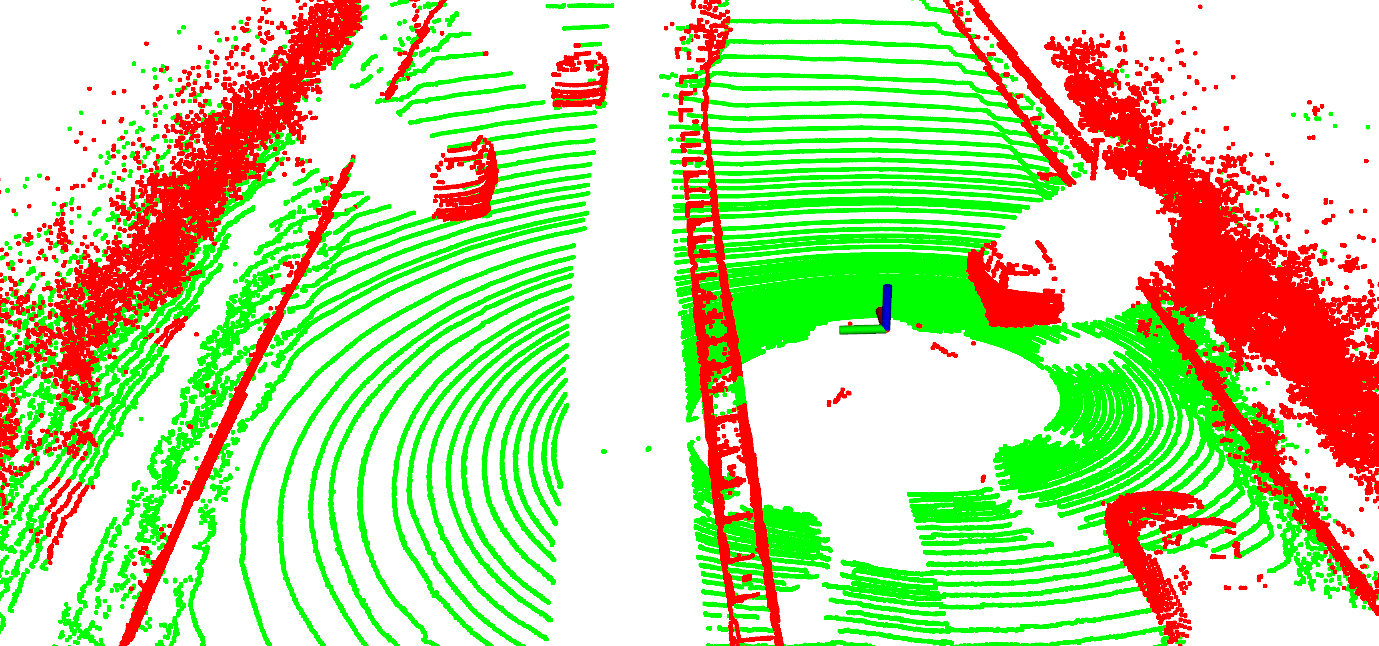}&\includegraphics[width=.33\textwidth,height=2.11cm]{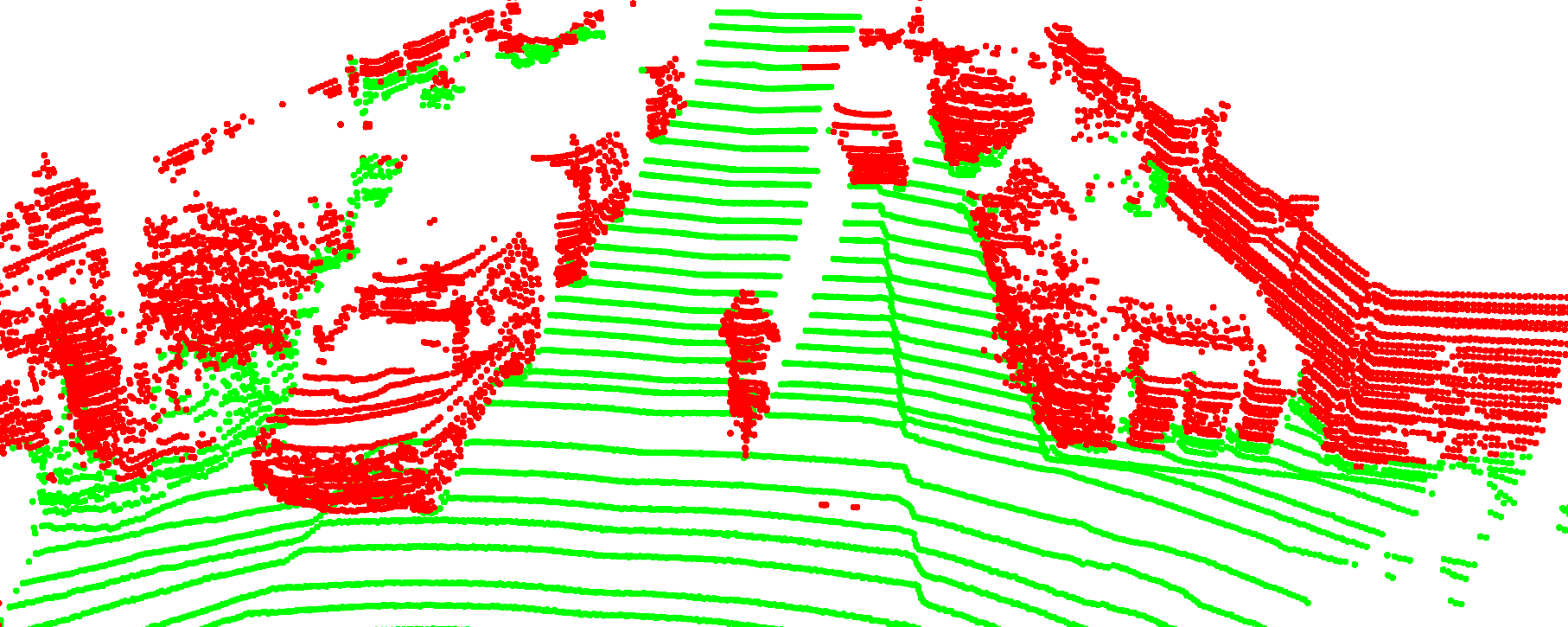}\\
    \rotatebox{90}{\begin{minipage}{2.11cm}\centering{TRAVEL}\end{minipage}}&
    \includegraphics[width=.37\textwidth,height=2.11cm]{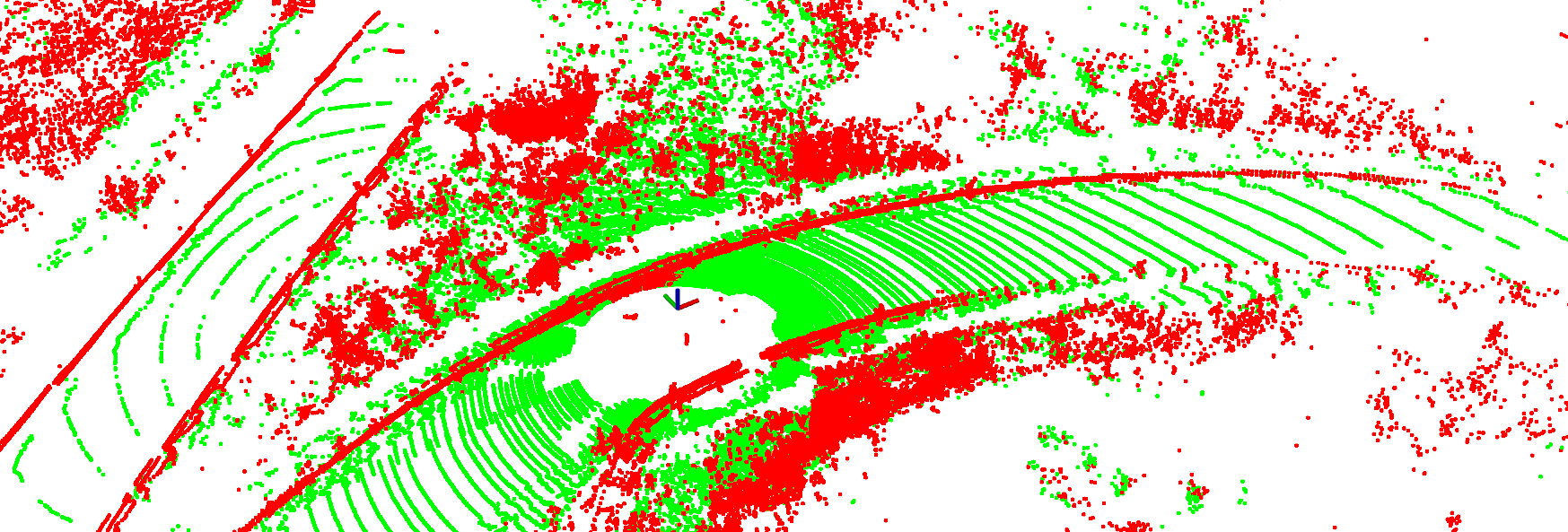}&\includegraphics[width=.27\textwidth,height=2.11cm]{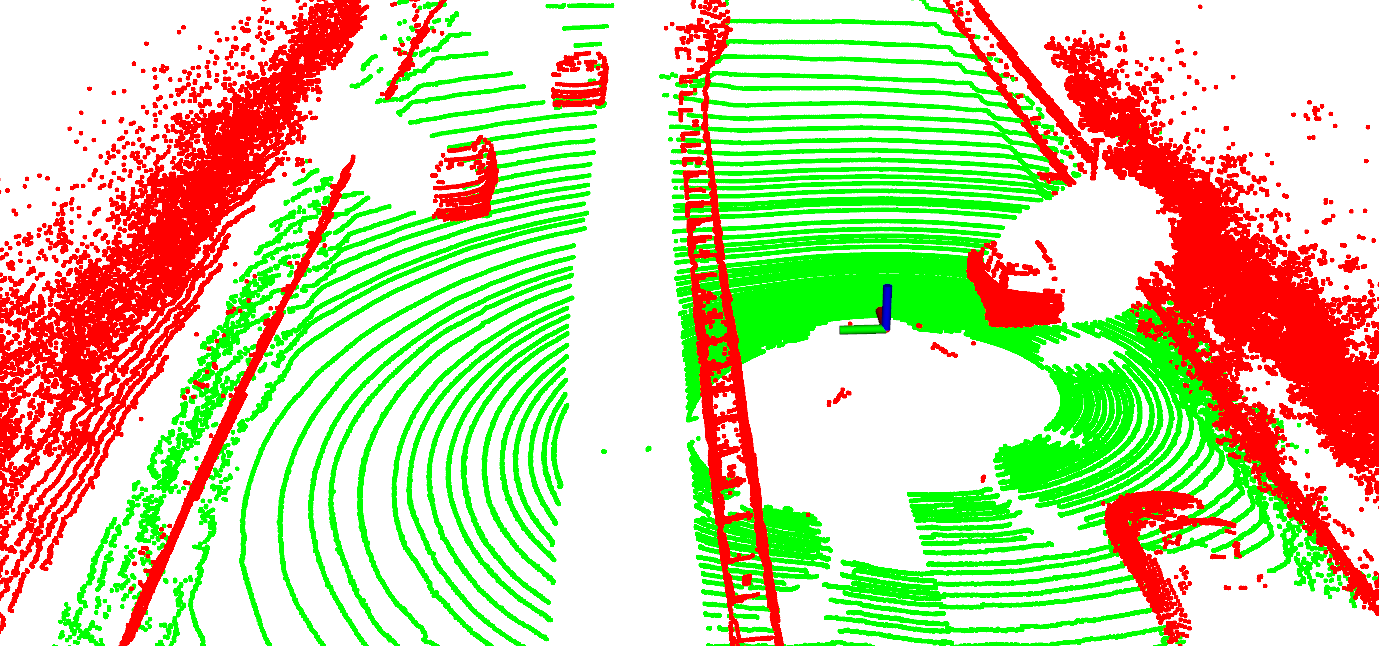}&\includegraphics[width=.33\textwidth,height=2.11cm]{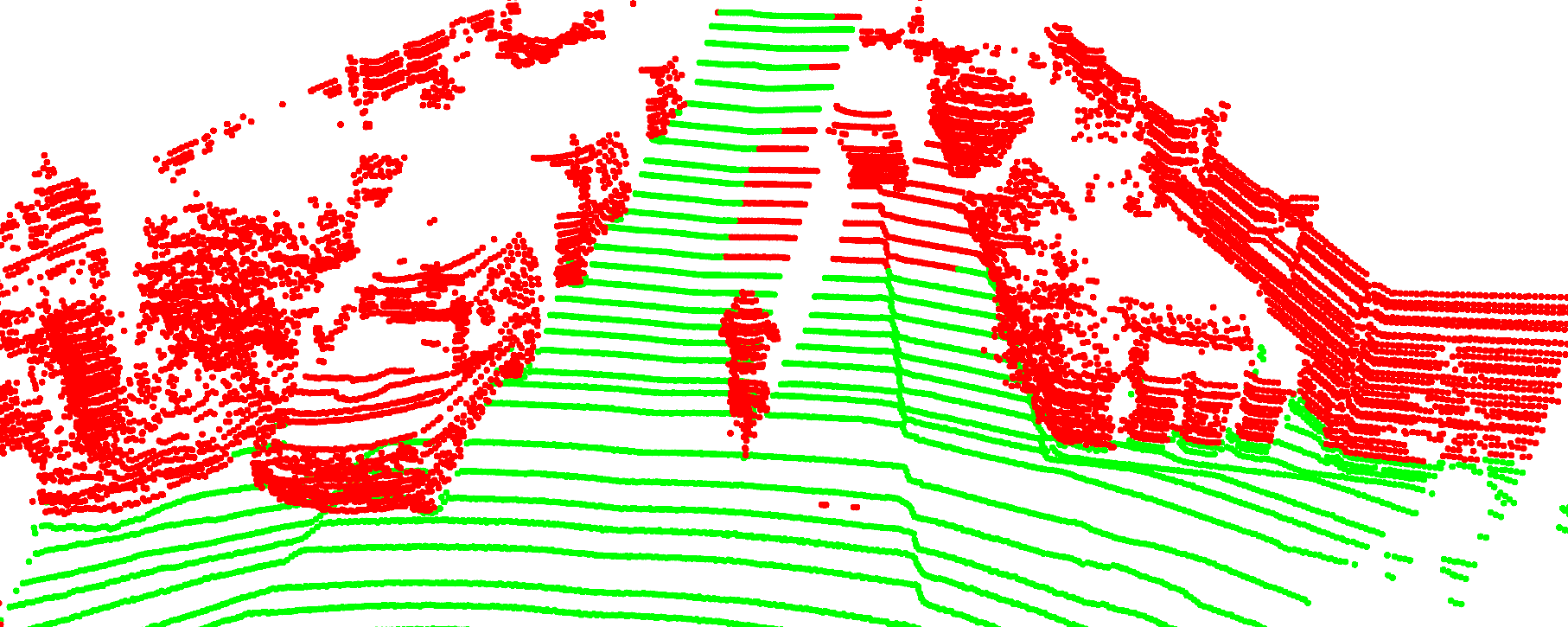}\\
    \rotatebox{90}{\begin{minipage}{2.11cm}\centering{LineFit}\end{minipage}}&
    \includegraphics[width=.37\textwidth,height=2.11cm]{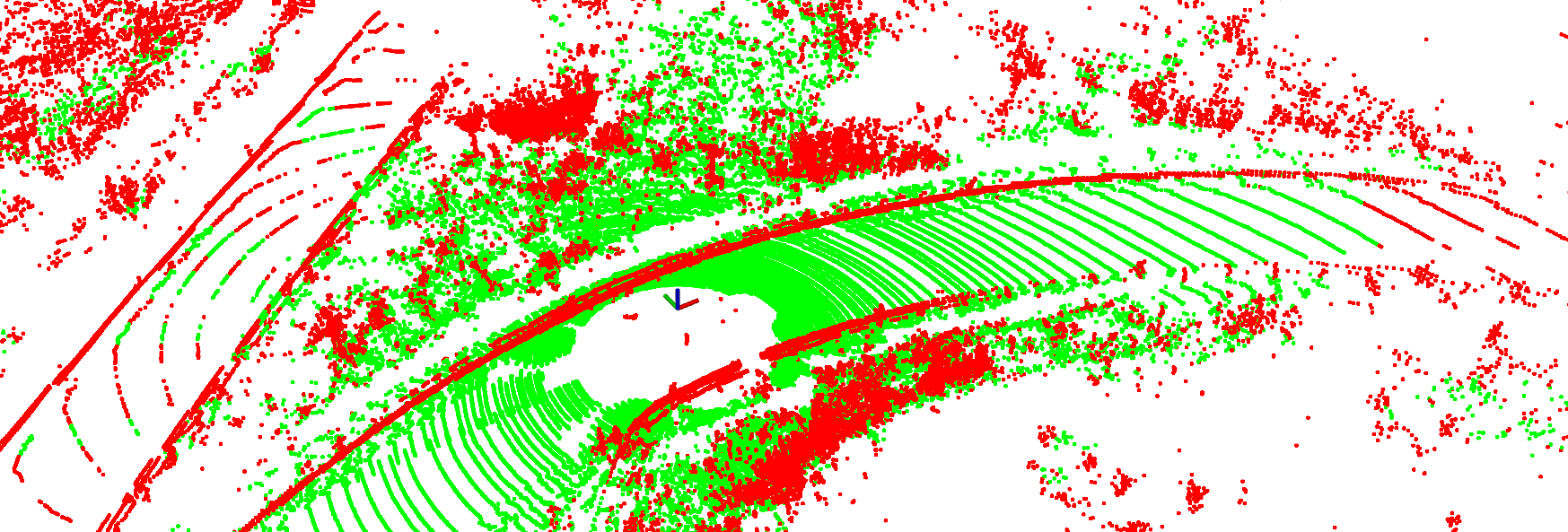}&\includegraphics[width=.27\textwidth,height=2.11cm]{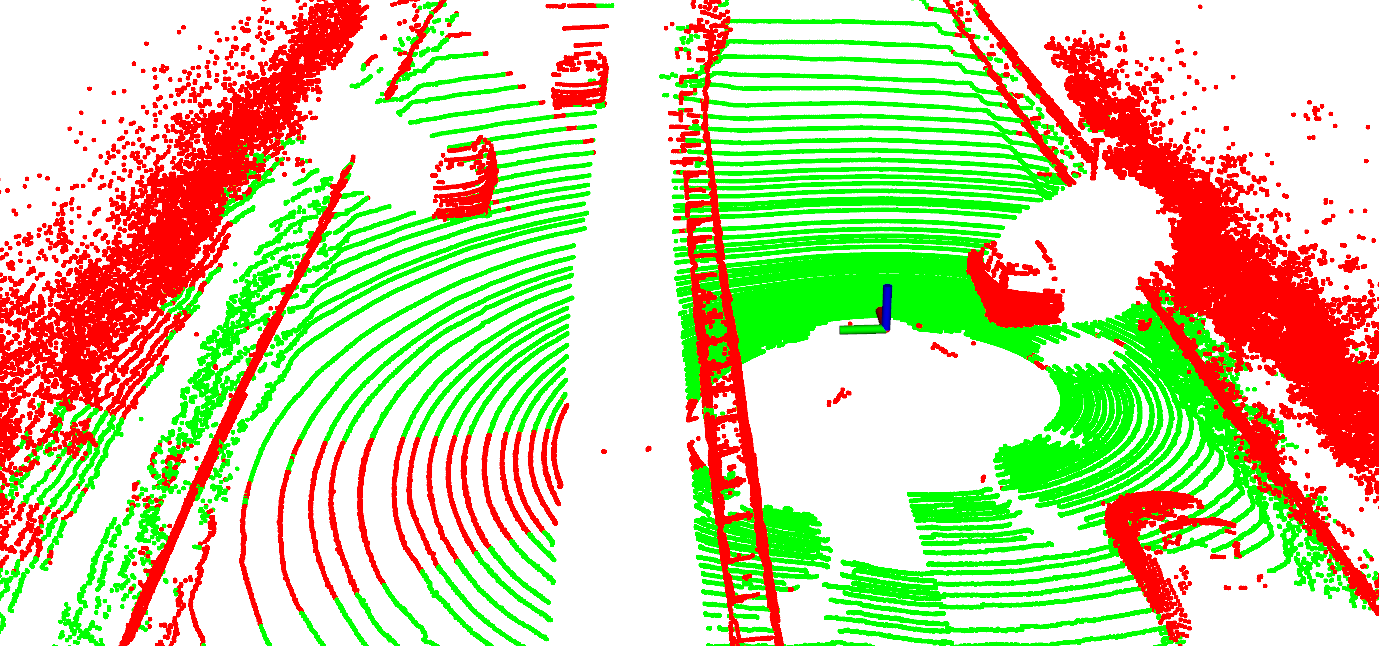}&\includegraphics[width=.33\textwidth,height=2.11cm]{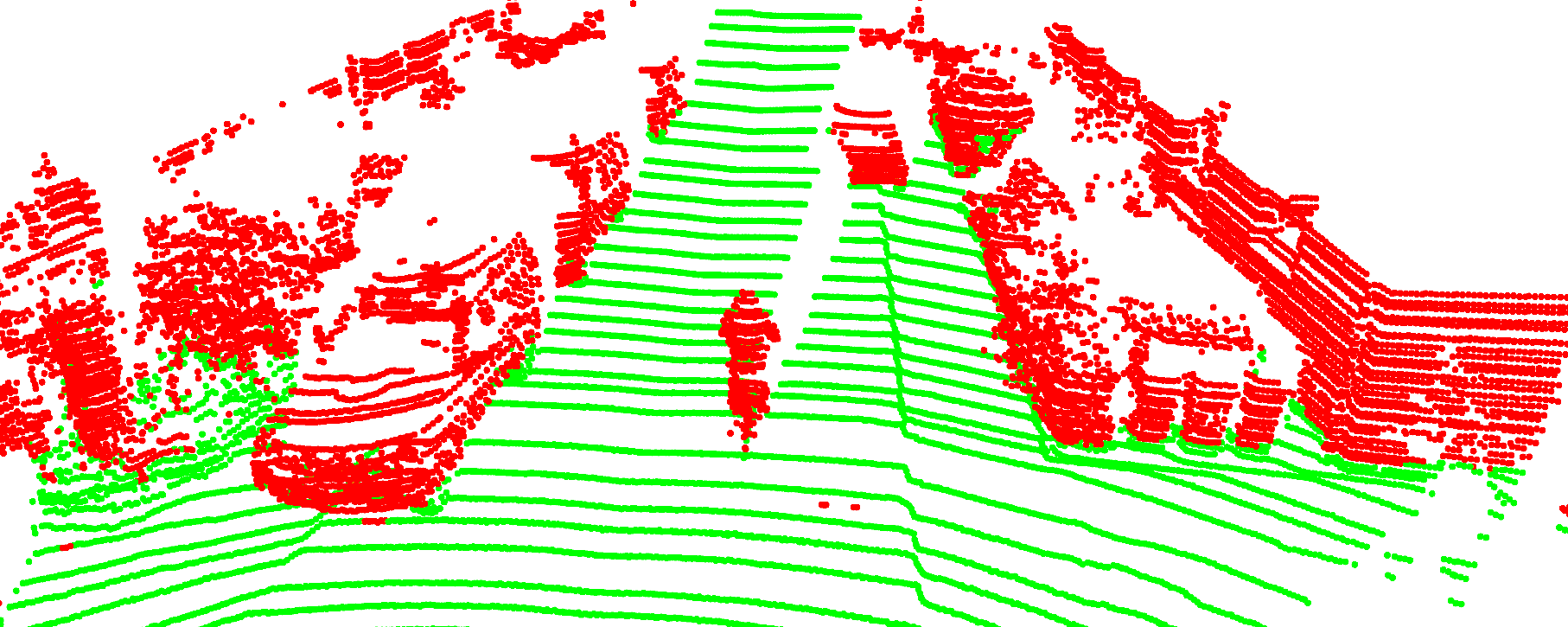}\\
    \rotatebox{90}{\begin{minipage}{2.11cm}\small\centering{DepthClustering}\end{minipage}}&
    \includegraphics[width=.37\textwidth,height=2.11cm]{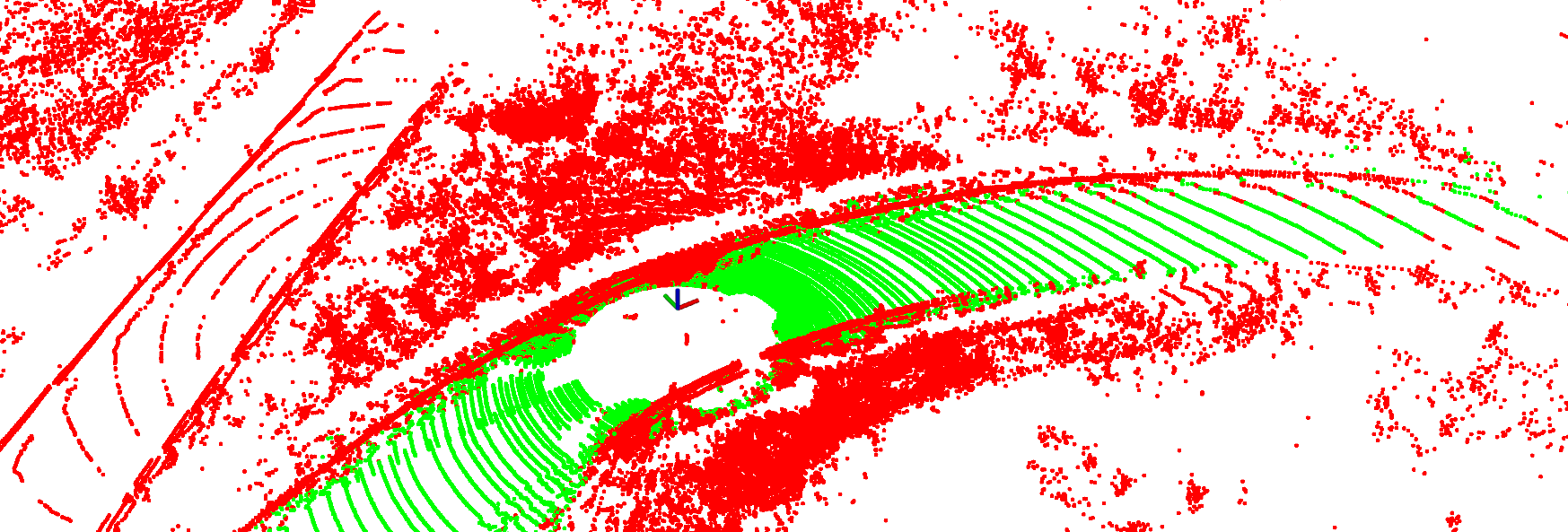}&\includegraphics[width=.27\textwidth,height=2.11cm]{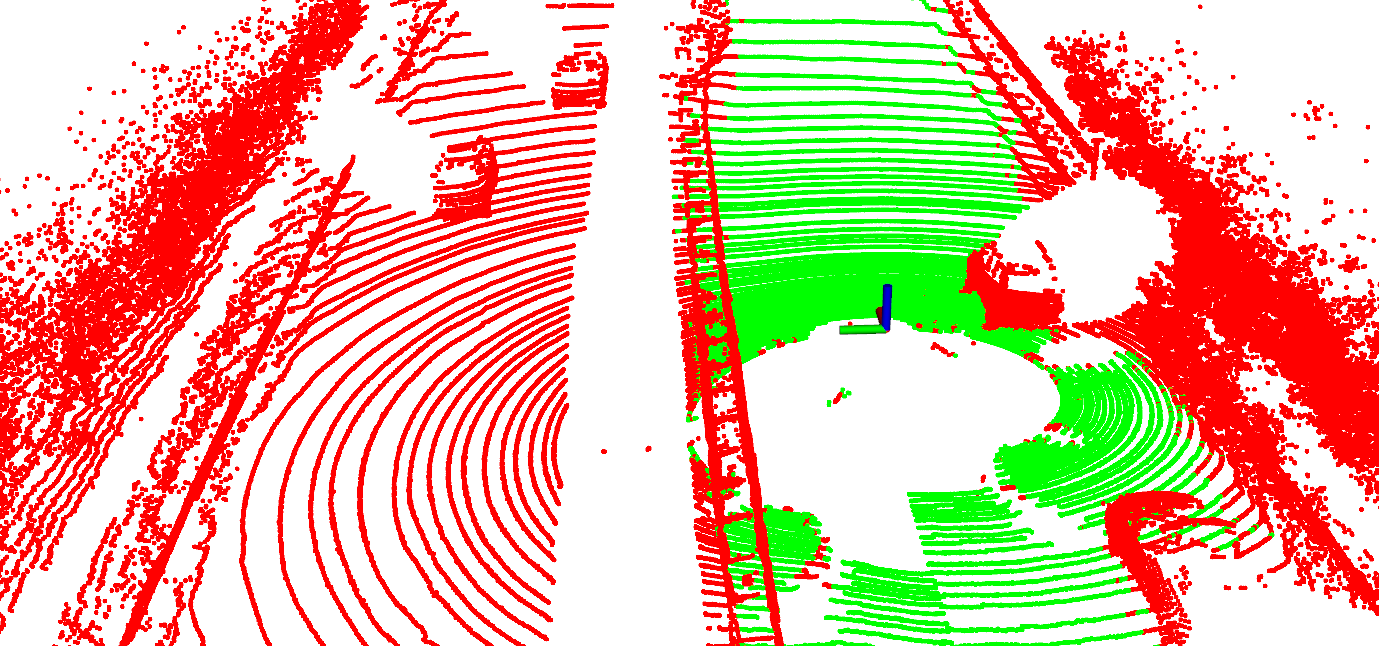}&\includegraphics[width=.33\textwidth,height=2.11cm]{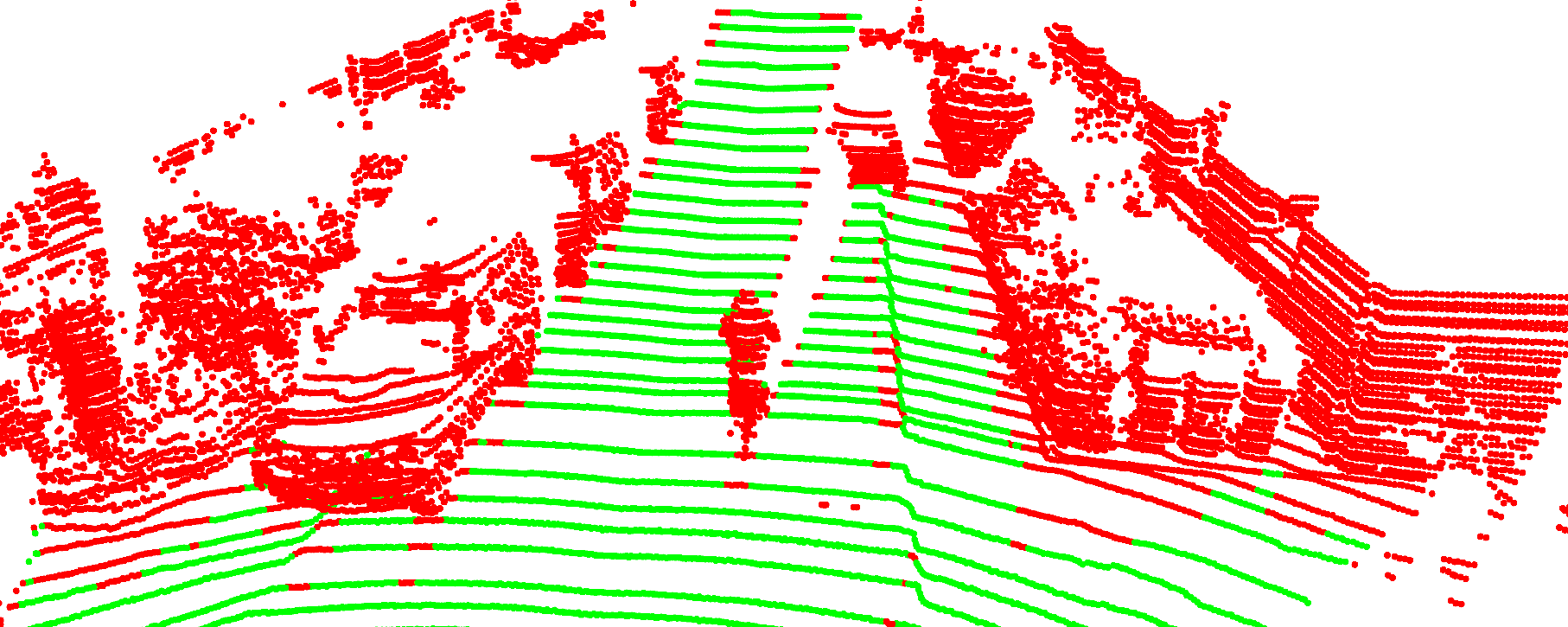}\\
    \rotatebox{90}{\begin{minipage}{2.11cm}\centering{JCP}\end{minipage}}&
    \includegraphics[width=.37\textwidth,height=2.11cm]{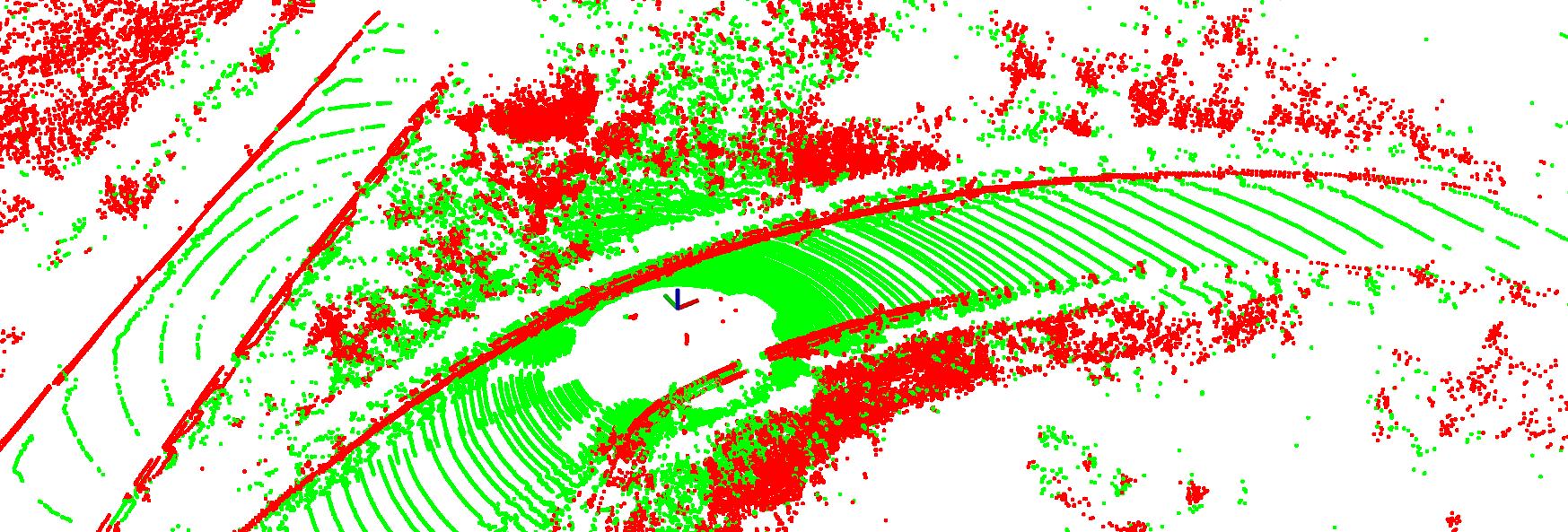}&\includegraphics[width=.27\textwidth,height=2.11cm]{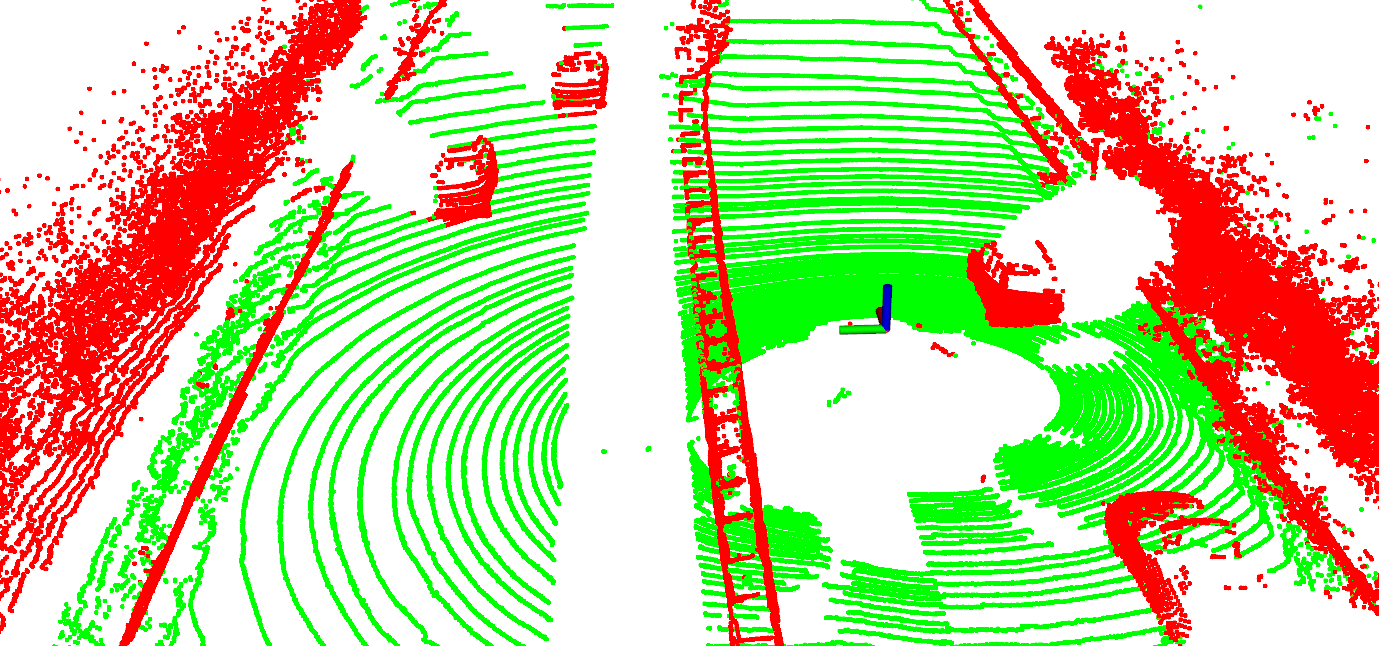}&\includegraphics[width=.33\textwidth,height=2.11cm]{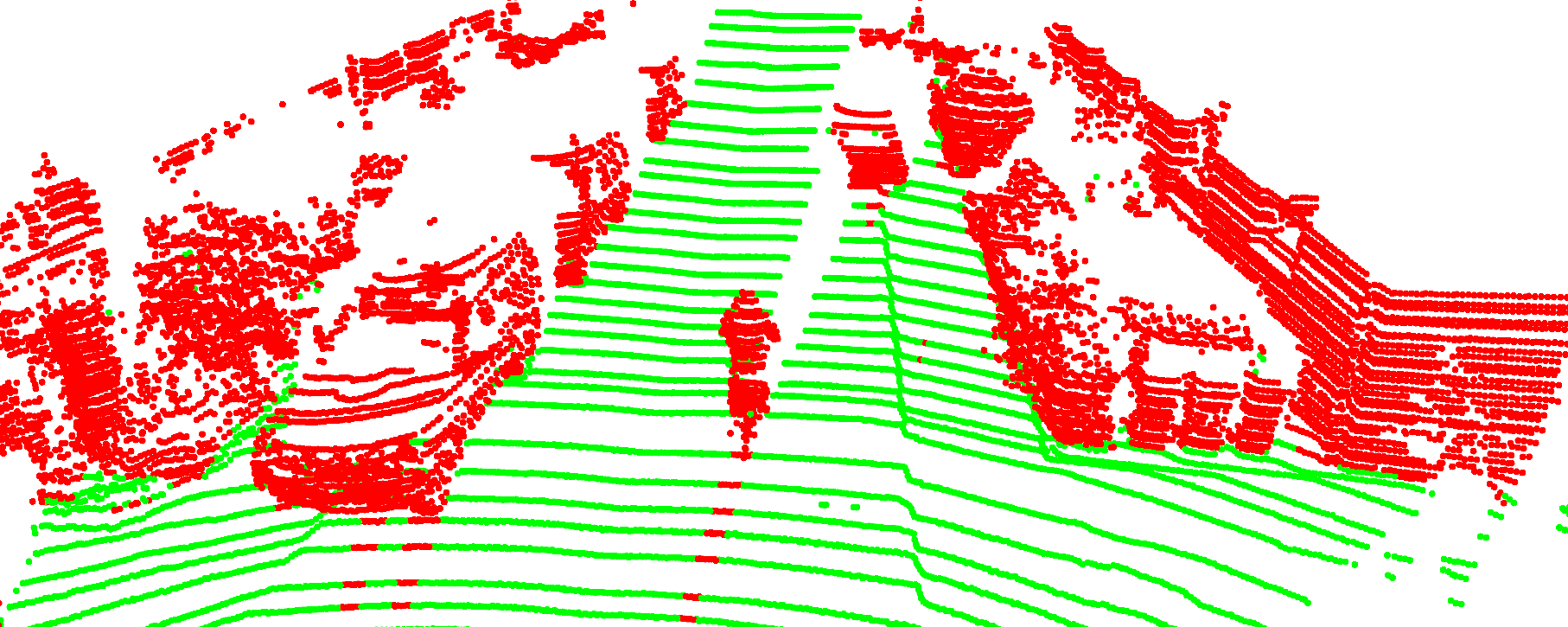}\\
    \rotatebox{90}{\begin{minipage}{2.11cm}\centering{GroundGrid}\end{minipage}}&
    \includegraphics[width=.37\textwidth,height=2.11cm]{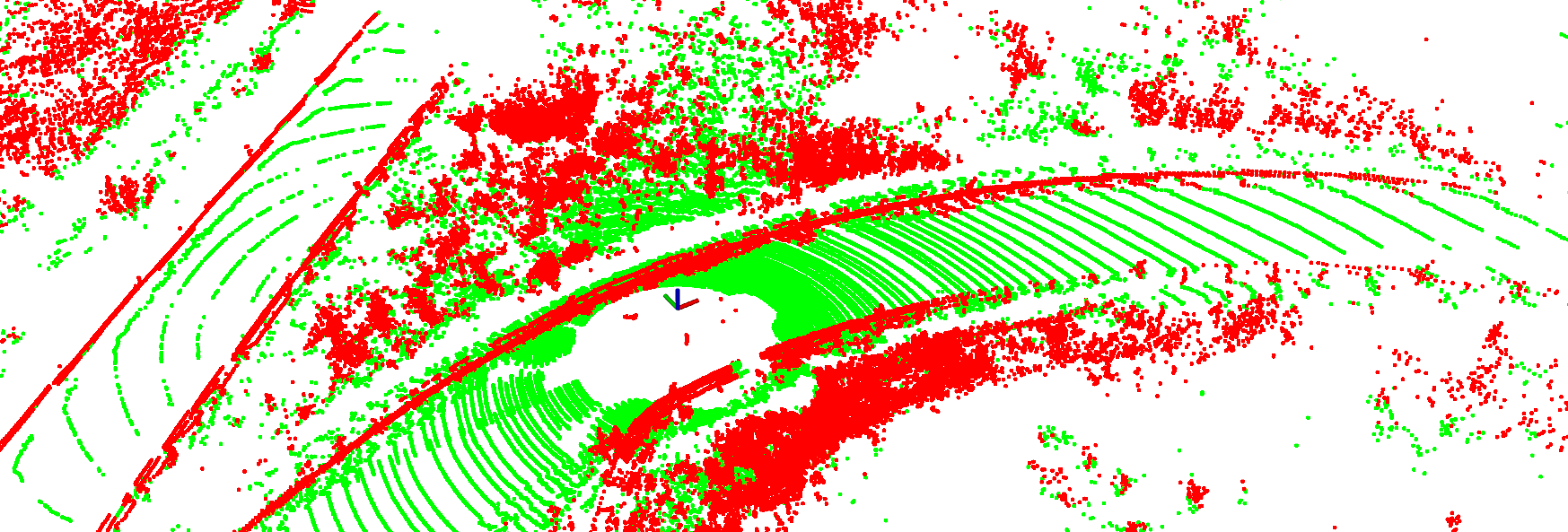}&\includegraphics[width=.27\textwidth,height=2.11cm]{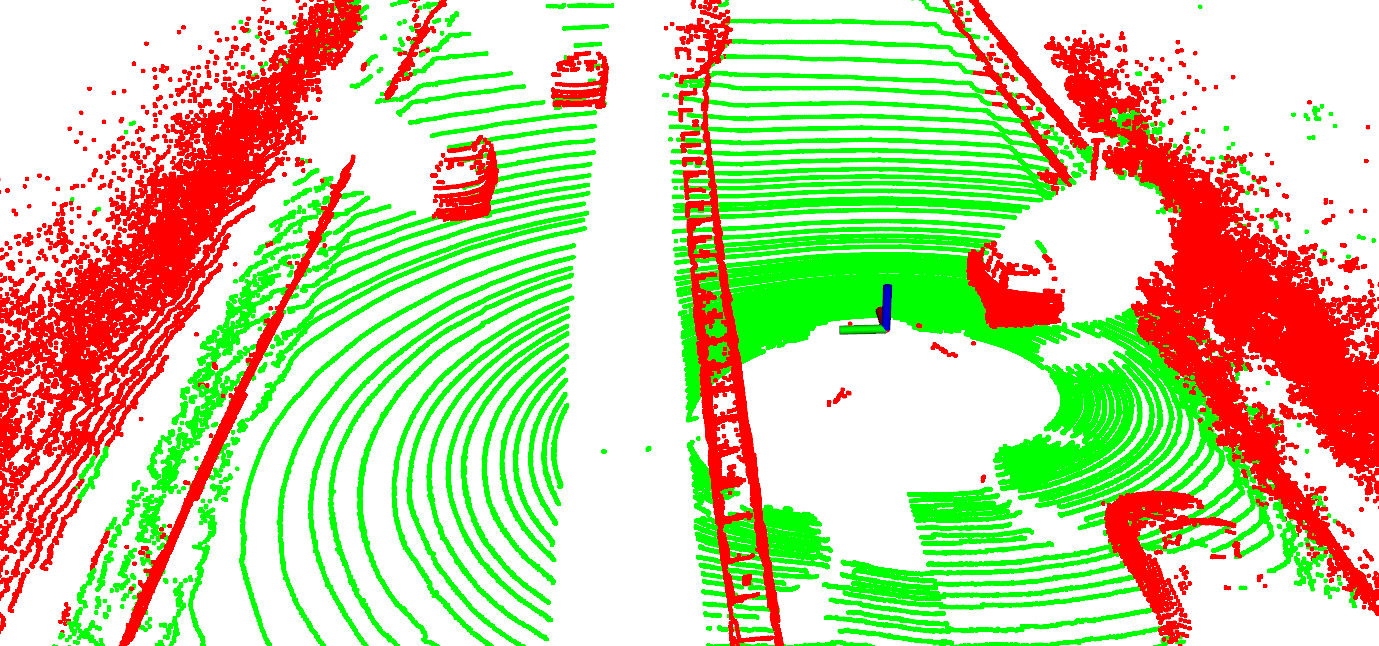}&\includegraphics[width=.33\textwidth,height=2.11cm]{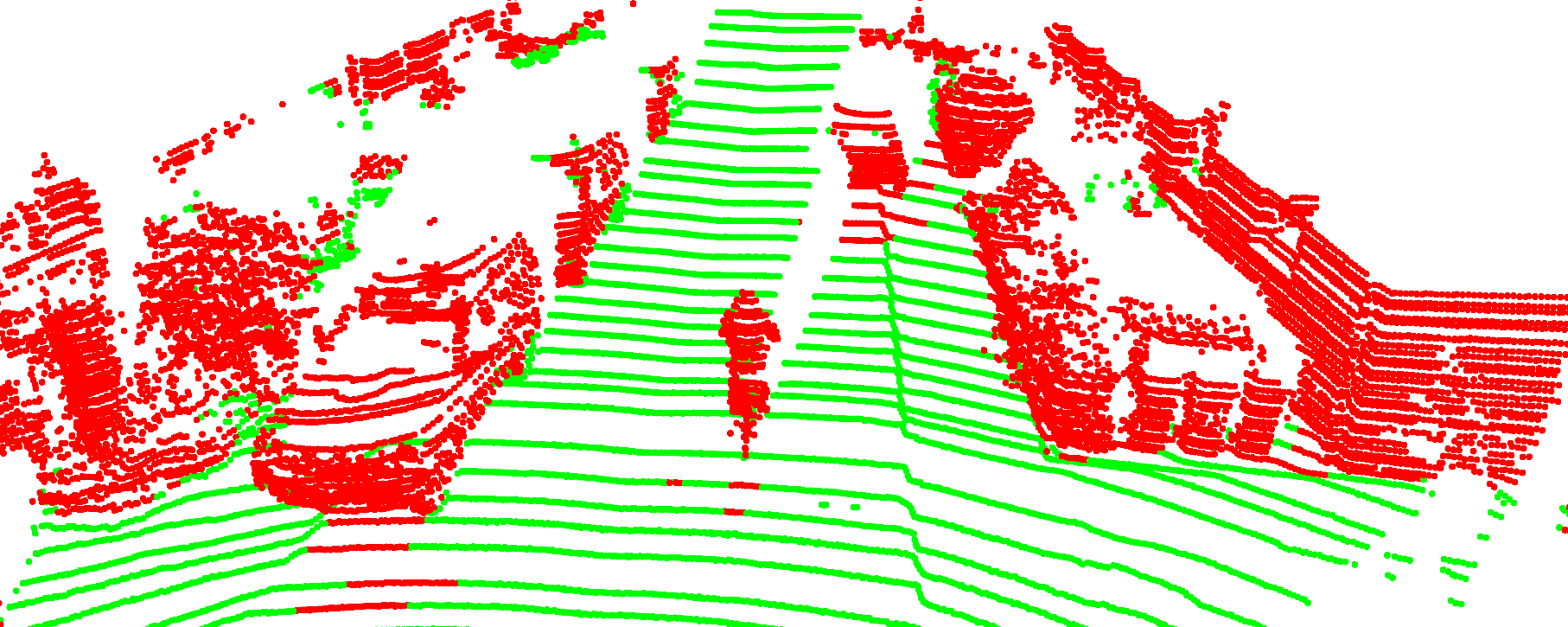}\\
    \rotatebox{90}{\begin{minipage}{2.11cm}\centering{DipG-Seg}\end{minipage}}&
    \includegraphics[width=.37\textwidth,height=2.11cm]{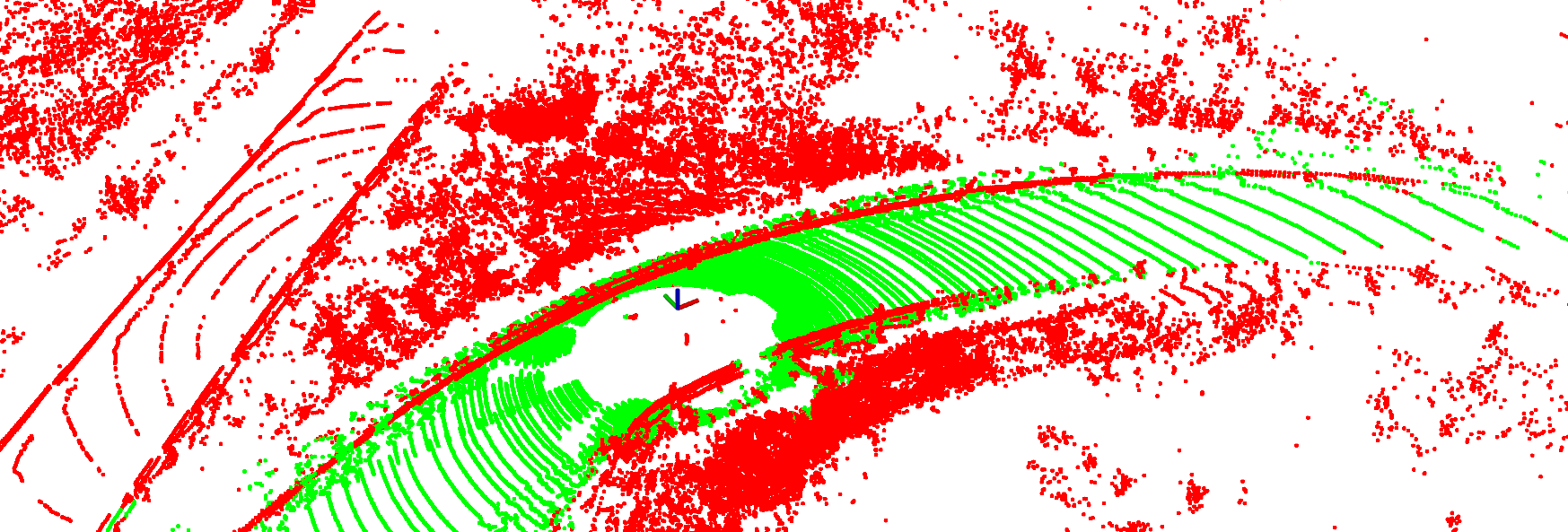}&\includegraphics[width=.27\textwidth,height=2.11cm]{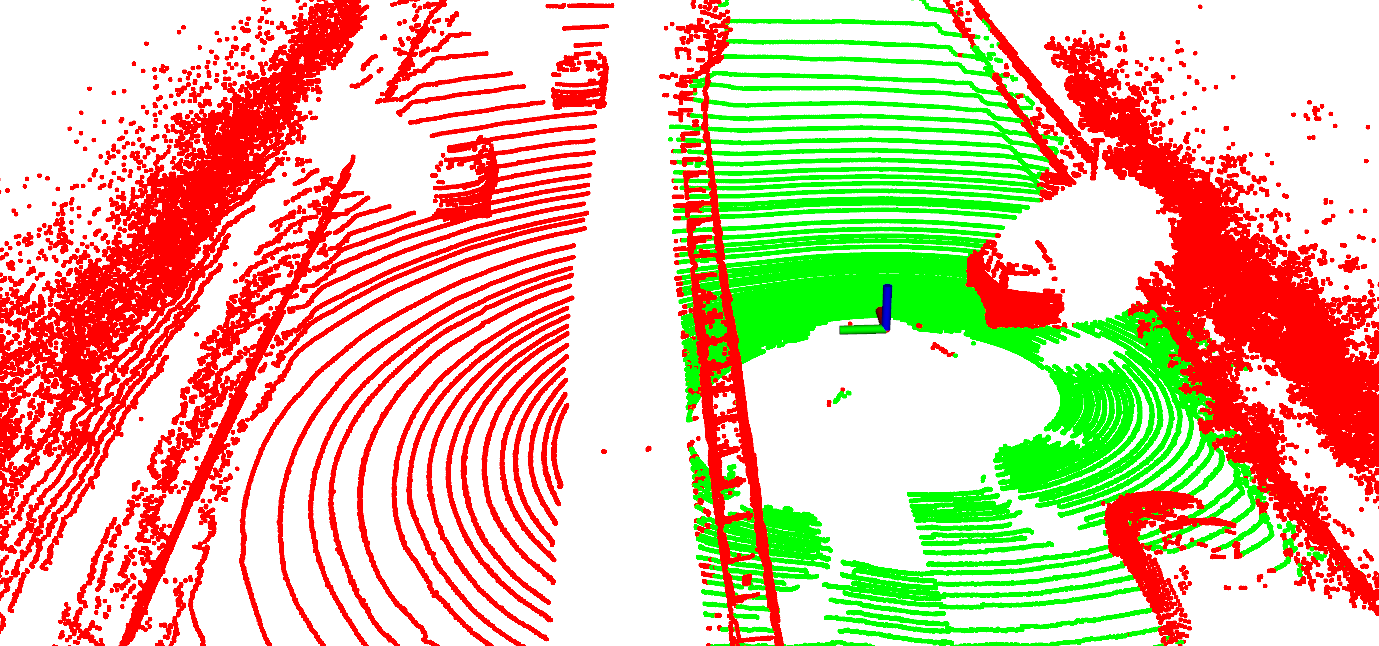}&\includegraphics[width=.33\textwidth,height=2.11cm]{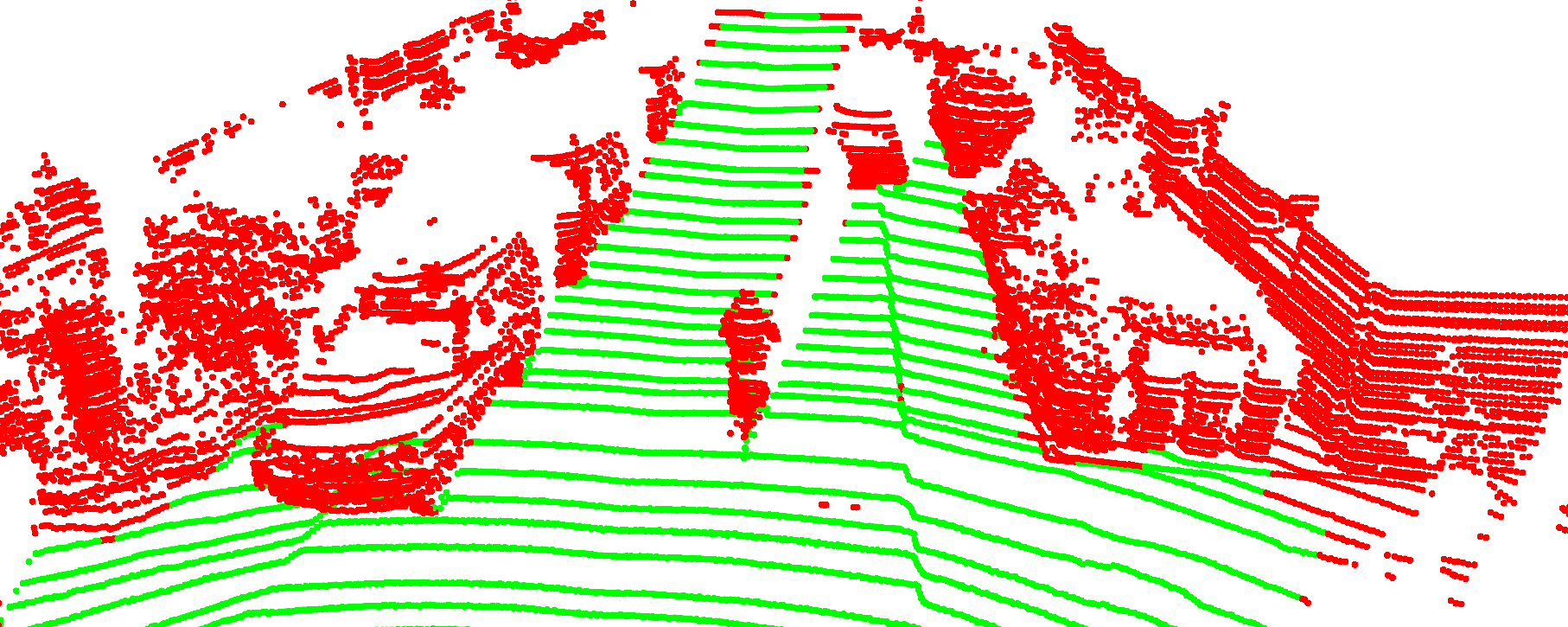}\\
    \rotatebox{90}{\begin{minipage}{2.11cm}\centering{FugSeg (ours)}\end{minipage}}&
    \includegraphics[width=.37\textwidth,height=2.11cm]{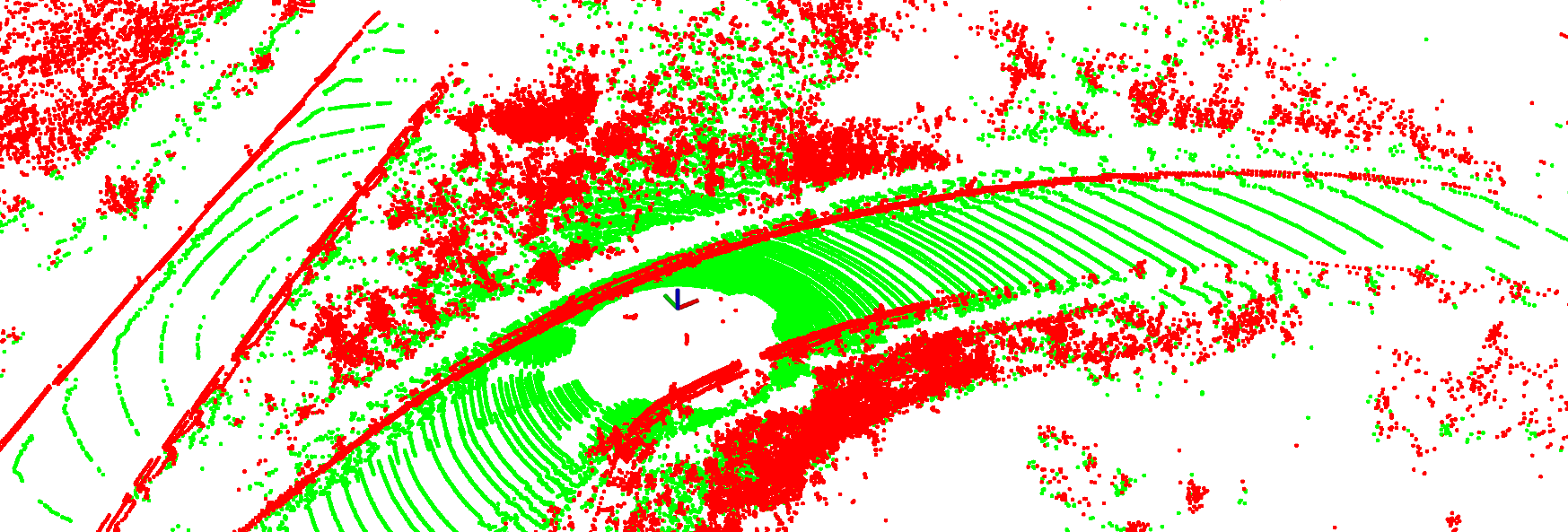}&\includegraphics[width=.27\textwidth,height=2.11cm]{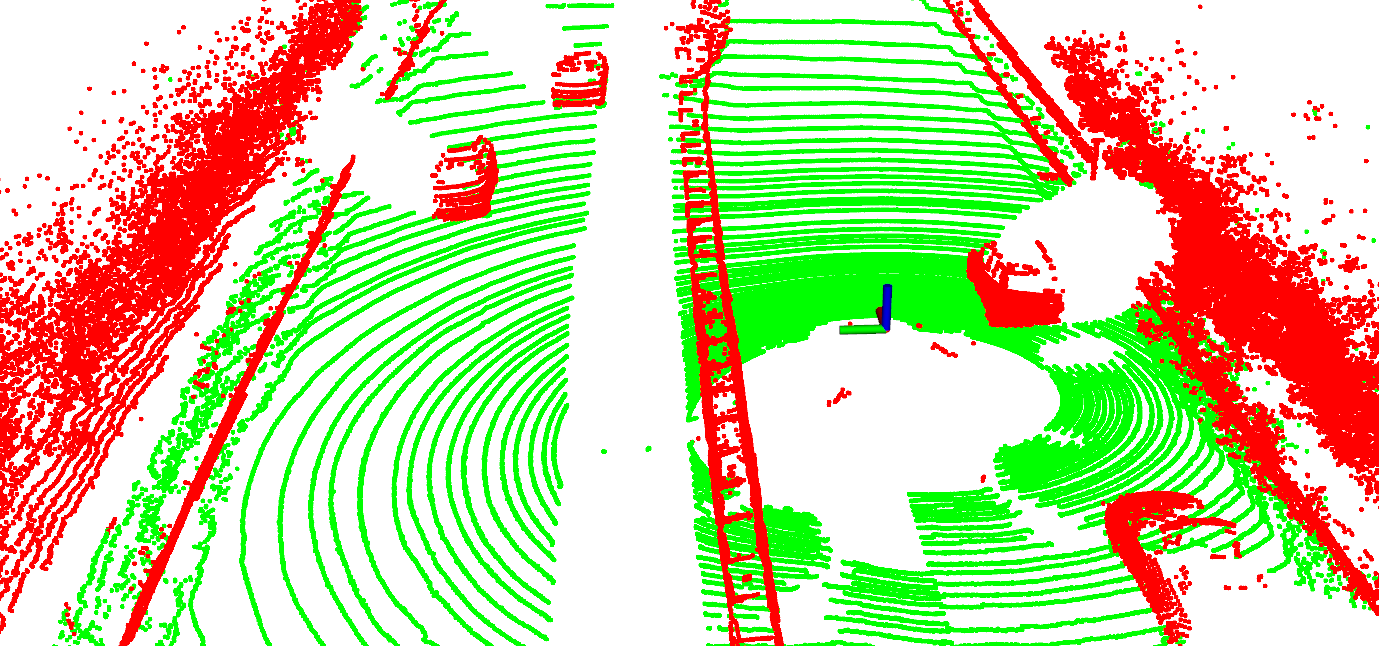}&\includegraphics[width=.33\textwidth,height=2.11cm]{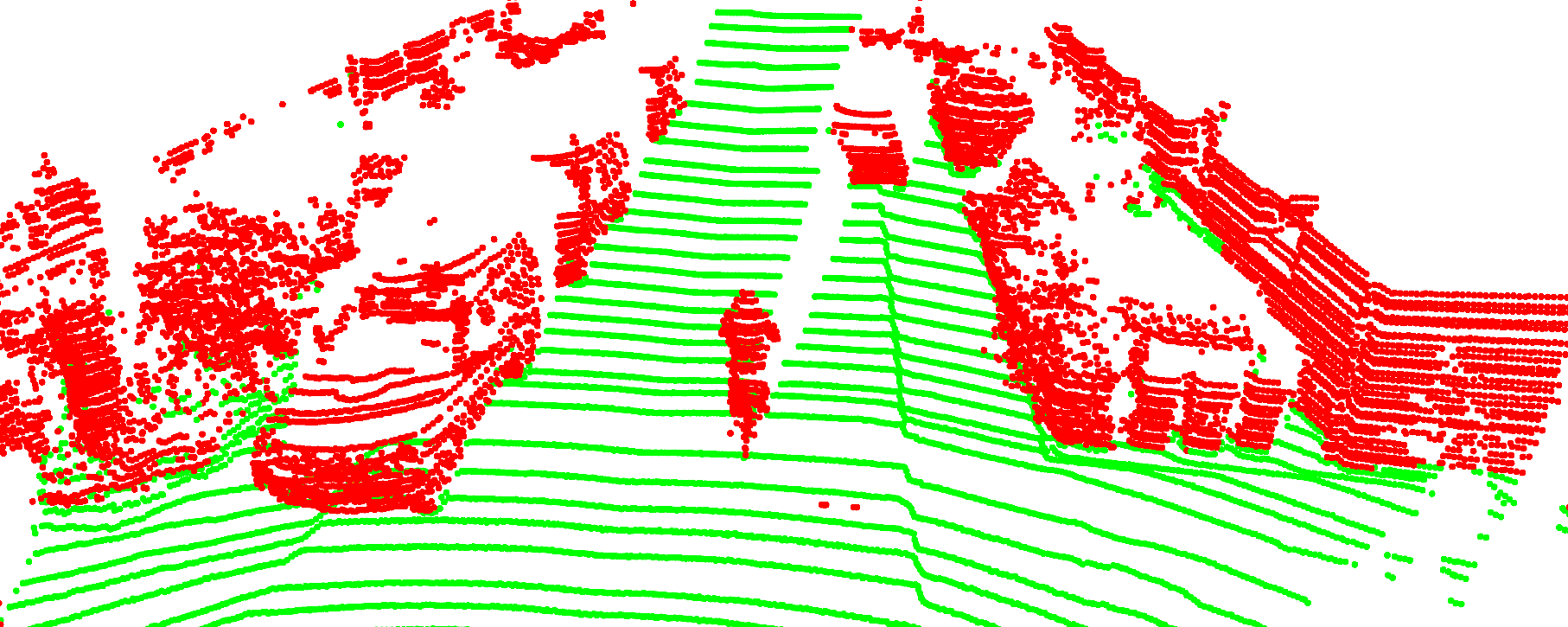}\\
\end{tabular}
    \caption{Qualitative comparison. Scenario 1: curvy slip road with a neighboring path on the left; scenario 2: bidirectional highway with a central barrier; scenario 3: urban road surrounded by cyclist, cars and other objects. Green: ground; red: non-ground.}
    \label{fig8}
\end{figure*}
\setlength{\tabcolsep}{6pt}
\renewcommand{\arraystretch}{1.0}

Figure~\ref{fig8} provides a qualitative comparison across three representative driving scenarios: slip road, highway and urban. In the first scenario, plane-fitting based methods RANSAC-300 and GPF-50 struggle to recover the complete ground surface due to the road slope and curvature. A similar effect is observed for LineFit, whose filtering condition relies on thresholding the absolute slope. Another challenge arises from the isolated neighboring path on the left, which is correctly segmented by only five methods: Patchwork++, TRAVEL, JCP, GroundGrid and FugSeg.

The second scenario presents a relatively flat highway separated by a central barrier. However, the left part of the highway is completely missed by the range-image based methods DepthClustering and DipG-Seg, likely due to inaccurately projected spatial relations between adjacent pixels\cite{9578697}. In addition, Patchwork++ over-segments the roadside vegetation, whereas LineFit incorrectly labels part of the left road surface as non-ground.

The third scenario mainly demonstrates the influence of traffic participants and their associated reflection noise. Notably, reflection noise from the cyclist and parked cars causes substantial interference for DepthClustering, JCP and GroundGrid. Furthermore, transitions between objects pose another challenge for range-image based approaches (see DepthClustering and DipG-Seg). Moreover, Patchwork++ and LineFit exhibit pronounced over-segmentation of roadside objects like the vegetation on the left, while TRAVEL under-segments the road surface between the cyclist and the vehicle ahead.

\subsection{Ablation Studies}
\label{sec_ablation}

\begin{table*}
\begin{center}
\caption{Ablation Studies for Individual Components. Runtime in [$ms$], and all other metrics in [\%].}
\label{tab2}
        \begin{tabular}{*{7}c r *{5}c r}
        \toprule
        \multirow[b]{2}{*}{\textbf{Component}} & \multirow[b]{2}{*}{\textbf{Method}} & \multicolumn{6}{c}{{SemanticKITTI}} & \multicolumn{6}{c}{{nuScenes}}  \\
        \cmidrule(lr){3-8}  \cmidrule(lr){9-14}
        & & \textbf{P} & \textbf{R} & {$\mathbf{F_1}$} & \textbf{A} & \textbf{mIoU} & \multicolumn{1}{c}{\textbf{t}} & \textbf{P} & \textbf{R} & {$\mathbf{F_1}$} & \textbf{A} & \textbf{mIoU} & \multicolumn{1}{c}{\textbf{t}}  \\
        \midrule
        \multirow[c]{3}{1.6cm}{\centering PGM: Radial Division}
        & manual        & \textbf{96.10} & 92.38 & 94.20 & 93.16 & 84.78 & \textbf{7.065}       &   \textbf{95.17} & 89.96 & 92.49 & 94.36 & 88.68  & \textbf{1.661}  \\
        & linear        & 95.86 & \textbf{97.87} & \textbf{96.85} & \textbf{96.30} & \textbf{90.99} & 10.382  &   95.08 & 95.78 & 95.43 & 96.38 & 92.71  & 2.718  \\
        & equidistant   & 95.98 & 97.72 & 96.84 & 96.27 & 90.90 & 7.385           &   95.05 & \textbf{96.18} & \textbf{95.61} & \textbf{96.52} & \textbf{92.97}  & 2.053  \\
        \hline
        \multirow[c]{4}{1.6cm}{\centering UGL}
        & SGL+TS        & \textbf{96.48} & 94.40 & 95.43 & 94.51 & 87.45 & \textbf{7.071}  &  \textbf{95.78} & 91.96 & 93.83 & 95.21 & 90.43 & \textbf{1.934}  \\
        & SGL+AS        & 96.39 & \textbf{95.36} & \textbf{95.87} & \textbf{95.04} & \textbf{88.50} & 7.156  &  95.59 & \textbf{92.78} & \textbf{94.17} & \textbf{95.45} & \textbf{90.91} & 1.954  \\
        \cline{2-14}
        & SGL+CGP+TS    & \textbf{96.08} & 97.46 & 96.77 & 96.17 & 90.69 & \textbf{7.331}  &  \textbf{95.23} & 95.84 & 95.53 & 96.46 & 92.87 & \textbf{2.028}  \\
        & SGL+CGP+AS    & 95.98 & \textbf{97.72} & \textbf{96.84} & \textbf{96.27} & \textbf{90.90} & 7.385           &   95.05 & \textbf{96.18} & \textbf{95.61} & \textbf{96.52} & \textbf{92.97}  & 2.053  \\
        \hline
        \multirow[c]{2}{1.6cm}{\centering EGE}
        & 5\texttimes5 weighting & \textbf{96.31} & 97.15 & 96.73 & 96.06 & 90.44 & \textbf{7.008}  &  \textbf{95.27} & 95.69 & 95.48 & 96.40 & 92.77 & \textbf{2.052}  \\
        & proposed      & 95.98 & \textbf{97.72} & \textbf{96.84} & \textbf{96.27} & \textbf{90.90} & 7.385           &   95.05 & \textbf{96.18} & \textbf{95.61} & \textbf{96.52} & \textbf{92.97}  & 2.053  \\

        \bottomrule
        \end{tabular}
\end{center}
\end{table*}
To assess the contribution of individual components to the overall performance, we conduct ablation studies utilizing the SemanticKITTI and nuScenes datasets. The results are summarized in Table~\ref{tab2}. During the evaluation of each component, all other components are held constant.

\subsubsection{Polar Grid Mapping}
In addition to the proposed equidistant radial division, the manual division method from\cite{9466396} and a linear radial division inspired by\cite{5548059} are also implemented. The manual radial division is carefully designed as a piecewise equidistant radial division method, which results in 14 cells per segment. The idea of linear radial division is to divide each segment with a linearly growing radial resolution $d(j)=a\cdot j+b$, where $j\in[0,M)$. This equation can be solved by constraining its integral $D(j)=\frac{a}{2}\cdot j^2+b\cdot j+c$ with boundary conditions $D(0)=r_0$ and $D(M)=r_M$, together with an initial state $d(0)=d_0$. For the linear method, $M$ and $d_0$ are set to $80$ and $0.05m$ in our experiment.

As shown in Table~\ref{tab2}, the manual method achieves the fastest runtime due to the smaller number of cells per segment (i.e., $M=14$), but its performance is surpassed by the other methods. The linear and equidistant methods exhibit similar performance; however, the proposed equidistant method is faster than the linear one by more than $25\%$.

\subsubsection{Uncertainty-aware Ground Labeling}
\label{sec_ablation_uagl}
As shown in Table~\ref{tab2}, compared with the traditional slope (TS), the proposed adaptive slope (AS) consistently improves FugSeg's performance on both datasets. In addition, the proposed cross-segment propagation (CGP) module significantly enhances recall by over $2\%$ on the SemanticKITTI dataset and nearly $4\%$ on nuScenes, demonstrating its effectiveness in handling occlusions and sparse measurements. Consequently, the configuration ``SGL+CGP+AS'' is identified as the most effective setup for the ground segmentation task.

\subsubsection{Estimation of Ground Elevation}
In addition to the proposed elevation estimation method, we implement a weighting-based baseline for comparison. Following\cite{rs13163239} and\cite{10319084}, the ground elevation of a cell is calculated as the weighted average height of its 5\texttimes 5 \textit{ground} neighbors, where the weighting factors are determined using the same formula as (\ref{equ.3.c.1}). As shown in Table~\ref{tab2}, the proposed method outperforms the 5\texttimes 5 weighting approach across most evaluation metrics on both datasets, demonstrating its effectiveness in handling non-flat ground.

\subsection{Parameter Studies}
\begin{figure}[!t]
    \centering
    \includegraphics[width=3.5in]{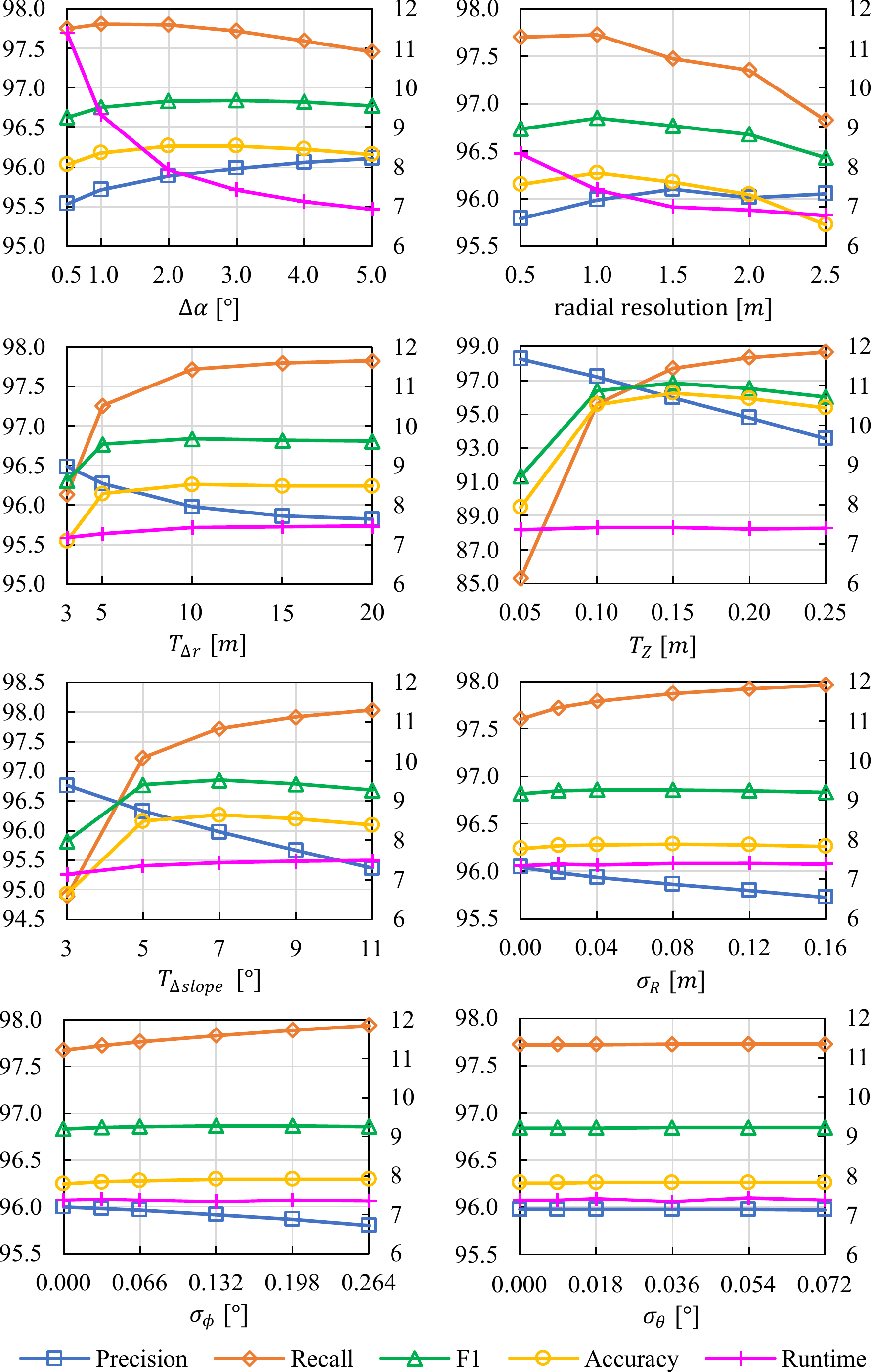}
    \caption{Impact of $\Delta\alpha$, $M$, $T_{\Delta r}$, $T_Z$, $T_{\Delta slope}$, $\sigma_R$, $\sigma_\phi$ and $\sigma_\theta$ on FugSeg's performance on the SemanticKITTI dataset. Left vertical axis: score in [\%]; right vertical axis: runtime in [ms].}
    \label{fig7}
\end{figure}

To investigate the robustness of FugSeg with respect to $\Delta\alpha$, $M$, $T_{\Delta r}$, $T_Z$, $T_{\Delta slope}$, $\sigma_R$, $\sigma_\phi$ and $\sigma_\theta$, extensive sensitivity analyses are conducted on the complete SemanticKITTI dataset. The results are presented in Figure~\ref{fig7}. Note that the mIoU measure is not plotted due to its shifted vertical range, but it basically follows the same trend as $\mathrm{F_1}$ and accuracy.

\subsubsection{Impact of $\Delta\alpha$ and $M$}
Parameters $\Delta\alpha$ and $M$ define the spatial resolution of the constructed polar grid map. As shown in Figure~\ref{fig7}, compared to others, $\Delta\alpha$ and $M$ directly affect FugSeg's efficiency. Increasing the resolution (i.e., decreasing $\Delta\alpha$ or increasing $M$) leads to higher computational demand. However, finer resolution does not necessarily improve performance, as the input point cloud may be divided into excessively small cells. Stable performance is observed around $\ang{3.0}$ and $1.0m$ (i.e., $M=80$), which reflects the characteristic spatial distribution of SemanticKITTI's LiDAR measurements.

\subsubsection{Impact of $T_{\Delta r}$ and $T_Z$}
Parameter $T_{\Delta r}$ constrains the baseline length when propagating the ground label from a \textit{ground} cell to the next unlabeled cell in Algorithm~\ref{alg1}, reflecting the sparsity and occlusion level of the point cloud. As shown in Figure~\ref{fig7}, overall performance improves as $T_{\Delta r}$ increases from $3m$ to $10m$, but saturates thereafter with gradually increasing runtime beyond $10m$. Parameter $T_Z$ specifies the permitted elevation deviation during point-level ground segmentation and jointly reflects the measurement accuracy and ground roughness. As shown in Figure~\ref{fig7}, larger $T_Z$ leads to higher recall but lower precision, as more points are misclassified as ground. However, when $T_Z$ approaches the measurement resolution (about $2cm$ on SemanticKITTI), recall drops sharply because ground and non-ground points become indistinguishable. A balance between precision and recall is achieved when $T_Z$ is in the range of $0.10$--$0.20m$.

\subsubsection{Impact of $T_{\Delta slope}$}
Parameter $T_{\Delta slope}$ is used to identify new ground cells during the proposed within- and cross-segment ground labeling. For better scalability, it constrains the change in slope rather than the slope itself. As shown in Figure~\ref{fig7}, decreasing $T_{\Delta slope}$ improves precision but significantly reduces recall, as a small $T_{\Delta slope}$ makes FugSeg overly conservative. However, increasing $T_{\Delta slope}$ enhances recall but leads to a drop in precision due to increased misclassification of non-ground cells. Balanced performance is observed when $T_{\Delta slope}$ is set between $\ang{5}$ and $\ang{9}$, suggesting an overall moderate terrain gradient in the SemanticKITTI dataset.

\subsubsection{Impact of $\sigma_R$, $\sigma_\phi$ and $\sigma_\theta$}
Parameters $\sigma_R$, $\sigma_\phi$ and $\sigma_\theta$ represent the measurement uncertainties of the LiDAR sensor in the radial, elevation, and azimuthal directions, respectively. Potential misestimation of these parameters may affect the performance of the adaptive slope in Equation~(\ref{equ10}), and consequently the overall segmentation results. As shown in Figure~\ref{fig7}, variations in these parameters have a relatively minor impact on FugSeg's performance, indicating that the proposed adaptive slope formulation is robust to potential misestimation of sensor noise parameters.

In summary, FugSeg demonstrates stable performance under moderate parameter variations, indicating that its effectiveness is not overly dependent on specific parameter settings. Nevertheless, the optimal configuration may vary depending on the characteristics of LiDARs and the operational environment.

\subsection{Real-world Experiments}
\begin{figure*}[!t]
    \includegraphics[width=1.0\textwidth]{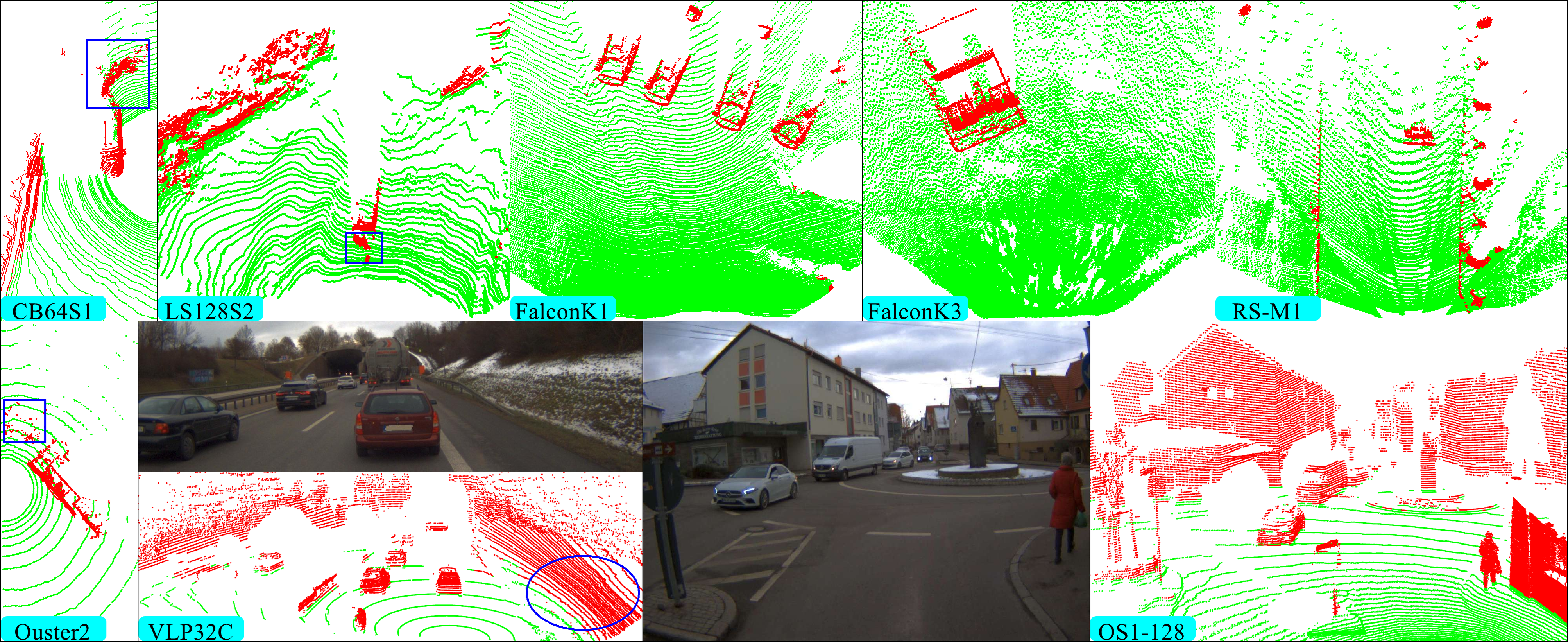}
    \caption{Performance of FugSeg on different LiDAR sensors across various environments. VLP32C and OS1-128 are self-collected measurements (including the camera views), and the remaining samples are from the LiDARDustX dataset. Blue rectangles highlight correctly segmented dust particles in unstructured environments, whereas the blue ellipse marks the misclassified roadside terrain due to its steep slope. More examples are available on the FugSeg webpage.}
    \label{fig12}
\end{figure*}

\begin{table}
\begin{center}
\caption{Contribution of FugSeg to the road boundary detection task, evaluated on the SemanticKITTI validation set.}
\label{tab8}
        \begin{tabular}{*{5}c}
        \toprule
        setup & \textbf{P} & \textbf{R} & {$\mathbf{F_1}$} & \textbf{t}  \\
        \midrule
        without FugSeg   & 67.17 & 75.90 & 71.27 & 66.173   \\
        with FugSeg    & 84.43 & 81.13 & 82.75 & 52.256   \\
        \bottomrule
        \end{tabular}
\end{center}
\end{table}

\begin{figure}[!t]
    \centering
    \includegraphics[width=3.4in]{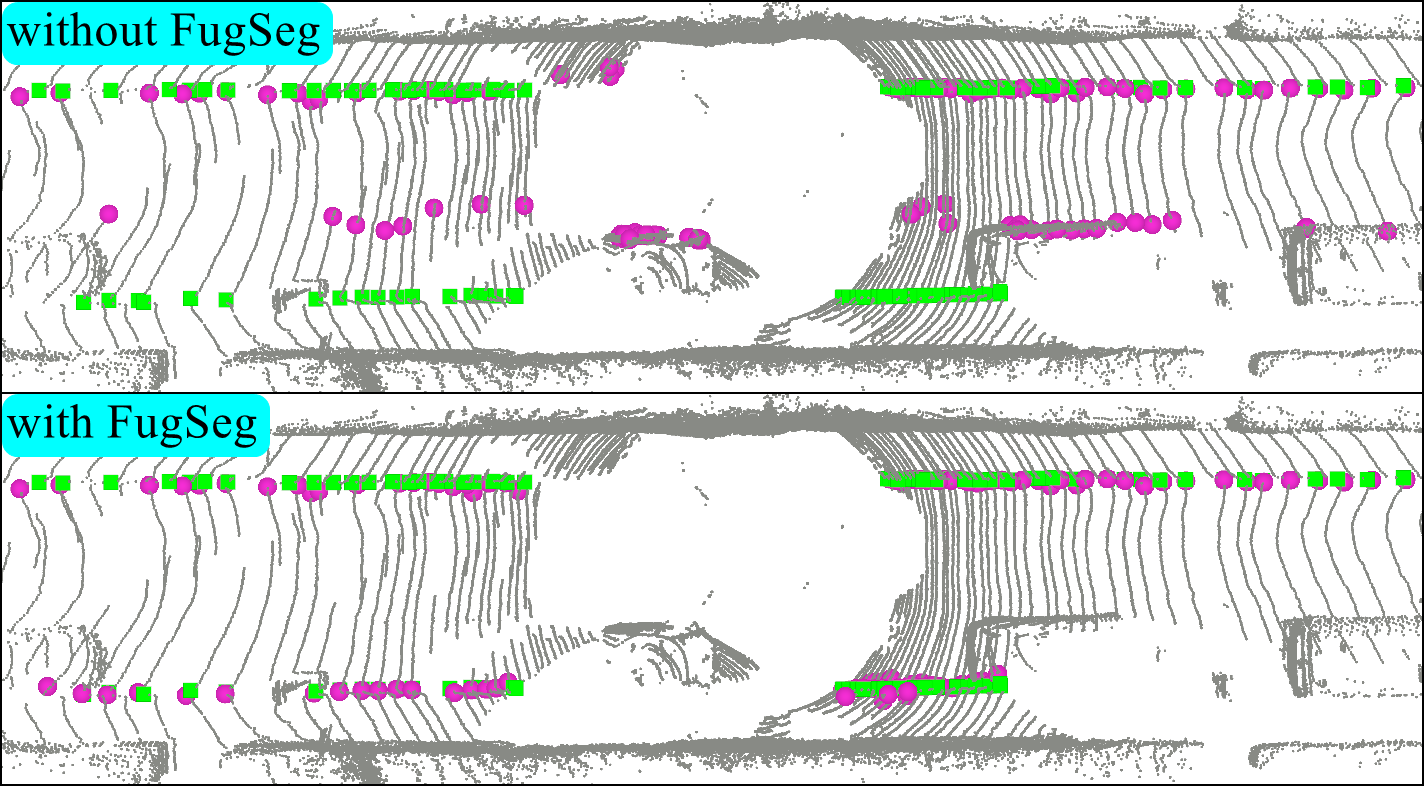}
    \caption{Road boundary detection with the support of FugSeg. FugSeg helps to find actual road boundaries (in green) in occluded scenes, leading to more accurate and efficient detection. Red: detected road boundaries.}
    \label{fig13}
\end{figure}

\setlength{\tabcolsep}{0.0pt}
\renewcommand{\arraystretch}{0.0}
\begin{figure}[!t]
    \centering
    \begin{tabular}{|c|}
        \hline
        \includegraphics[width=3.4in]{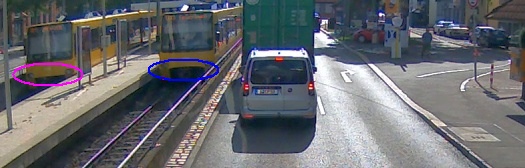}  \\
        \includegraphics[width=3.4in]{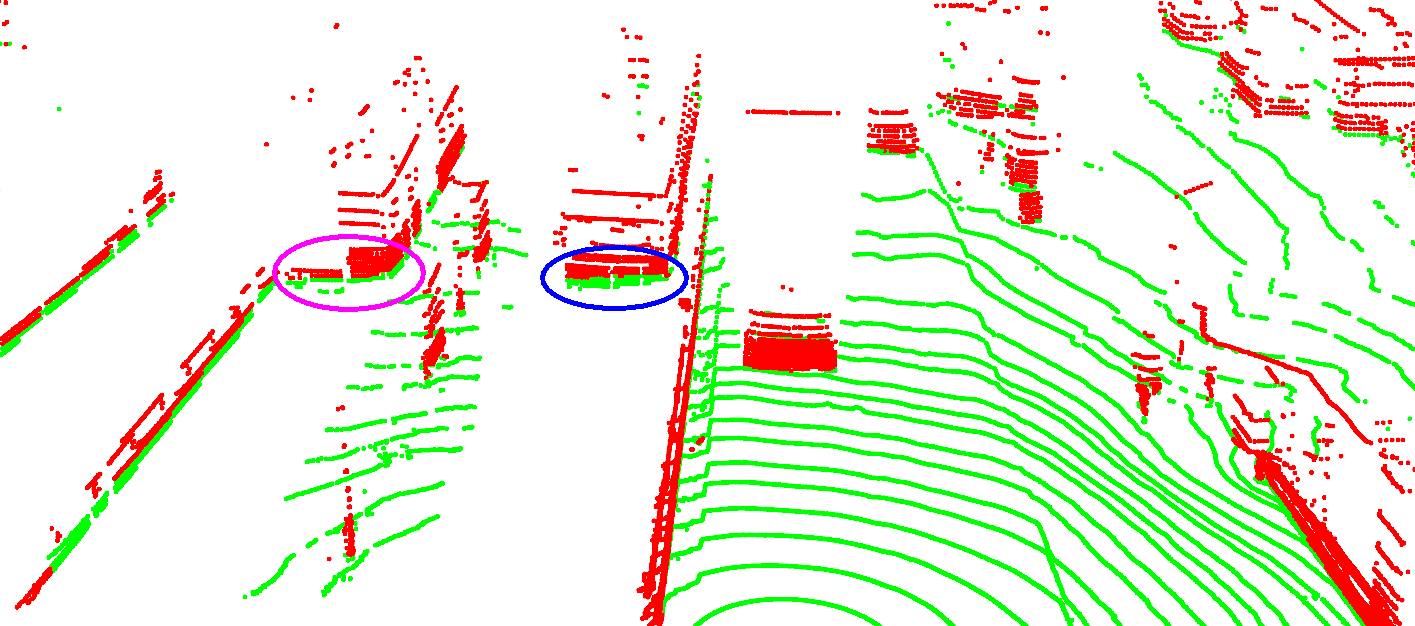}  \\
        \hline
    \end{tabular}
    \caption{A highly cluttered scene in which the chassis of parked trams are erroneously classified as ground (highlighted by ellipses), while the platform ground surface is correctly segmented.}
    \label{fig14}
\end{figure}
\setlength{\tabcolsep}{6pt}
\renewcommand{\arraystretch}{1.0}

To further validate the generalizability of FugSeg across different LiDAR sensors and environments, we conduct additional experiments on real-world measurements beyond the public datasets. The self-collected dataset includes both sparse and dense LiDAR scans and covers structured and unstructured terrain. As shown in Figure~\ref{fig12}, FugSeg consistently demonstrates robust performance across all tested sensors and environments. Notably, it effectively identifies dust particles as non-ground in unstructured scenes, highlighting its resilience to terrain undulation and environment noise.

To demonstrate FugSeg's applicability, we integrate it into a downstream task: road boundary detection. As shown in Table~\ref{tab8} and Figure~\ref{fig13}, with the support of FugSeg, the road boundary detector achieves improved efficiency and performance, confirming its effectiveness in practical applications.

However, FugSeg still faces challenges in highly cluttered and ambiguous scenes. In the VLP32C scene shown in Figure~\ref{fig12}, FugSeg struggles to accurately segment steep roadside terrain. Figure~\ref{fig14} presents another example in which the chassis of parked trams are erroneously classified as ground, even though the platform ground surface is correctly segmented. This misclassification is likely caused by the similar height and slope of the tram chassis relative to the surrounding ground, which confuses the slope-based ground propagation strategy. Future work will investigate potential solutions to improve FugSeg's performance in such challenging scenarios.

\subsection{Runtime}
\begin{table}
\begin{center}
\caption{Component-wise runtime distribution (in millisecond).}
\label{tab4}
        \begin{tabular}{*{8}c}
        \toprule
        Dataset & PGM & UGL & EGE & PGS & Sum  \\
        \midrule
        SemanticKITTI   & 4.676 & 0.329 & 0.262 & 2.118 & 7.385   \\
        nuScenes        & 1.185 & 0.203 & 0.210 & 0.455 & 2.053   \\
        KITTI-360        & 4.989 & 0.262 & 0.221 & 1.900 & 7.372   \\
        LiDARDustX        & 1.700 & 0.119 & 0.137 & 0.734 & 2.690   \\
        \bottomrule
        \end{tabular}
\end{center}
\end{table}

Table~\ref{tab4} summarizes the runtime distribution across individual components. More than half of the computation time is consumed by the polar grid mapping, followed by the point-level ground segmentation, which accounts for approximately 20--30\% of the total runtime. The remainder arises from the within- and cross-segment ground labeling and the ground elevation estimation. This distribution highlights the importance of employing efficient data representations rather than directly processing unordered 3D point clouds. It is worth noting that all reported runtimes were obtained under single-threaded execution. To further improve FugSeg's efficiency, point-level operations such as PGM and PGS can be optimized through parallel processing and hardware acceleration.

\section{Conclusion}
This work presents FugSeg, a fast uncertainty-aware ground segmentation method for LiDAR-based perception systems. By employing a polar grid map, FugSeg achieved consistent generalizability across LiDAR sensors with different scanning patterns. A within- and cross-segment labeling strategy was introduced to identify not only directly visible ground regions, but also isolated or occluded ground cells that are commonly missed by existing methods. To enhance robustness in complex terrain, we proposed an adaptive slope formulation that incorporates measurement uncertainties, along with a fine-grained elevation estimation module for accurate point-level labeling. Reflection noise was explicitly handled through the proposed \textit{noisy ground} cells, which improved FugSeg's reliability in highly dynamic driving environments. Extensive experiments on eleven LiDAR sensors spanning five datasets demonstrate FugSeg's robustness under diverse environmental conditions, and its superior efficiency reduces the development cost in large-scale applications. To support further research, we will release our code publicly.

For future work, we plan to integrate FugSeg into downstream tasks such as traversability analysis and object detection. We also aim to evaluate its contribution to the complete ITS system presented in Figure~\ref{fig10}, once supporting infrastructure like connectivity and cloud computing is in place.

\bibliography{literature.bib}
\bibliographystyle{IEEEtran}

\begin{IEEEbiography}[{\includegraphics[width=1in,height=1.25in,clip,keepaspectratio]{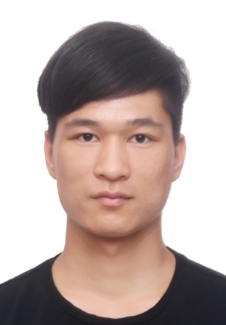}}]{Yu Li}
    received the B.Eng. degree in geodesy and geomatics engineering from Wuhan University, Wuhan, China, in 2016, and the M.Sc. degree in geomatics engineering from the University of Stuttgart, Stuttgart, Germany, in 2019. He is currently pursuing the Ph.D. degree with the University of Stuttgart, and meanwhile working as a full-time software engineer at Daimler Truck AG. His current research interests include environment perception and autonomous driving.
\end{IEEEbiography}

\vspace{-25pt}

\begin{IEEEbiography}[{\includegraphics[width=1in,height=1.25in,clip,keepaspectratio]{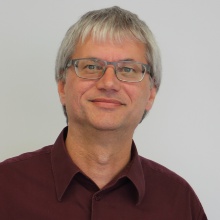}}]{Volker Schwieger}
    was born in Springe near Hannover. He studied geodesy at the University of Hannover and obtained his doctorate in 1997 on the subject of GPS monitoring measurements at the Geodetic Institute of the University of Hannover. In 2000 and 2001 he worked at the German Research Centre for Geosciences Potsdam (GFZ), where he was responsible for determining satellite orbits. In 2002 he moved to the Institute for Applications of Geodesy to Engineering at the University of Stuttgart. He obtained his habilitation there in 2004 with a dissertation on the subject of kinematic measurements, sensor fusion and modelling. He has been director of the institute since 2010, having renamed it to Institute of Engineering Geodesy. He also coordinates international and national research groups on the subjects of GNSS, measuring systems and positioning, mainly in the framework of International Federation of Surveyors (FIG, Fédération Internationale des Géomètres). Additionally, from 2015 to 2021 he served as a Dean of the Faculty of Aerospace Engineering and Geodesy of the University of Stuttgart. In 2019 he was awarded a Dr. h.c. degree at the Technical University of Civil Engineering Bucharest in Romania. Since 2022, he has been an honorary member of FIG.
\end{IEEEbiography}

\vfill

\end{document}